\documentclass[10pt,twocolumn]{article}

\usepackage{epsfig}
\usepackage{graphicx}
\usepackage{times}
\usepackage{amsthm,amsmath}
\usepackage{amssymb, verbatim}
\usepackage{fullpage}
\usepackage{ctable}
\usepackage{multirow}

\usepackage[top=2.0cm,bottom=2.0cm,left=2.0cm,right=2.0cm]{geometry}

\newcommand{\abheading}[1]{\vspace{0.1cm}\noindent\textbf{#1}}

\begin{document}

%%%%%%%%% TITLE
\title{Review of Statistical Shape Spaces for 3D Data with Comparative Analysis for Human Faces}

\author{Alan Brunton\thanks{Fraunhofer Institute for Computer Graphics Research IGD, Germany, alan.brunton@igd.fraunhofer.de} \and Augusto Salazar\thanks{Grupo Autom\'{a}tica y Electr\'{o}nica, Instituto Tecnol\'{o}gico Metropolitano - ITM, Colombia, augustosalazar@itm.edu.co} \and Timo Bolkart\thanks{MMCI, Saarland University, Germany, \{tbolkart, swuhrer\}@mmci.uni-saarland.de} \and Stefanie Wuhrer\footnotemark[3]}

%\begin{keyword}
%statistical shape spaces \sep statistical model fitting \sep 3D data \sep generative models \sep 3D vision
%\end{keyword}

\date{}

\maketitle

\begin{abstract}
With systems for acquiring 3D surface data being evermore commonplace, it has become important to reliably extract specific shapes from the acquired data. In the presence of noise and occlusions, this can be done through the use of statistical shape models, which are learned from databases of clean examples of the shape in question. In this paper, we review, analyze and compare different statistical models: from those that analyze the variation in geometry globally to those that analyze the variation in geometry locally. We first review how different types of models have been used in the literature, then proceed to define the models and analyze them theoretically, in terms of both their statistical and computational aspects. We then perform extensive experimental comparison on the task of model fitting, and give intuition about which type of model is better for a few applications. Due to the wide availability of databases of high-quality data, we use the human face as the specific shape we wish to extract from corrupted data.
\end{abstract}

%--------------------------------------------------------------------------------------------------------------------------------------------------------------------------------------------------------

\section{Introduction}

Whether through laser range scanners, stereo reconstruction or structured light, methods and systems for 3D sensing and acquisition are now commonplace. However, these systems typically incur some amount of noise. Further, if we are interested in a particular object within data, it may be occluded by other objects. To extract the shape of an object of a particular class from such data it is often advantageous to use a statistical shape model to ensure that the extracted shape is valid for that object class. Another way of saying this, is that the shape is restricted to lie in a statistical shape space. In this paper, we review and describe, within a common mathematical framework, a wide variety of statistical shape spaces used for robust fitting to noisy and ambiguous data. We further perform a thorough theoretical analysis and experimental comparison of variants of such a model in which statistics are learned over global and local extents.

Such a statistical model must be learned from a database of consistently parametrized (i.e. registered) instances from the object class in question. One class of shape for which there exist a number of available databases is the shape of the human face, and this is the shape which we use for evaluation in this paper. Furthermore, the 3D shape of human faces is important to a wide variety of applications, ranging from tele-presence to virtual avatar control to face recognition. However, we emphasize that the principles, models and algorithms discussed in this paper are applicable to any class of shapes for which a database containing parametrized data is available.

The main reason to use a statistical shape model to fit to data, instead of a template fitting with a non-rigid iterative closest point (ICP) approach and some kind of regularization constraint, is that by learning a statistical model for a class of shapes, we can significantly reduce the search space, which results in the ability to reconstruct the underlying shape in the presence of severe noise or occlusions. These and other types of ambiguities are often present in real-world data captured in uncontrolled environments.

The purpose of this paper is to provide researchers, engineers and end-application developers interested in employing existing statistical models or developing new ones with a targeted, focused review that includes a thorough analysis and comparison of the costs and benefits of two statistical models for the task of model fitting to noisy, corrupted or incomplete 3D data. The task of fitting a 3D shape model to ambiguous data is important for many applications, such as recognition tasks (identity and expression recognition in the case of faces, sex recognition and gait analysis in the case of human bodies), tele-presence, virtual avatar control, segmenting and extracting organ shapes from medical images for surgical planning or diagnosis, and so on. We specifically maintain a targeted scope to provide an in-depth analysis of the behaviors of various statistical models. Toward this end, we analyze different statistical models from those that model global geometry variation to those that model local geometry variation. We experimentally compare a purely global model to a purely local model, thereby providing a quantitative comparison of two models on either end of the spectrum of commonly used statistical models. Thus, the contribution of this paper is three-fold:
\begin{itemize}
	\item An analysis of the theoretical properties, within a common mathematical framework, of a wide variety of seemingly very different statistical shape spaces.
	\item A quantitative and qualitative analysis, and extensive comparative evaluation of the practical performance of both global and local statistical shape models.
	\item We publish the learned statistical models and code to use them~\cite{bsbw:statmods:2013}, thus allowing others to try them out and potentially fit them to other input modalities and for any application.
\end{itemize}

We begin by reviewing the statistical shape spaces, or statistical shape models, that have been used for human face shapes, human body shapes, and medical data (Section \ref{sec_related}). We give a common mathematical framework in which these models relate to one another and discuss the key differences between them. We then give a mathematical description of the process of learning or training a statistical shape space, and an in-depth focus on two specific models for human faces that represent extremes in the range of models (Section \ref{sec_model_training}). This is followed by a mathematical description of the process of fitting a statistical model to ambiguous, partial or corrupted data via energy minimization (Section \ref{sec_model_fitting}). Again, this is accompanied by an in-depth analysis of the process for the two shape spaces. We then provide an extensive experimental comparison of the two models (Section \ref{sec_comp_eval}), followed by a discussion of the implications of the results (Sections \ref{sec_impl_obs} and \ref{sec_conclusions}).

%--------------------------------------------------------------------------------------------------------------------------------------------------------------------------------------------------------
\section{Statistical Shape Spaces}
\label{sec_related}

To extract an object from noisy input data, it helps to have a small basis in which the shape of the object can be represented. Traditionally, such bases have been generated for specific classes of models by artists. An example of such an artist-generated basis are blendshape models (as for instance used by Li et al.~\cite{Li2010}), which can be used to encode a face performing different expressions. This way of generating a basis requires expertise about the possible deformations, and modeling the shapes is tedious. More recently, machine learning has been used to find a small basis from a set of training shapes using statistical analysis. This gives an easy and fully automatic way to find a small set of basis functions.

In this vein, we formulate statistical shape analysis of a given object class as the task of finding a \emph{statistical shape space} that efficiently and informatively represents the shape of objects of that class. We define a statistical shape space as a \emph{shape space} equipped with a probability distribution, or \emph{prior}, measuring how likely it is that an object of the given class would have a particular parametric representation in the shape space. The shape space itself is defined by the set of coefficients obtained by projecting the shapes onto the set of basis functions. (We use the terms basis functions and basis vectors somewhat interchangeably in the following; strictly speaking a basis function is only relevant for continous surfaces, and in practice basis vectors are used for discrete data.)

Thus, we focus on statistical shape analysis as a \emph{generative} technique. A surface containing $n$ vertices in $\mathbb{R}^3$ is represented by $d$ shape parameters or coefficients, which form a vector $\textbf{s}\in\mathbb{R}^d$. A generator function 
\begin{equation}
\textbf{F}(\textbf{s}):\mathbb{R}^d\rightarrow\mathbb{R}^{3n}
\label{eq:generator}
\end{equation}
generates from these shape parameters a surface representation (either mesh or point cloud) of $n$ vertices. These shape parameters, and by extension the surface, can be fit to input data of varying modalities (3D point clouds, 3D voxel images, 2.5D depth, 2D images, sparse measurements, etc.), so long as there is a way to measure the distance between the surface and the data, or the quality of the fitting.

As we see in the following review, by far the most common form of statistical analysis used for shapes is principle component analysis (PCA), which seeks a basis in which variance of the training data is maximized. The resulting basis vectors are the directions of greatest variation within the training data. Projecting the training samples onto this basis results in a diagonal sample covariance matrix. If the underlying distribution of the data is assumed to be multi-dimensional Gaussian, then this corresponds to the maximum likelihood estimate of the parameters of the density function. As a result, the resulting shape space is often equipped with a Gaussian prior. If this assumption does not hold, then a Gaussian prior may be arbitrarily far from the true prior, and choosing the correct prior may be challenging. A mathematical description of how to learn a statistical shape space using PCA is given in Section \ref{sec_model_training_global}. 

The core aspect that varies between statistical shape spaces is the space in which PCA is performed. Careful selection of the space in which to perform PCA can lead to significantly better statistical properties. For many models, this amounts to a change of basis, followed by separate analyses in different sub-spaces of the transformed space. The resulting basis of the learned space is then formed by composing the decomposition basis with the sub-space PCA basis (directions of greatest variation within the subspace). For others, this amounts to a nonlinear transformation followed by a global analysis. 

This section reviews work on performing statistical shape analysis of 3D data for image processing applications. The categorization of the statistical models in different application domains is summarized in Table \ref{tab_related}, where models and methods are grouped by the type of data they were applied to and by the extent of the basis functions used. The table further contains information on whether the models were designed to analyze articulation variations separately from shape variations (here, articulation can refer to facial expression, body posture, or the pose of bones before and after an operation). 

In this study, we focus on shape variations over a sample from a population, and hence in Sections \ref{sec_model_training_global}, \ref{sec_model_training_local} and we analyze and compare statistical models for faces without expression variations. In Section \ref{sec_comp_eval} we perform an extensive experimental comparison of the models. This allows us to better examine the differences between the statistical models themselves with respect to the model-fitting task.

The first step to performing shape analysis is to acquire and register a set of \emph{training shapes} that capture the shape variability that is of interest for a particular application. Subsequently, statistical shape analysis is performed on the registered training shapes: the shapes are projected onto a basis of choice and a probability distribution is fitted to the resulting coefficients to obtain a prior distribution for the shapes of interest. Without correspondence information, this statistical analysis is not possible. However, as indicated in the last column of Table \ref{tab_related}, a few methods simultaneously compute a parametrization of a population of shapes while building a statistical model. 

Computing correspondences between a population of shapes is a challenging problem, and a detailed discussion about possible approaches is beyond the scope of this work. We refer the reader to recent surveys~\cite{VanKaick2011,tam_survey_2013} for more information. However, we emphasize that the quality of the registration greatly affects the quality of the resulting statistical models, and by using a high-quality registration in this study, computed as discussed in Section \ref{sec_exper_setup}, we are able to better analyze the properties of models themselves rather than the effects of gross mis-registration.

In computer vision, statistical 3D shape models are commonly used to infer the three-dimensional shape of an object from images, mostly for the purpose of image manipulation. While recently, different classes of shapes have been considered~\cite{cashman_fitzgibbon_shape_dolphins,alcantara_pami2009}, shape models of human faces and human bodies are of special interest due to their immense applicability in human--machine interaction. In medical image analysis, statistical shape models are commonly used to segment medical images and to find correspondences and abnormalities of anatomical shapes. In the following, we review statistical shape spaces used to analyze human faces (Section \ref{sec_related_faces}), human bodies (Section \ref{sec_related_bodies}) and medical data (Section \ref{sec_related_medical}).

\begin{table*}
\centering
{\small
\begin{tabular}{l l l l l }
\hline
Type of data	& Influence of 		& Methods & Articulation variation	& Simultaneous \\
							& basis functions & 			  & analyzed separately			& parametrization \\
\hline
Faces \\
				& global										& Morphable model (PCA)~\cite{Blanz1999,amberg_etal_iccv07,patel_smith_morphableModelRevisited_09,Yang2011} & $-$ & $-$ \\
				& global										& Morphable model (PCA)~\cite{amberg_etal_fg08} & $\surd$ & $-$ \\
				& global										& Multilinear model~\cite{Vlasic2005,Dale2011,Yang2012,Bolkart2013} & $\surd$ & $-$ \\
				& part-based								& Part-based model~\cite{basso_verri_2007,haar_veltkamp_2008,smet_vanGool_2010} & $-$ & $-$ \\
				& part-based								& Part-based model~\cite{kakadiaris_etal_2007_deformable_model} & $\surd$ & $-$ \\
				& localized detail					& Hierarchical pyramids~\cite{Golovinskiy_2006} & $-$ & $-$ \\
				& local											& Local wavelet model~\cite{Brunton2011} & $-$ & $-$ \\
\hline
Bodies \\
				& global										& PCA model~\cite{Allen2003,seo_etal_shape_from_silhouette,chen:learning,boisvert_etal_shape_from_silhouette,wuhrer_shu_acc_shape_measurement} & $-$ & $-$ \\
				& global										& SCAPE model~\cite{anguelov_srinivasan_koller_thrun_rodgers_05_shapecomp,guan_etal,balan_black_08_naked_truth,ParametricReshaping2010,Jain:2010:MovieReshape,weiss_etal_bodyShapeFromKinect_2011} & $\surd$ & $-$ \\
				& global										& SCAPE model~\cite{Hirshberg_2012} & $\surd$ & $\surd$ \\
				& global										& Rotation-invariant encoding~\cite{HasStoSunRosSei09,hasler_etal_smi09} & $\surd$ & $-$ \\
				& global										& Multilinear model~\cite{Hasler2010} & $\surd$ & $-$ \\
				& global										& Posture-invariant model~\cite{wuhrer_etal_2012_pose_inv_statistics} & $-$ & $-$ \\
				& part-based								& Segmented PCA model~\cite{Xi_etal_segmented_body_2007} & $-$ & $-$ \\
				& part-based								& Part-based multilinear model~\cite{Chen_2013_CVPR} & $\surd$ & $-$ \\
\hline
Medical Data \\
				& global										& Active shape model (PCA)~\cite{cootes_etal_95_ASM,cootes_taylor_01_asm} & $-$ & $-$ \\
				& global										& Active shape model (PCA)~\cite{davies_twining_taylor_mdl} & $-$ & $\surd$ \\
				& global										& PGA model~\cite{fletcher_etal_pga_2004} & $-$ & $-$ \\
				& part-based								& Part-based model~\cite{toews_etal_06} & $-$ & $-$ \\
				& part-based								& Part-based multilinear model~\cite{lecron_etal_2012} & $\surd$ & $-$ \\
				& local											& Local wavelet model~\cite{davatzikos_etal,nain_etal_MICCAI05,nain_etal_MICCAI06,li_etal_CVPR07,hierarchical-diffusion-wavelet-shape-priors,cortical_folding} & $-$ & $-$ \\
\hline
\end{tabular}
}
\caption{Organization of statistical shape fitting methods reviewed.}
\label{tab_related}
\end{table*}

\subsection{Human Face Shapes}
\label{sec_related_faces}

We start our review by summarizing the use of statistical shape models of human faces. Blanz and Vetter~\cite{Blanz1999} proposed the first statistical shape model for 3D models of human faces. The model, called \emph{morphable model}, captures both 3D shape and texture information and can be used to predict a 3D face shape from a single input image. Texture is an important cue for human faces, and can greatly help with fitting directly to images. However, in this paper, we focus on statistical analysis of \emph{shape} for clarity. A parametrized database of 3D face scans, mostly in neutral expression, is used to learn the statistical model using standard PCA. Given an input image in neutral expression, the learned shape space is searched to find the textured shape that best explains the input image using an optimization technique. This successful approach resulted in multiple follow-up works~\cite{amberg_etal_iccv07, patel_smith_morphableModelRevisited_09}. Amberg et al.~\cite{amberg_etal_fg08} extended the morphable model to model expression variation by computing offsets from the neutral pose and performing PCA over these offset surfaces. With this method, they are able to include expression variation while maintaining a single linear model.

Blanz and Vetter~\cite{Blanz1999} experimented with manually segmenting the morphable model into four regions. A morphable model is then learned for each region and the regions are fitted to the data independently and merged in a post-processing step. This \textit{part-based} model was shown to lead to a higher data accuracy than the global morphable model~\cite{Blanz1999}. As this approach is suitable to obtain good fitting results in localized regions, it has been used in multiple follow-up works~\cite{basso_verri_2007,kakadiaris_etal_2007_deformable_model,haar_veltkamp_2008}. All of these methods proceed by manually segmenting the faces, by learning an independent morphable model for each part, and by fitting the parts to data independently. In this case, a basis arises out of the set of indicator functions on the segments, which are then composed with the union of PCA bases on the segments. As a result, the areas at the boundaries of the individual parts are not necessarily continuous, and a post-processing step is used to merge the fitted patches. Smet and Van~Gool~\cite{smet_vanGool_2010} proposed a similar method where the segmentation is found automatically by clustering the vertices based on features derived from their displacements over the training set. To address the potential discontinuities at the boundaries of the segments, they smoothly weight the segments to obtain regionalized basis functions for the training data. To estimate the optimal weights, a complex iterative algorithm is used. In each iteration, the training set is randomly partitioned into two disjoint subsets, where one set is used as test set to be reconstructed from the model learned on the other. This is followed by a two-step procedure. First, optimal reconstruction coefficients are estimated by weighted least-squares. Second, the optimal weights are estimated by solving many independent linear systems. Since for high-resolution training data this iterative algorithm converges slowly, they employ a coarse-to-fine approach to accelerate convergence.

Brunton et al.~\cite{Brunton2011} used a statistical analysis based on wavelet models to learn many localized independent prior distributions at multiple scales for the 3D shape of human faces. This allows to capture and combine localized shape variations in different areas of the face in a multi-resolution framework independently. They used this information to predict the 3D face shape from point clouds generated by stereo reconstruction. In this model, a basis is formed by composing the wavelet basis with the local PCA bases. As in the work by Blanz and Vetter, all faces are assumed to have a neutral facial expression. While PCA models such as the morphable model and part-based PCA models are commonly used in computer vision to model human faces, localized prior distributions based on wavelet models are only beginning to influence this field. In the following, we will refer to this method as \textit{local wavelet model}.

Golovinskiy et al.~\cite{Golovinskiy_2006} proposed a  statistical model based on hierarchical pyramids to synthesize geometric facial details. This statistical model, which allows for spatially varying geometric detail across the face, models the difference between a smooth face and a high-resolution face with geometric details, such as wrinkles.

To allow for varying facial expressions, Vlasic et al.~\cite{Vlasic2005} used a statistical method based on tensor algebra called the \textit{multilinear model} to analyze a set of faces captured of subjects performing a variety of facial expressions. This approach can be viewed as an extension of morphable models to higher dimensions. It is formed by taking the Cartesian product of two linear statistical shape spaces that capture variations according to two different attributes (i.e. identity and expression). The resulting basis is then the Cartesian product of the two bases associated with each attribute. This model allows for the prediction of a 3D face shape in multiple possible expressions from a single photograph. 

Yang et al.~\cite{Yang2011} exchanged the expression of a face in a single image based on a different input image of the same subject. For this application, they built multiple PCA spaces (one per expression) and combined these spaces for their application. A follow-up paper~\cite{Yang2012} used a multi-linear model to enhance or dampen expressions in videos.

More recently, much work has focused on extracting a set of frames of three-dimensional face shapes from a video stream showing a face. The output of this type of algorithm is a four-dimensional sequence showing the three-dimensional face shape in motion. Dale et al.~\cite{Dale2011} extended the method by Vlasic et al. to compute such a four-dimensional sequence. This information is then used to exchange faces in video sequences; either keeping the sequences of expressions and replacing the identity of the subject, or keeping the identity and replacing the sequence of expressions. Bolkart and Wuhrer~\cite{Bolkart2013} learn a multi-linear model and fit it to sequences of 3D faces performing various expressions, producing a 4D parametrization, which can be used to animate a static scan with a specified expression.

Another avenue of recent work is to track two-dimensional range images over time. These images can be captured using depth sensors, such as the Kinect sensor. Weise et al.~\cite{Weise2011} proposed a method to track a three-dimensional face model over time using prior information on the deformation model, and to use the tracked model to drive the animation of a virtual character in real-time. Unlike the other methods discussed in this section, this method learns a statistical prior that is subject-specific. In this paper, we do not consider subject-specific priors.

\subsection{Human Body Shapes}
\label{sec_related_bodies}

Allen et al.~\cite{Allen2003} proposed a statistical model for human body shapes acquired in a similar posture that is similar to the morphable face model introduced by Blanz and Vetter. One main difference is that Allen et al. only learn information about the 3D shape and not about texture. We denote this model by \textit{global PCA model} in the following. This model has been used to predict a 3D human body shape in a similar posture from one or more images~\cite{seo_etal_shape_from_silhouette, chen:learning, boisvert_etal_shape_from_silhouette} and from measurements~\cite{wuhrer_shu_acc_shape_measurement}. Xi et al.~\cite{Xi_etal_segmented_body_2007} used a part-based PCA model to infer body shapes from silhouettes. Wuhrer et al.~\cite{wuhrer_etal_2012_pose_inv_statistics} performed PCA on a local shape representation based on the Laplace operator to analyze human body shape variations independently of posture changes. This is an example of a non-linear transform followed by a global PCA in the transformed space, where the transformed space has been chosen to be invariant to local rigid deformations, making the resulting statistical analysis posture invariant.

To allow for posture variation, Anguelov et al.~\cite{anguelov_srinivasan_koller_thrun_rodgers_05_shapecomp} proposed the \emph{SCAPE model}. This model learns a PCA shape space for body shape variations using a database containing multiple subjects in a similar posture. Furthermore, the model learns a mapping from posture parameters (based on a skeleton) to shape changes using a database containing one subject in multiple poses. The model then combines the two variations using the assumption that body shape and posture are decorrelated. Since the SCAPE model successfully models human body shape and posture, it has been used to predict a 3D body shape in arbitrary posture from a single image~\cite{guan_etal}. Furthermore, this model can be used to predict a 3D human body shape in arbitrary posture based on a set of input images of a dressed person~\cite{balan_black_08_naked_truth}. Such a 3D prediction can then be used to modify the input image~\cite{ParametricReshaping2010}. Just like in the case of human faces, more recently, much work has focused on finding a four-dimensional sequence of three-dimensional human body shapes in motion from a video sequence. Jain et al.~\cite{Jain:2010:MovieReshape} used this to modify human body shapes in video sequences. Weiss et al.~\cite{weiss_etal_bodyShapeFromKinect_2011} proposed to use the SCAPE model to compute a 3D body scan from noisy Kinect data.

A recent approach by Hirshberg et al.~\cite{Hirshberg_2012} treats the problems of learning a SCAPE model and computing point-to-point correspondences between a set of training data simultaneously by solving a single variational problem.

A different avenue to allow for posture variation is to model shape and posture changes as correlated. This assumption is relevant, since the difference of the human body shape of the same subject in different postures depends on the body shape, e.g. on how muscular the subject is. Hasler et al.~\cite{HasStoSunRosSei09} proposed a shape space that jointly captures shape and posture variations by performing PCA on a rotation-invariant encoding of the shapes. This shape space is then used to predict the body shape of a dressed subject~\cite{hasler_etal_smi09}.

An alternative for correlating human shape and posture variations is to use a multilinear model (as Vlasic et al.~\cite{Vlasic2005} do for face shapes). This avenue was explored by Hasler et al.~\cite{Hasler2010} for the application of predicting 3D body shape and posture from an image. Chen et al.~\cite{Chen_2013_CVPR} proposed a part-based multilinear model using a manual segmentation into body parts. 

\subsection{Medical Data}
\label{sec_related_medical}

In medical imaging, one is especially interested in a body part, such as an organ or part thereof. Tasks of interest include finding the shape of interest in a medical image. This decomposition of the image is often called \emph{segmentation}. To solve this task, Cootes et al.~\cite{cootes_etal_95_ASM} propose the use of a statistical prior called \emph{active shape model}, which is similar to the morphable model for faces. The active shape model learns the distribution of a set of registered and aligned training shapes using PCA, and uses this prior information to segment a given medical image. This model is commonly used for image segmentation, see Cootes and Taylor~\cite{cootes_taylor_01_asm} and references therein.

The quality of the registration of the training shapes directly influences the quality of the statistical shape model. Davies et al.~\cite{davies_twining_taylor_mdl} used this observation to derive an approach that jointly optimizes the registration and a shape model built using PCA. The main idea of this approach is that a good shape model should have a small information-theoretic description length. Hence, to simultaneously parameterize the shapes and build a statistical model, the approach by Davies et al. solves a variational problem that aims to minimize the description length of the PCA model.

PCA assumes that the data can be approximated well using a linear model. However, in medical imaging, many data sets form a non-linear manifold in a high-dimensional space. To capture the structure of this high-dimensional manifold, Fletcher et al.~\cite{fletcher_etal_pga_2004} generalized PCA to this setting. This approach called \emph{principal geodesic analysis (PGA)} considers geodesic distances between shapes measured along the high-dimensional manifold instead of Euclidean distances.

One problem with active shape models is that they capture global shape variations. In medical imaging, one is often interested in detecting localized shape anomalies, as these can give insights in whether or not a specific organ is affected by a disease, for instance. In an active shape model, such local variations may be distributed over several principal components, and they may be controlled by principal components that capture a small percentage of the overall shape variability. 

To remedy this, part-based models have been proposed for medical imaging. Toews et al.~\cite{toews_etal_06} proposed a part-based model that captures the statistical variations of geometry and color. This model is used to analyze MRI volumes of the brain. Lecron et al.~\cite{lecron_etal_2012} used a different part-based model to analyze variations of the spine both within and across subjects. Each vertebra is considered one part and analyzed using a multilinear statistical model.  

Part-based models work well for applications where a segmentation into parts is meaningful. However, in many medical imaging tasks, a natural segmentation is difficult to define. To address this problem, Davatzikos et al.~\cite{davatzikos_etal} used statistical analysis based on local wavelet models to learn a localized prior distribution of contours in images. Here, the localized regions with large variations are found automatically and do not need to be predefined. Nain et al.~\cite{nain_etal_MICCAI05} extended this technique to use wavelets to perform a statistical analysis of three-dimensional shapes. Shape priors based on different types of wavelet models have been used to segment medical images~\cite{nain_etal_MICCAI06, li_etal_CVPR07, hierarchical-diffusion-wavelet-shape-priors}. Yu et al.~\cite{cortical_folding} show that statistical wavelet models can be used to analyze cortical folding patterns, which is a challenging task.

\subsection{Comparison}
\label{sec_related_comparison}
In the following, we provide an analytical comparative evaluation of a global PCA model and a local wavelet model. By doing so, we compare two methods on either end of the spectrum of the commonly used statistical shape models. It is already known~\cite{Blanz1999} that manually segmented part-based decompositions improve fitting accuracy for human face shapes. A thorough numerical comparison has not been performed for automatically and object-class non-specific decompositions such as wavelets. We expect that, like part-based models, the local wavelet model will obtain locally more accurate results than the global model, since the basis functions have limited extent. However, since the part-decomposition for statistical models is to our knowledge always shape-specific, the local model can more readily be adapted to various application areas. In particular, when segmenting volumetric medical data for a particular organ or part of the brain, it may be impossible to identify distinct parts on which part-specific statistical models can be learned. However, one may still want localized fitting for these shapes.

Applications of statistical model fitting are wide-ranging and require very different evaluation criteria. To evaluate them all would be impractical, and to evaluate one or two would unduly reduce the scope of this study. Our comparison (Section \ref{sec_comp_eval}) therefore uses as a test meta-application the task of model fitting to point cloud data of human faces, and from the results we give an intuition of which model would be most appropriate for which end application (Section \ref{sec_impl_obs}). All of the figures and tables in the following sections depend on the specific training data used, but they nonetheless illustrate how the geometric information captured in the statistical models can be evaluated and visualized. This comparison can provide a guide of how the locality of the basis functions of a model influence its performance for model fitting. Reconstruction of 3D shape from silhouettes or images, or segmentation of volumetric data, can also be viewed as a model fitting, with additional or different ambiguities, and therefore we infer that we can reasonably expect the statistical models to behave similarly in these scenarios.

Furthermore, we allow those developing end applications to test the performance of both models in their specific scenario by publishing the statistical models and code to use them online~\cite{bsbw:statmods:2013}. The data and code provided will also allow others to derive fitting energies and code for different input modalities.

%-------------------------------------------------------------------------------------------------------------------------------------------------------------------------------------
\section{Learning a Statistical Shape Space}
\label{sec_model_training}

\begin{figure*}
\centering
\includegraphics[width=12.0cm]{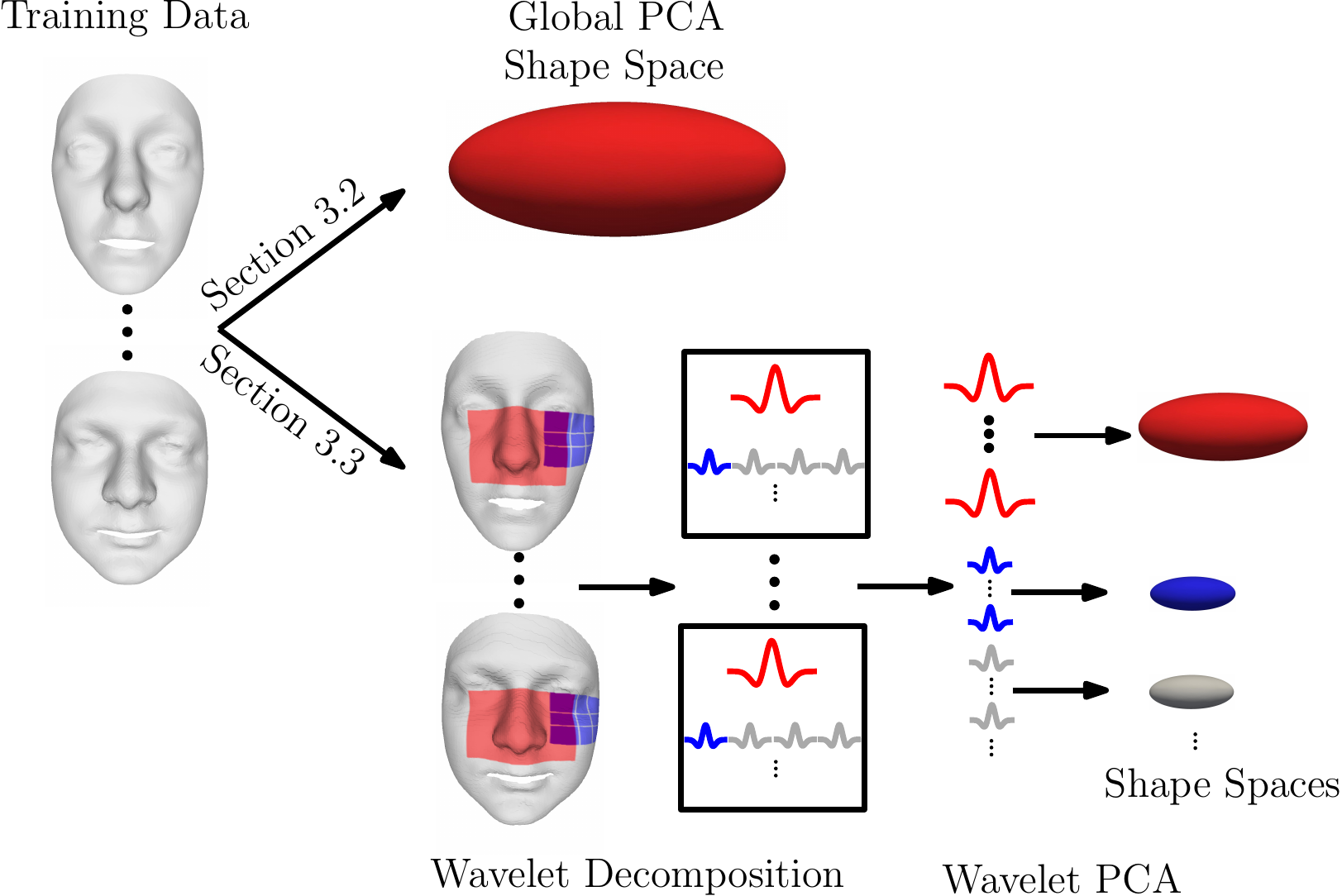}
\caption{\emph{Overview of the model training.}}
\label{fig_training}
\end{figure*}

For learning a statistical shape space, or training a statistical shape model, we assume we are given $T$ training shapes in full correspondence. Figure~\ref{fig_training} gives an overview of the model training. The key difference between the two models discussed in detail here is the basis in which a prior probability distribution is fit to the training data. 

We pre-align the data to remove rotation, translation, and uniform scale differences using generalized Procrustes analysis (GPA)~\cite{dryden_mardia_shape_analysis}. Note that by removing uniform scale differences, we only consider shape differences and not size differences of the models. For data-fitting this is desirable due to different measurement units used by different acquisition systems, but in general this is application dependent~\cite{dryden_mardia_shape_analysis}. GPA iteratively aligns each model to the mean shape and recomputes the mean. Removing transformations that are not of interest using GPA is an important pre-processing step that yields better statistical models.

Learning a statistical shape space requires determining the basis functions or vectors, which define the shape space, and fitting a probability distribution to the resulting shape space coefficients from the training set. PCA-based methods perform these two steps simultaneously, selecting as a basis the directions of greatest variations in the data and computing a diagonal sample covariance matrix for the data projected onto this basis, which corresponds to a maximum likelihood estimate of a multi-dimensional Gaussian distribution. Part-based and wavelet-domain methods decompose the shapes into a localized basis before proceeding with PCA. The result is a basis consisting of the localized basis functions composed with the localized principal components. The learned prior is the product of the localized multi-dimensional Gaussian distributions. Note that the assumption of a Gaussian density function is only introduced when equipping the shape space with a Gaussian prior.

With this framework, and by restricting ourselves to linear shape spaces for the purposes of this study, the generator function Eq.~(\ref{eq:generator}), can be written as a combination of the basis functions
\begin{equation}
\label{eq:generator:basis}
\textbf{F}(\textbf{s}) = \overline{\textbf{F}} + \Phi\textbf{s} = \overline{\textbf{F}} + \sum_{i=1}^d \Phi_i s_i
\end{equation}
where $\overline{\textbf{F}}$ is the mean shape computed over the training set, $\Phi\in\mathbb{R}^{3n\times d}$ is a matrix, $\Phi_i\in\mathbb{R}^{3n}$ are its columns, and as before $\textbf{s}\in\mathbb{R}^d$ is a vector of shape parameters. It is precisely the choice of $\Phi$ that determines the properties of the shape space, and determines the prior distribution learned from the training samples.

\subsection{Evaluating a Statistical Model}
\label{sec_model_training_eval}

If a statistical model with a small number of basis functions is fitted to a face, the result contains little shape detail, because the model only represents a small proportion of the variability of the training data. Keeping a large number of basis functions may cause the model to overfit to the training data. That is, the learned space may contain a bias towards shapes present in the training set. Overfitting can occur when the model is underconstrained by the training samples. This may occur if the model itself has too many degrees of freedom, or when it is learned in a very high-dimensional space, such as when the model is computed directly from very high-dimensional training samples.

To pick a number of basis functions $d$ that preserves a high amount of variability yet does not overfit the training data, we use the following three error measures similar to \textit{compactness}, \textit{generalization}, and \textit{specificity}~\cite{Styner2003}. We use a slight modification of the original error measures to obtain results that are independent of the size of the training data.

\textit{Compactness} measures  how much variability of the training data is explained by the learned statistical model. That is, we want to measure what fraction of the total variability of the training data is captured by $d$ model parameters. This provides a measure of how well a given number of parameters explains the training data. 

\textit{Generalization} measures the ability of the model to represent data, which are not part of the training set. To calculate this measure, we learn a PCA model on a subset of the training data, where one subject is excluded. The excluded subject is projected to the PCA space, reconstructed, and the distance between the source and the reconstruction is measured. To measure the distance between two faces, we use the average Euclidean vertex distance computed between all corresponding vertices. We perform this measurement for all subjects. The mean and standard deviation are then considered.

\textit{Specificity} measures the similarity between reconstructions from the statistical model and the training data. This estimates the plausibility of a random face represented using the learned shape space. To calculate specificity we choose a set of random points sampled from the probability distribution of the learned statistical shape space. For each of these points we reconstruct the shape using Eq.~(\ref{eq:generator:basis}) and compute the distance to the closest face in the training data. The distance between two faces is computed as above. The mean and standard deviation for the random sample are then considered.

\subsection{Global PCA}
\label{sec_model_training_global}

Principal component analysis aims to reduce the complexity of a set of data. Due to its simplicity it is widely used for shape analysis. PCA is a linear transformation of a set of vectors from $\mathbb{R}^{3n}$ to $\mathbb{R}^{d}$ with $d < 3n$. A vector $\textbf{f} \in \mathbb{R}^{3n}$ is expressed by the scalar weights $s_i$ in a $d$-dimensional subspace, spanned by the orthogonal vectors $\textbf{V}_i$, by
\begin{equation}
\textbf{F}(\textbf{s}) = \overline{\textbf{F}} + {\sum_{i=1}^{d} s_i \textbf{V}_i}.
\end{equation}
For each parameterized shape of the training set we have one vector $\textbf{F}_i^{(train)} \in \mathbb{R}^{3n}$ that contains an ordered coordinate set of all points of the $i$-th training shape. The vectors $\textbf{V}_i$ are the eigenvectors of the data covariance matrix 
\begin{equation}
\Sigma_\textbf{F} = \frac{1}{T} {\sum_{i=1}^n (\textbf{F}_i^{(train)}-\overline{\textbf{F}})(\textbf{F}_i^{(train)}-\overline{\textbf{F}})^T },
\end{equation}
where $\overline{\textbf{F}}$ is the mean of the training data. The eigenvectors $\textbf{V}_i$ are ordered with respect to the non-increasing corresponding eigenvalues $\lambda_i$. The eigenvalues $\lambda_i$ measure the variability captured by the $i$-th principal component. More specifically, $\textbf{V}_i$ captures $100\frac{\lambda_i}{\sum_{i=1}^{T-1} \lambda_i}\%$ of the variability of the training data. The rank of the data covariance matrix is at most $\min(3n-1, T-1)$ and therefore the number of distinct non-zero eigenvalues and hence, the number or principal components, is at most $\min(3n-1, T-1)$. 

Thus, we get our basis directly from the data via the principal components: in matrix form $\Phi_G$ has columns $\Phi_{Gi}=\textbf{V}_i$, for $i=1,\ldots,d$, where $d\leq\min(3n-1, T-1)$. Note that every basis vector $\Phi_{Gi}$ has global support in general: all $3n$ elements are in general non-zero, and every vertex is influenced.

\begin{figure}[ht]
\centering
\includegraphics[width=6.0cm]{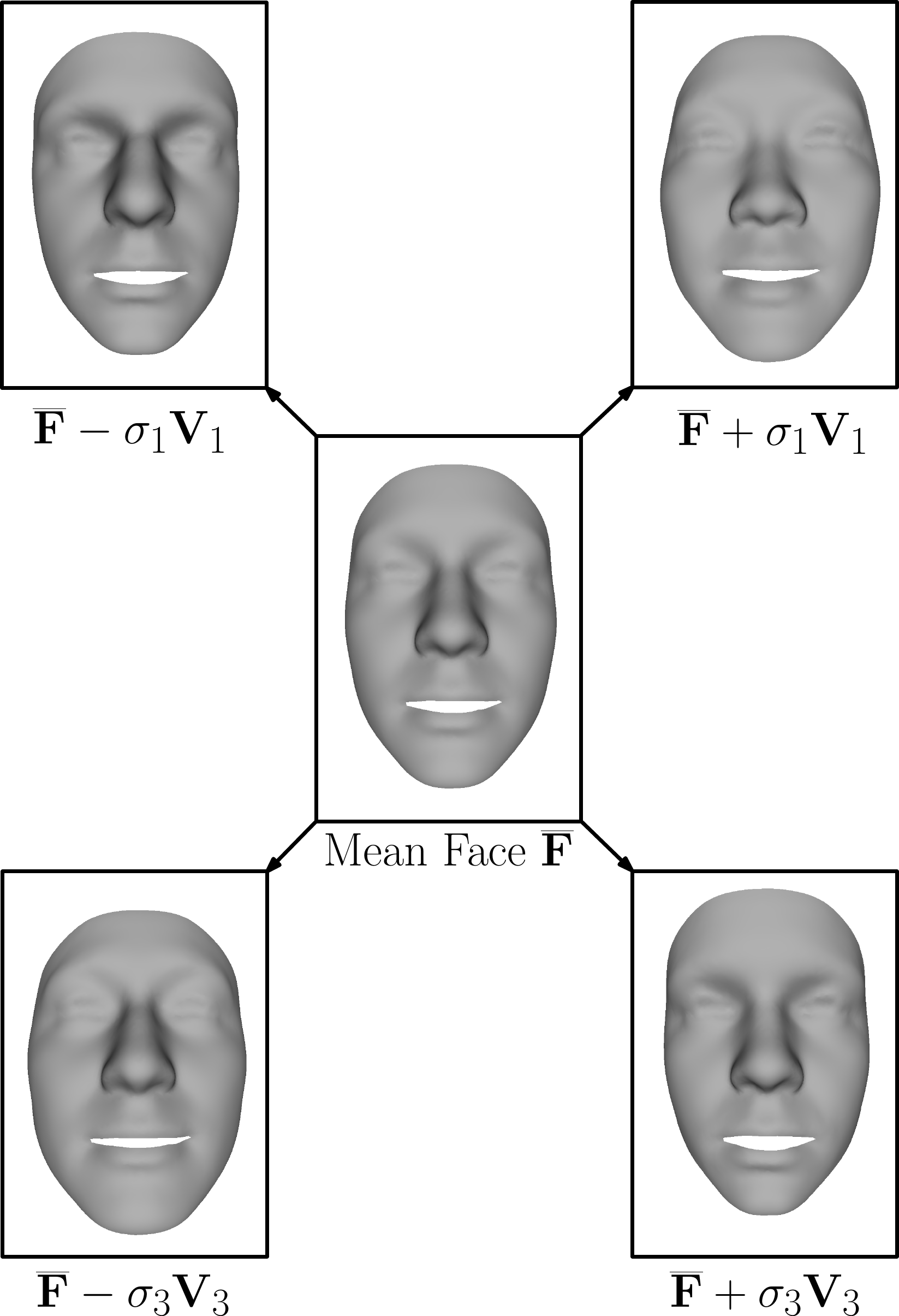}
\caption{\emph{Variations of two principal components.}}
\label{fig_Variation_plot}
\end{figure}

\abheading{Properties} The global shape space represents the high-dimensional differences of the training faces in a low-dimensional shape space that is spanned by the corresponding eigenvectors of the $d$ largest eigenvalues of the data covariance matrix. 

Figure~\ref{fig_Variation_plot} visualizes the variations along two principal components in the range of $-3\sigma_i$ to $+3\sigma_i$, where $\sigma_i$ denotes the standard deviation of the $i$-th principal component.

The amount of details that can be expressed by the global statistical model is limited by the details present in the training data. It would be useful in some applications to increase the variability of the training data by increasing the mesh resolution of the training data by inserting new vertices as linear combinations of existing vertices. Unfortunately, this is not possible using this global approach. If a vertex expressed as a fixed linear combination of existing vertices is inserted to each training mesh, the corresponding additional vertex in the fitted surface is identical to the corresponding fixed linear combination. Therefore, if an additional vertex is chosen to be placed on a triangle, the corresponding additional vertex is located on the corresponding triangle of the fitted result. Hence, using fixed linear combinations to add points to the surface of the model and fitting this extended surface to a target face leads to the same result as fitting the original model to the target face and adding the points into the resulting surface using the fixed linear combinations. This is a key difference to the local model reviewed in Section~\ref{sec_model_training_local}. 

Computing the compactness statistical measure is straightforward for the global PCA model. For $d$ principal components, compactness is defined as 
\begin{equation}
C \left( d \right) = {\sum \limits_{i=1}^{d} \lambda_i} / {\sum \limits_{i=1}^{T-1} \lambda_i},
\end{equation}
where $\lambda_i$ is the $i$-th eigenvalue of the data covariance matrix. Computation of generalization and specificity are also straightforward.

\subsection{Wavelet PCA}
\label{sec_model_training_local}

The core principle behind wavelet transforms is to project sampled data onto a set of basis functions that are localized in space and frequency. 
In our context, we use the wavelet basis as a prefix, and extract data driven basis for individual wavelet coefficients using PCA.
%Wavelet transforms were originally defined on regularly sampled Euclidean domains~\cite{mallat_wavelet_tour_1999}. 
Second-generation or lifting wavelets~\cite{sweldens_lifting_1996} are computed in time linear in the number of samples in the original signal using local lifting operations and sub-sampling at each scale. 
The samples are partitioned into maximally correlated subsets, \emph{e.g.} odd and even samples for signals on 1D domains. One subset is then used to predict the other, and the residual of this prediction is then called the detail, or wavelet, coefficients. The detail coefficients are then used to \emph{update} the other subset, giving approximation, or scaling, coefficients. The process is repeated on the scaling coefficients. We refer the reader to Sweldens~\cite{sweldens_lifting_1996} for a more thorough explanation.

While wavelets were originally defined regularly sampled Euclidean domains~\cite{mallat_wavelet_tour_1999}, spherical wavelets~\cite{spherical_wavelets} are defined on subdivision surfaces, often topological spheres. A commonly used wavelet basis is a biorthogonal generalized B-spline basis~\cite{bspline_subdiv_wavelets} that uses the Catmull-Clark subdivision scheme and has been applied in multiple application domains~\cite{li_etal_CVPR07,Brunton2011}. The prediction and update operators are B-spline interpolations from the neighboring vertices. This scheme is stable for linear and cubic B-splines; our comparison uses the linear basis as we found that this produced satisfactory results.

Aside from wavelet coefficients, one might use any localized basis as a prefix to PCA or other data-driven bases. However, wavelet bases have the important property that they decorrelate the data, meaning the resulting coefficients can be analyzed separately. Since individual coefficients are of much lower dimension than the whole surface, this greatly reduces the risk of overfitting, given the same number of training samples $T$.

Performing PCA over the whole set of wavelet coefficients would result in the same principal components as the global model, because the wavelet transform is a linear transform, and PCA essentially just rotates the data so that the coordinate axes align with the directions of greatest variations. Instead, this method performs PCA locally on each coefficient, which is a 3D vector quantity, over the database.

First, let us denote the mean of each wavelet coefficient over the database by
\begin{equation}
\bar{\textbf{s}}^k = \frac{1}{T} \sum_{i=1}^T \textbf{s}_i^k,
\end{equation}
where $k$ indexes the coefficients.

While we can perform statistical analysis on each $\textbf{s}^k$ independently of other values of $k$, we must consider their three components together. Each $\textbf{s}^k$ is a 3D vector representing either the scale (absolute value) or the detail (relative value) of the shape at a particular frequency and spatial location. However, the coordinate axes in general do not correspond to the directions of greatest variation in the database. Therefore, we perform PCA on each set of coefficient vectors, to obtain 3D vectors $\textbf{r}^k_i$ that represent the position along the directions of greatest variation, and $3\times3$ matrices $U^k$ that transform these coordinates to our original world coordinate system, as in
\begin{equation}
\label{eqn_coord_pca}
\textbf{s}^k_i = \bar{\textbf{s}}^k + U^k \textbf{r}^k_i 
\end{equation}
where we write $\textbf{s}^k = [x^k_s,y^k_s,z^k_s]^T$ and $\textbf{r}^k = [x^k_r,y^k_r,z^k_r]^T$ to denote the components of these vectors. Applying the transform $(U^k)^T$ to the data diagonalizes the covariance matrix, thus making each component independent.

The reconstruction of the face shape from the model is then given by the inverse wavelet transform
\begin{equation}
\label{eqn_wavelet_transform}
\textbf{F}(\textbf{s})_i = \sum_{o\in V(0)} \phi_0^o(i) \textbf{s}^o + \sum_{j=0}^{J-1} \sum_{l\in W(j)} \psi_j^l(i) \textbf{s}^l
\end{equation}
where $i$ is the vertex index in the reconstructed surface, $j$ is the level of wavelet coefficient, $J$ is the number of levels used, $V(0)$ is the set of scaling functions at level zero, $W(j)$ is the set of wavelet functions at level $j$, $o$ and $l$ are the coefficient indices, $\phi_0^o$ is the scaling function at the coarsest level centered on location $o$, and $\psi_j^l$ is the wavelet function at level $j$ and location $l$. While the transform is expressed here in terms of basis functions, it is computed using lifting operators, which amount to weighted averages of a vertex's local neighborhood, and it can be expressed as a matrix multiplication. The basis functions themselves, $\phi_j$ and $\psi_j$, are B-spline approximations to Gaussian and Mexican-hat functions. For more details, see Bertram et al.~\cite{bspline_subdiv_wavelets}.

We can now construct the combined basis $\Phi_W$ as follows. Observing that the inverse wavelet transform, Eq.~(\ref{eqn_wavelet_transform}), is a linear operator on the vector of concatenated wavelet coefficients $\textbf{s}$, we can write it as a $3n\times3n$ matrix $D^{-1}$. Thus, we have
\begin{equation}
\textbf{F}(\textbf{s}) = D^{-1}\textbf{s}
\end{equation}
and from Eq.~(\ref{eqn_coord_pca}) we have
\begin{equation}
\textbf{F}(\textbf{s}) = D^{-1}\bar{\textbf{s}} + D^{-1}U\textbf{s}
\end{equation}
where $\bar{\textbf{s}}$ is the concatenation of the coefficient means $\bar{\textbf{s}}^k$, and $U$ is a block-diagonal $3n\times3n$ matrix with the matrices $U^k$ on the diagonal. Therefore, because $D^{-1}\bar{\textbf{s}} = \overline{\textbf{F}}$, we have
\begin{equation}
\Phi_W = D^{-1}U
\end{equation}
as our combined basis, and the dimensionality of our shape space is $d=3n$. Note that $\Phi_W$ has full rank.

To use a spherical wavelet basis to represent shape, it must be a subdivision surface, in our case Catmull-Clark subdivision hierarchy. For training, the surfaces in the database are typically stored as triangle meshes without subdivision structure. Thus, we must resample the surfaces with the proper structure. The subdivision scheme uses quadrilateral elements, although it can handle extraordinary vertices. We resample the triangle meshes using the custom, yet straightforward, technique by Brunton et al.~\cite{Brunton2011} tailored to the fact that we are dealing with faces, which are topologically like a disc. In principle, however, any quad-remeshing technique can be used.

\abheading{Properties} The local model has the benefit that it avoids overfitting, and as a consequence we can keep all variability present in the training set. Intuitively, the \emph{local} surface properties of any given surface point are not likely to be specific to one set of faces or another. Whereas for the global model a bias in the training set, over-representation of one sex or a particular ethnicity or age range, can cause the lesser principal components to be highly specialized to that set, the geometry of a local surface patch is likely to be less dependent on the training data. The consequence is a somewhat unexpected behavior: by training and combining many low-dimensional models, which due to the limited flexibility of the training space ($\mathbb{R}^3$) have reduced sensitivity to bias in the training set, we get a final model with much greater flexibility, because truncation becomes unnecessary.

\begin{figure*}[htb]
\centering
\begin{tabular}{c c c c}
\includegraphics[height = 2.2cm]{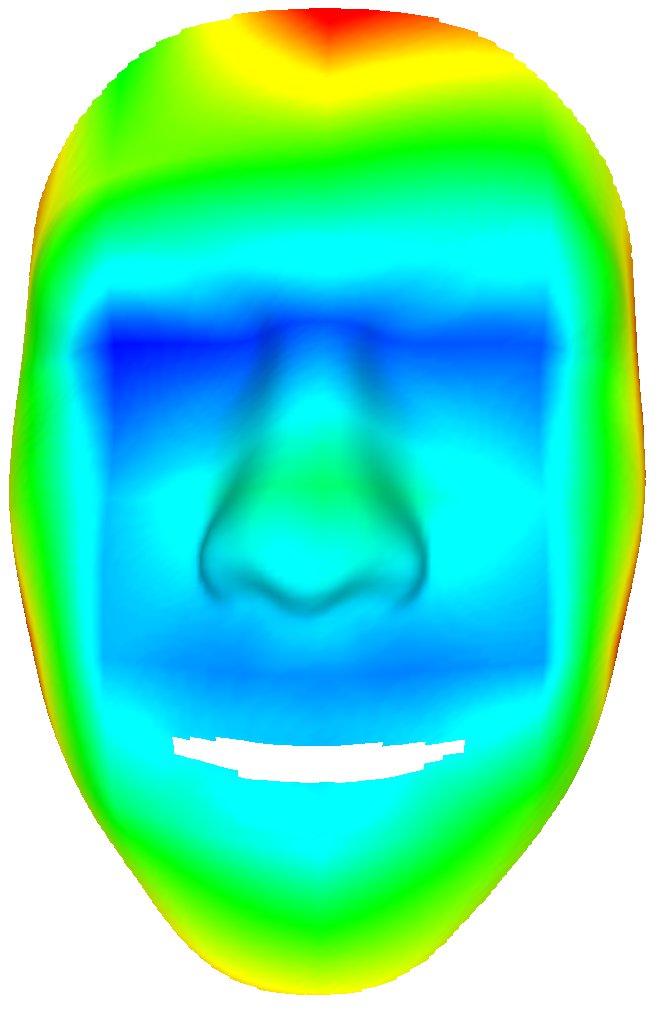}
\includegraphics[height = 2.1cm]{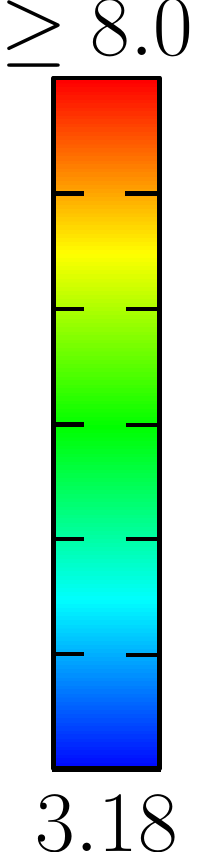} &
\includegraphics[height = 2.2cm]{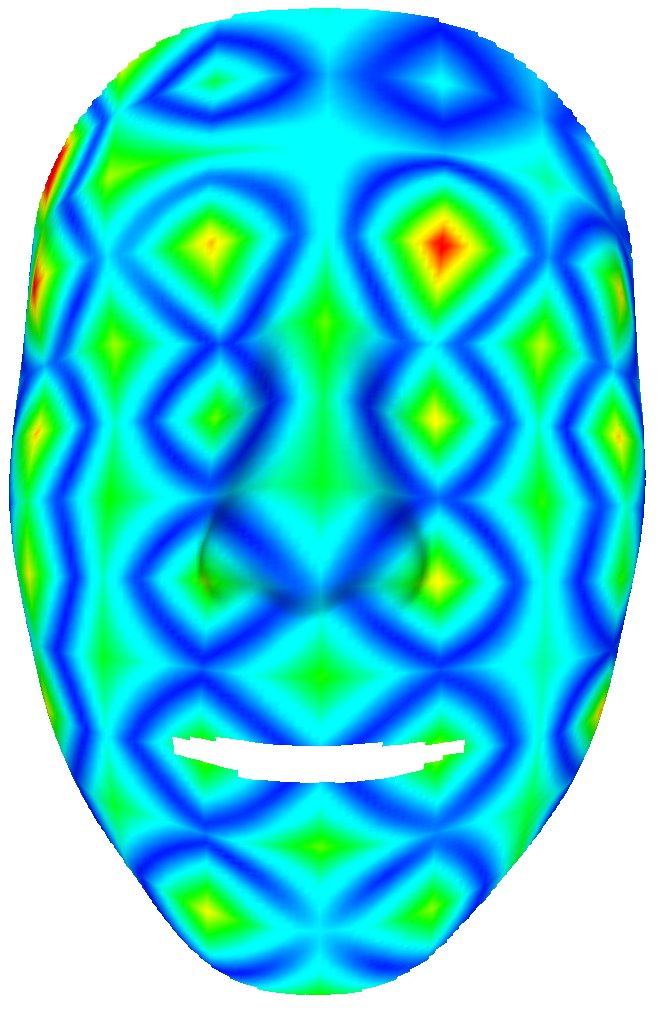}
\includegraphics[height = 2.1cm]{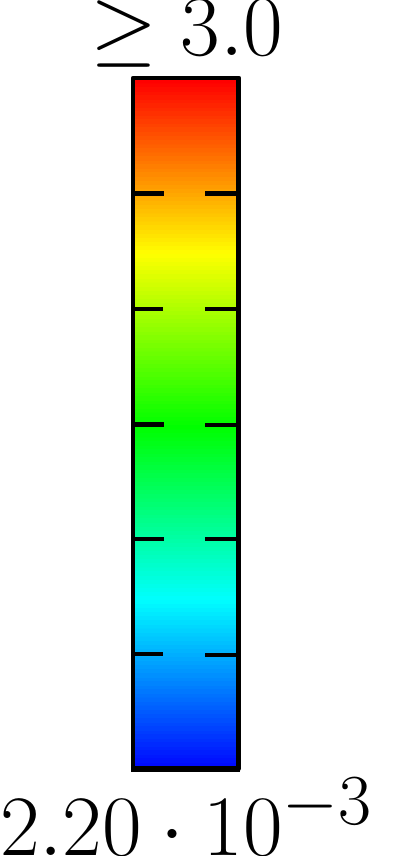} &
\includegraphics[height = 2.2cm]{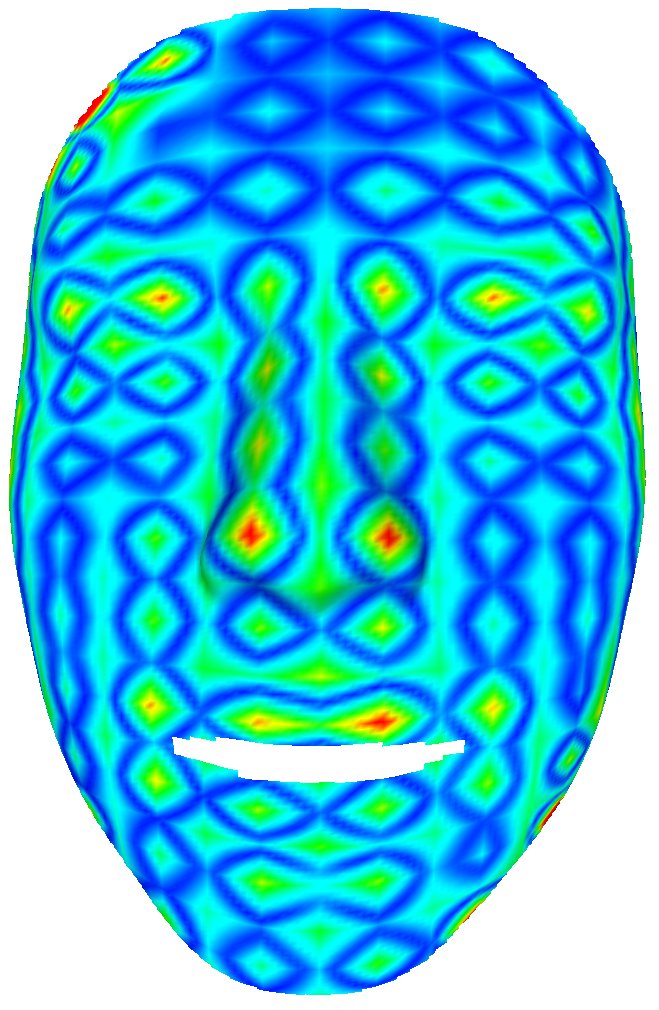}
\includegraphics[height = 2.1cm]{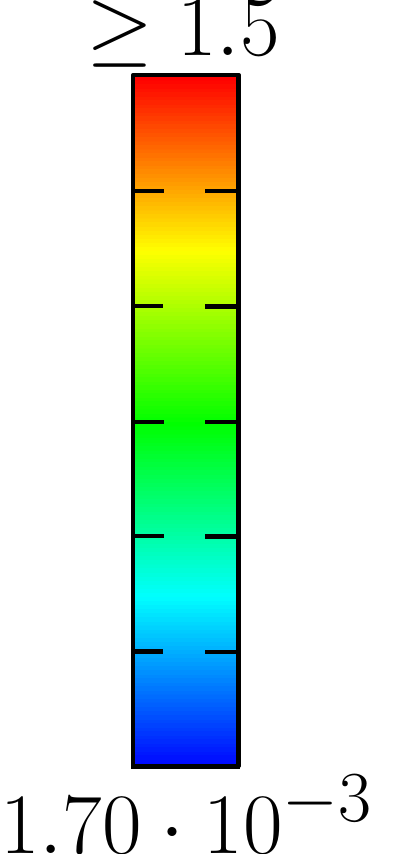} &
\includegraphics[height = 2.2cm]{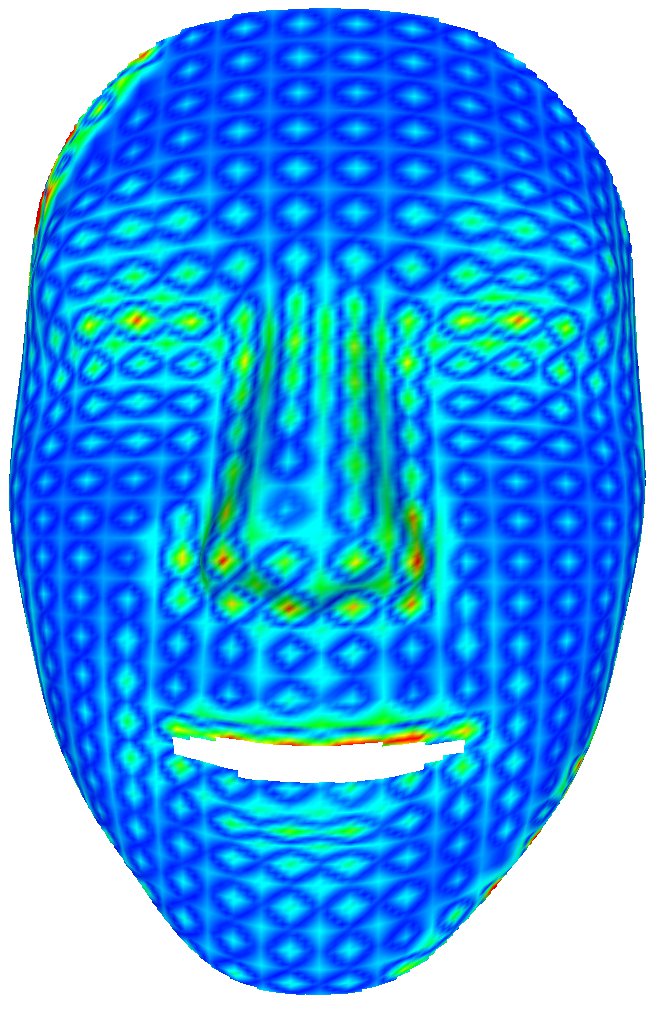}
\includegraphics[height = 2.1cm]{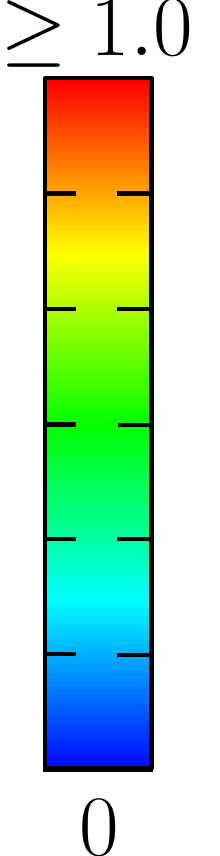} \\ %&
\small{level 0} & \small{level 1} & \small{level 2} & \small{level 3} %& \small{level 4} & \small{level 5} \\
\end{tabular}
\caption{\emph{Mean shape color-coded with the magnitude of shape variability for different levels. All units are in millimeters.}}
\label{fig_local_variability}
\end{figure*}

Figure~\ref{fig_local_variability} visualizes the mean shape color-coded with the magnitude of the shape variability for four levels of the wavelet subdivision, which corresponds to the localized shape variations at different scales. At finer scales, the variation quickly localizes around major facial features and reduces in magnitude.

The dimensionality of the local model is \emph{statistically} more favorable. If, as is usually the case, the number of vertices is much greater than the number of training examples, $n\gg T$, then the global model has problems of fitting to the particularities of the training set. In the local model, many independent statistical priors are learned, each with dimension 3. We have many more training examples than that. The independence of the local priors further allows an exhaustive search of the parameter space. Thus, we have no danger of getting trapped in local minima.

The drawback of these properties, in particular of retaining all the variability of the training data, is that the local model is a much higher-dimensional representation than the global model. Thus, the dimensionality of the local model can be \emph{computationally} much less favorable. There is, however, a trade-off that can be made by using the wavelet basis progressively, and working with the localized priors only up to a certain level.

As the lifting operations of the wavelet transform amount to local weighted averages of vertex coordinates, the transform can be expressed as a matrix multiplication, if the surface is expressed as a vector containing the vertex coordinates. Because the transform is biorthogonal, this matrix is square and has full rank. In contrast, the global PCA basis $\Phi_G$, has rank $d\leq\min(3n-1, T-1)$. As discussed in Section \ref{sec_model_training_global}, resampling the surface at linear combinations of vertex coordinates (eg., within a triangle), does not increase the rank of the transform. However, because $\Phi_W$ has full rank, we can obtain more detail by linearly upsampling the training surfaces. This means that we can resample the training shapes at high resolution to obtain a statistical model that captures fine shape detail.

The statistical measures generalization and specificity can be computed in the same manner as for the global PCA model. While there is no simple formula for the compactness of the wavelet model, if we retain all $3n$ shape parameters, and hence all variation, the compactness measure is fixed at $100\%$. In this case, the generalization measure is $0$.

%--------------------------------------------------------------------------------------------------------------------------------------------------------------------------------------------------------
\section{Fitting a Statistical Shape Model}
\label{sec_model_fitting}

\begin{figure*}
\centering
\includegraphics[width=13.0cm]{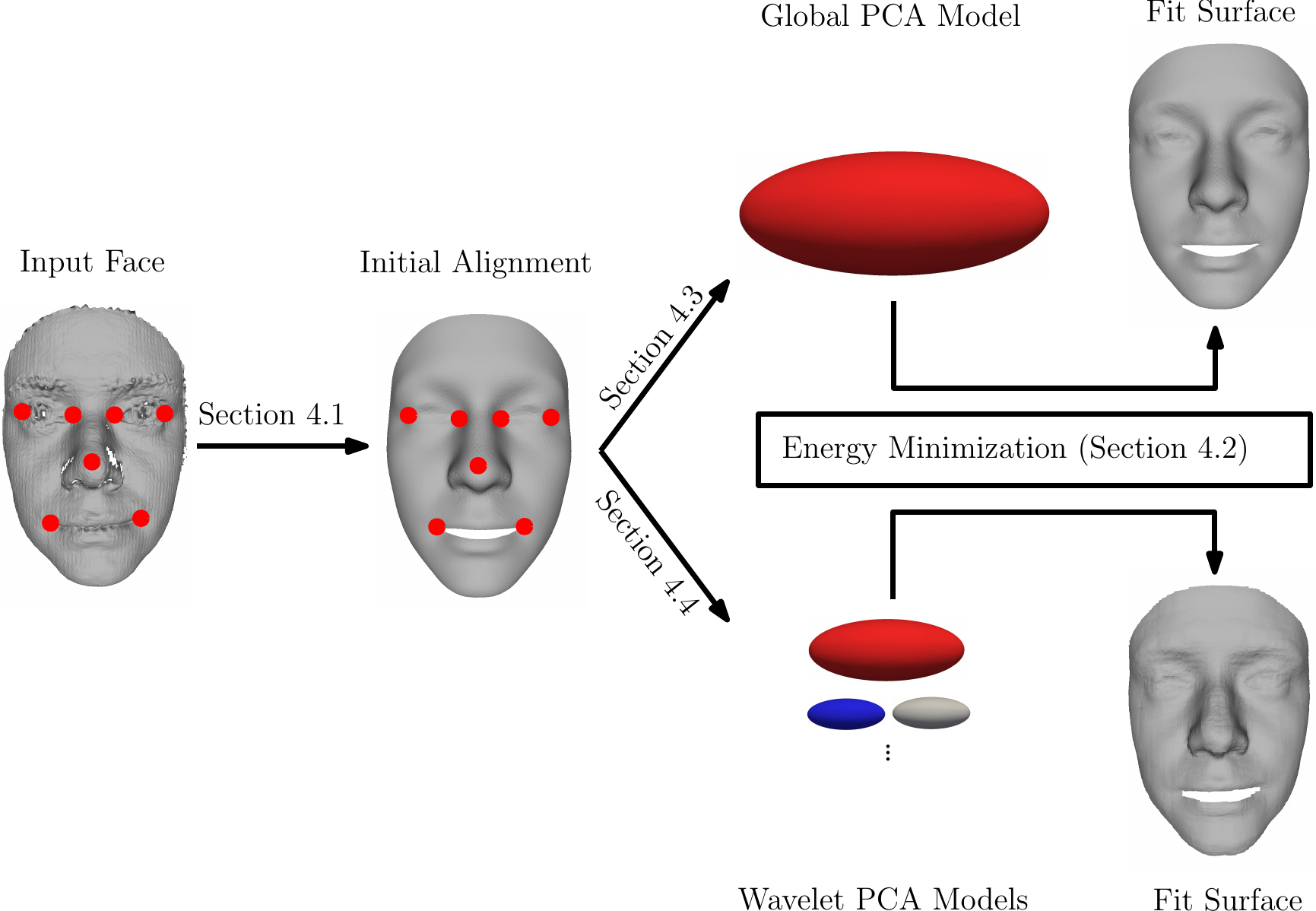}
\caption{\emph{Overview of the model fitting.}}
\label{fig_fitting}
\end{figure*}

In this section we give an overview of the generic model fitting approach shown in Figure~\ref{fig_fitting}. This assumes that a statistical shape model has been learned, for instance using one of the approaches given in Sections~\ref{sec_model_training_global} and~\ref{sec_model_training_local}. As can be seen from Figure \ref{fig_fitting}, the process begins with an initial alignment using either landmarks or feature points, described in Section \ref{sec_model_fitting_init_align}, regardless of the statistical shape space. Subsequently, we minimize an energy function with respect to the model parameters (Section \ref{sec_model_fitting_energy}), which are specific to the statistical shape space, and may affect the minimization strategy (Sections \ref{sec_model_fitting_global} and \ref{sec_model_fitting_local}).

\begin{figure}
\centering
\begin{tabular}{c c}
\includegraphics[width=2.25cm]{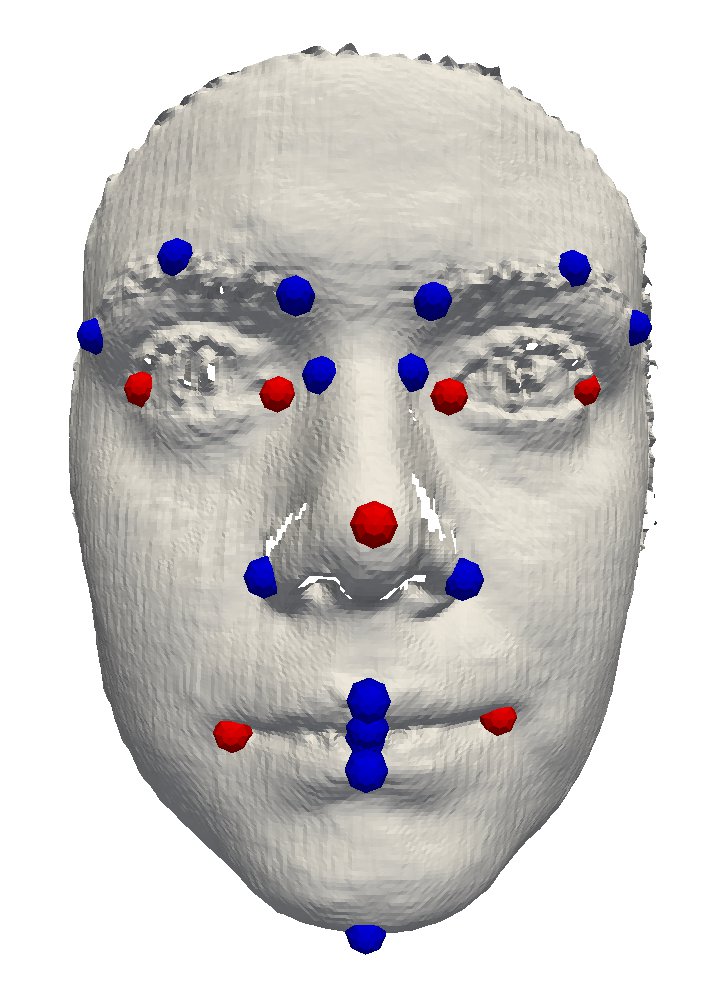} &
\includegraphics[width = 2.25cm]{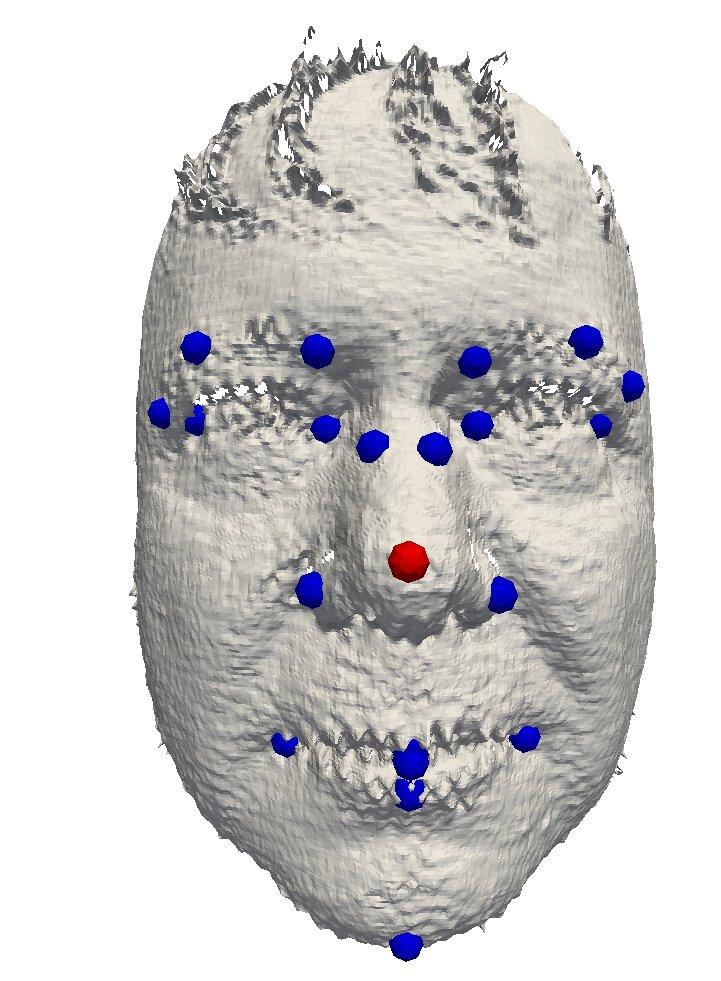} \includegraphics[width = 2.25cm]{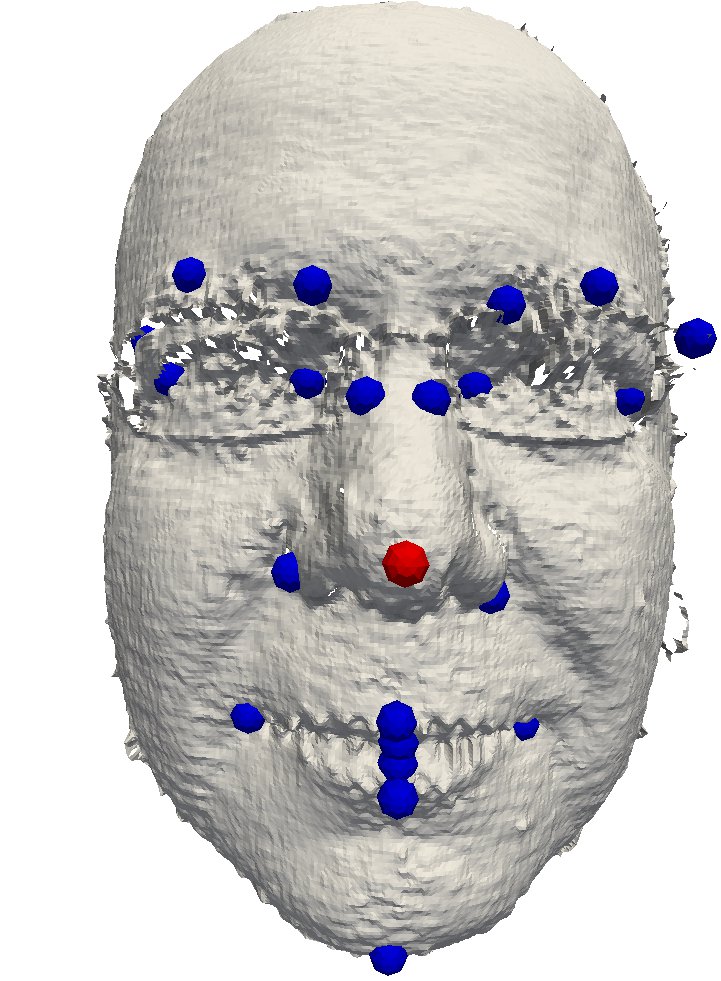} \\
(a) & (b)
\end{tabular}
\caption{\emph{Bosphorus scan with landmarks. (a) Red landmarks are used for initial alignment, blue landmarks are used for error evaluation. (b) Landmarks for two scans of the same identity. The position of the red landmark differs slightly for the two scans.}}
\label{fig_bosphorus_lnd}
\end{figure}

\subsection{Initial Alignment}
\label{sec_model_fitting_init_align}
To fit a statistical shape model to an input data set, we first need to align the input data and the statistical shape model to be in the same global coordinate system. Since we consider only shape differences in the training data, the initial alignment aims to find the rotation, translation, and uniform scaling that best aligns the statistical shape model with the input data.

To compute such an initial alignment, corresponding landmarks are commonly used. These landmarks can be manually located on the mean shape of the aligned training database once. On the input data, the landmarks can be predicted in a fully automatic way~\cite{creusot_landmark_labelling_3dor2010,salazar_auto_expr_face_reg_arxiv2012}. However, since we use a test database that contains a set of landmarks, we choose to use a subset of these landmarks (the ones shown in red in Figure~\ref{fig_bosphorus_lnd}a) to compute an initial alignment. This approach removes a potential source of fitting error due to landmark prediction inaccuracies.

Another commonly used way to rigidly align two shapes is to use automatically detected features. We test a method of this flavor in our experiments. The method we use proceeds by finding corresponding features on the mean face and the input scan using spin images~\cite{JohnsonHebert1997} and by performing random sample consensus~\cite{FischlerBolles1981}. This fully automatic method is expected to lead to less accurate alignments than the use of the given landmarks.

\subsection{Energy Minimization in Shape Space}
\label{sec_model_fitting_energy}
Our goal is to fit the statistical shape model to the input data as closely as possible while staying in the learned shape space. To fit our model to data, we minimize an energy function that amounts to the sum of squared distances between each model vertex and its nearest neighbor in the input point cloud. For our experiments, we use the following commonly used basic energy to pull the model towards the data
\begin{equation}
\label{eqn_nearest_neighbor}
E_{data}(\textbf{s}) = \sum_{i=1}^{n} \min\left( \left\| \textbf{f}_i - \textbf{p}_{\mbox{NN}(i)} \right\|_2^2, \tau \right)
\end{equation}
where $\textbf{f}_i$ is vertex $i$ of $\textbf{F}(\textbf{s})$ (see Eq.~(\ref{eq:generator}) and (\ref{eq:generator:basis})), $\textbf{p}\in P$ is a point in the input point cloud, $\mbox{NN}(i)$ returns the index of the nearest neighbor in $P$ of $\textbf{f}_i$, and $\tau$ is a truncation threshold to add robustness against outliers. We compute nearest neighbors with a k-d tree using the implementation in ANN~\cite{ANN}.

When fitting a statistical shape model to data, the space of possible solutions should only contain likely shapes, thus ensuring that only plausible results are possible. A common and intuitive approach is to use the (negative logarithm of the) learned prior distribution as an energy term. In the case of PCA, it is common to assume a multi-dimensional Gaussian centered on the mean shape. In terms of energy minimization, this amounts to placing a soft constraint that the solution should be close to the mean. By design, however, this introduces a bias into the optimization, and results using this technique tend to lose distinctiveness and look similar to the mean. 

\begin{figure}
\centering
\includegraphics[width=8.0cm]{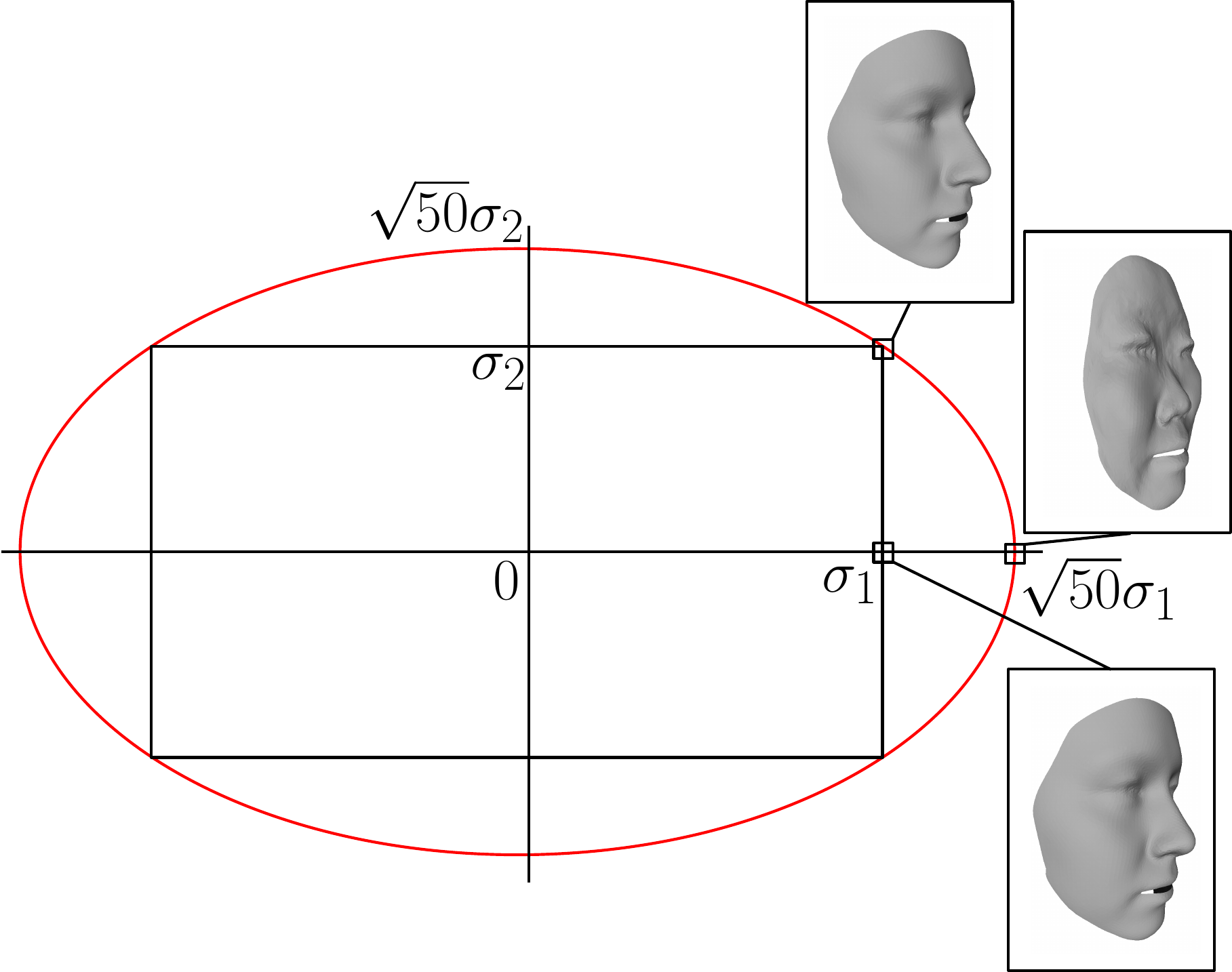}
\caption{\emph{Unrealistic shapes may occur for large numbers of dimensions $d$ along the directions of the principal components when maintaining a fixed Mahalanobis distance from the mean shape. Here, a global statistical shape space with $d=50$ and $c=1$ is shown.}}
\label{fig_prior_plot}
\end{figure}

Patel and Smith~\cite{patel_explore_id_manifold_eccv2010} proposed an alternative prior that is aimed at maintaining the distinctiveness of the models. They model the shape space as a manifold that is at a constant Mahalanobis distance from the mean. This is based on the observation that the squared Mahalanobis distances from the mean of a set of $d$-dimensional normally distributed vectors form a $\chi_d^2$ distribution with expected value equal to $d$. Hence, Patel and Smith restrict the shapes to be on the hyper-ellipsoid at Mahalanobis distance $\sqrt{d}$ from the mean in order to preserve shape distinctiveness. While this approach models distinctiveness using the expected Mahalanobis distance from the mean, it does not consider the normal distributions along each dimension of the shape space. That is, the modeled shape space may contain highly unlikely shapes along the directions of the principal components, as can be seen for the shape at the intersection of the hyper-ellipsoid shown in red and the $x$-axis in Figure~\ref{fig_prior_plot}. 

To simultaneously avoid mean-shape bias and highly unlikely shapes, in our experiments we constrain the shape to lie within a region of the shape space where the prior probability is sufficiently high, i.e. the shape is sufficiently likely. Ideally, in the case of PCA models, one would impose ellipsoidal constraints corresponding to a probability isosurface of a given value. However, from an optimization standpoint, it is much easier and more efficient to impose linear constraints. Hence, we constrain the shape to lie within the hyper-box of $\pm c \sigma_i$ about the mean shape, where $\sigma_i$ is the standard deviation of the training data along dimension $i$ of the learned statistical space, and $c$ is a parameter controlling the amount of deviation allowed. Figure~\ref{fig_prior_plot} shows a two-dimensional plot of this hyper-box. We demonstrate that in this shape space, by minimizing an energy function with only the data term given in Eq.~(\ref{eqn_nearest_neighbor}), we can maintain distinctiveness, while avoiding unrealistic shapes. This is equivalent to a prior probability of the form
\begin{equation}
P(\textbf{s}) = \prod_{i=1}^d P_i(s_i)
\end{equation}
where
\begin{equation}
\label{eqn_hyperbox}
P_i(s_i) = 
\begin{cases}
1 & |s_i| \leq c\sigma_i \\
0 & \mbox{otherwise} 
\end{cases}
\end{equation}
if we assume the shape parameters $s_i$ are centered (mean subtracted). We call this a \emph{hyper-box prior}.

A common post-process to fitting the parameters of the statistical models, is to then leave the statistical shape space and perform a fine-fitting of the vertex positions directly, similar to a template fitting method. We deliberately do \emph{not} do this for two reasons. First, such a step is most often necessary when the learned shape space is not sufficiently generalizable to express novel shapes. This can occur due to insufficient or poor training data. However, as detailed in Section \ref{sec_comp_eval}, we train from clean data containing a good sampling of both sexes and different ethnicities with a high-quality registration. Thus, our learned models are of high-quality. 

Second, we wish to study the properties of the statistical shape spaces themselves, and leaving the space in a post-process would inevitably introduce additional uncertainty in analyzing the fitting results in Section \ref{sec_comp_eval}. This is particularly true when we evaluate the fitting in the presence of occlusions.

\subsection{Global PCA}
\label{sec_model_fitting_global}

\abheading{Energy Minimization} While the energy in Eq.~(\ref{eqn_nearest_neighbor}) is not strictly differentiable at all points, it is continuous, and the number of points where it is not differentiable is small. Hence, we can minimize it using a bounded Quasi-Newton method~\cite{liu_nocedal_lbfgsb}. The coordinate bounds on the parameters enforce the condition given in Eq.~(\ref{eqn_hyperbox}). This minimizer gives super-linear convergence rates without the need for an explicit inverse Hessian. Note that this optimization technique does not guarantee to find the global optimum of the energy function.

\abheading{Computational Complexity} Let $t_G$ denote the number of iterations required for the minimization to converge, and let $m$ denote the number of data points in the target shape. The complexity of building a k-d tree of $m$ points in $3D$ is $O(m\log{m})$, and a single nearest neighbor search takes $O(m^{2/3})$ time~\cite{lee_kdquadtree_1977}. Given the nearest neighbor indices for all $n$ points, a single evaluation of the energy given in Eq.~(\ref{eqn_nearest_neighbor}) takes $O(n)$ time, and a single evaluation of its gradient takes $O(nd)$ time. Thus, the overall time complexity of fitting the global PCA model to a data set is $O(m\log{m} + t_G n(d+m^{2/3}))$. For this model, each training shape contains $n=5996$ vertices.

\subsection{Wavelet PCA}
\label{sec_model_fitting_local}

\abheading{Energy Minimization} We minimize the energy in Eq.~(\ref{eqn_nearest_neighbor}) using a global search of each parameter. That is, we sample uniformly within the range given by the hyper-box prior given in Eq.~(\ref{eqn_hyperbox}) for each component of $\textbf{r}^k$ for each $k$ sequentially, starting with the coarsest resolution coefficients and progressively increasing the resolution. For each sampled value, we reconstruct the surface using Eq.~(\ref{eqn_coord_pca}) and (\ref{eqn_wavelet_transform}), and evaluate the nearest neighbor energy (Eq.~(\ref{eqn_nearest_neighbor})). The parameter value that minimizes this energy is then taken as the estimate for this parameter.

\abheading{Computational Complexity} Since we use a sampling approach to minimize the energy for the local model, the complexity depends not on the number of iterations of a nonlinear optimization, but on the number of samples $t_L$ per parameter. The number of wavelet coefficients is equal to the number of vertices in the subdivision mesh, which we denote by $n$. Note, however, that $n$ increases as the number of subdivision levels $J$ increases. In our experiments, we resample all training shapes with $n=24897$ vertices. The complexity of the nearest neighbor search remains unchanged, as does the cost of evaluating the energy. However, the energy must be evaluated $t_L$ times for each of the $n$ coefficients. Hence, the overall complexity is $O(m\log{m} + n(m^{2/3} + t_L n))$, which is dominated by the $O(n^2 t_L)$ part. Thus, assuming $n\gg t_G$, we expect the local model, with its higher-dimensional representation to take much longer to fit to the same point cloud. However, a trade-off between detail and running time can be made by fitting coefficients only up to some level less than the number of levels in the wavelet decomposition.

%--------------------------------------------------------------------------------------------------------------------------------------------------------------------------------------------------------
\section{Comparative Evaluation}
\label{sec_comp_eval}

In this section we evaluate both the fitting speed and quality of the global and local models. We evaluate both models for different initial alignment strategies and for different types of severe occlusion. Furthermore, we evaluate the models qualitatively for noisy scan data acquired using affordable stereo and range camera setups.

\subsection{Experimental Setup} 
\label{sec_exper_setup}

\abheading{Training Data}
For training, we use the neutral expressions of $T=100$ subjects from the BU-3DFE database~\cite{bu-3dfe}. This database contains relatively clean surfaces without occlusions, and a typical cropped face contains about 7500 vertices. Furthermore, each cropped face is equipped with 83 landmark points. 

\abheading{Parameterization} We parametrize the database using the method of Salazar et al.~\cite{salazar_auto_expr_face_reg_arxiv2012} that deforms a template to each input face. This method is capable of predicting landmark points to aid in the template fitting. However, since we are given manually placed landmarks, our algorithm uses these instead of predicted ones. This removes a potential source of error during registration. The template we use contains 5996 vertices. We choose this low-resolution template for parametrization since the database has low resolution and does not contain small shape details. While the BU-3DFE database contains six additional expressions in four different levels, we consider only neutral expressions for our comparison. The resulting registration is of high-quality, which has been verified by manually inspecting each registered face.

\abheading{Test Data} We use a subset of 20 subjects (10 female and 10 male) of the Bosphorus database~\cite{bosphorus} to test our algorithm. Each subject is present in five occlusion levels: without occlusion, with glasses, with an occlusion of one eye by a hand, with an occlusion of the mouth by a hand, and with an occlusion of parts of the face by hair. Examples for each occlusion class can be seen in the left column of Figure~\ref{fig:someResults}. We chose this database as it allows the evaluation of different methods in the presence of severe occlusion. The resolution of this database is fairly high, and a typical face contains about 35000 vertices. 

Each face is annotated with up to 22 landmarks. Figure~\ref{fig_bosphorus_lnd}a shows a model of the Bosphorus database with 22 landmark positions. The landmarks shown in red are used to compute an initial alignment of the test face to the learned shape space and the landmarks shown in green are used for error evaluation. Not all of the landmarks may be present in the database for two reasons: first, landmarks may be missing due to occlusion, and second, some landmarks are placed erroneously and we manually removed these landmarks.
The landmarks that are present are not perfectly located, as shown in Figure~\ref{fig_bosphorus_lnd}b. Here, the location of the landmark at the nose tip is slightly shifted for two models of the same identity. We observed that in our test set the location of the landmarks usually varies by as much as $1cm$ across different scans.

\abheading{Implementation Details} In all experiments, we set the nearest neighbor distance truncation threshold $\tau$ to $10mm$, and we set the parameter $c$ controlling the size of the hyper-box prior to $1.0$. For the global model, we use 30 principal components. For the local model, we use a base grid of size $5\times 7$ and $J=6$ levels of subdivision.

\begin{figure*}[htb]
\centering
\begin{tabular}{ccccccc}
\includegraphics[height = 2.5cm]{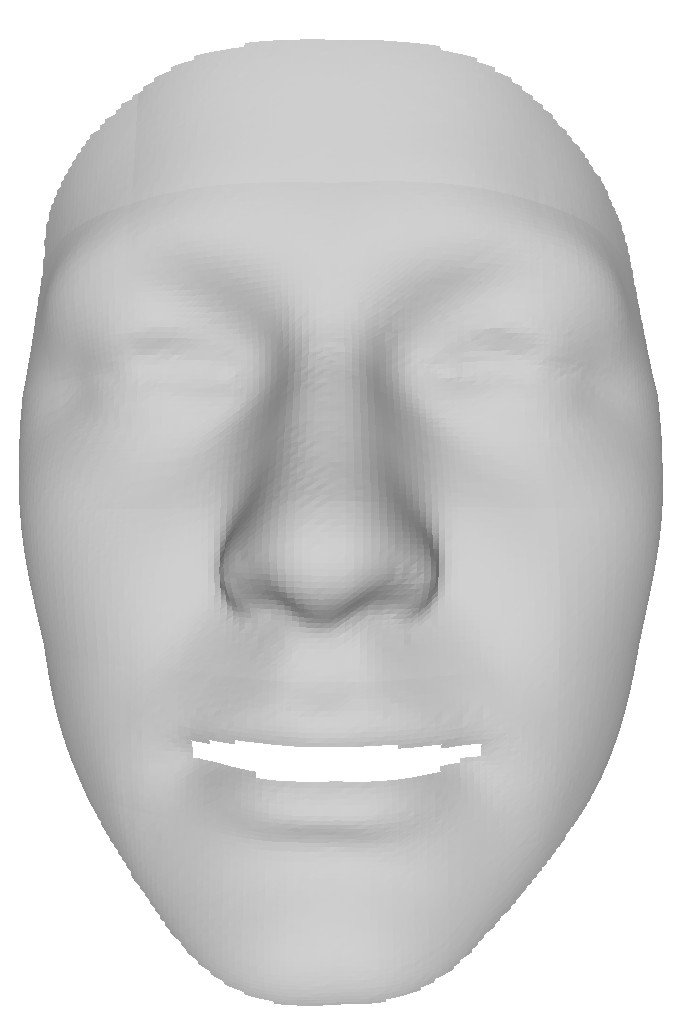} &
\includegraphics[height = 2.5cm]{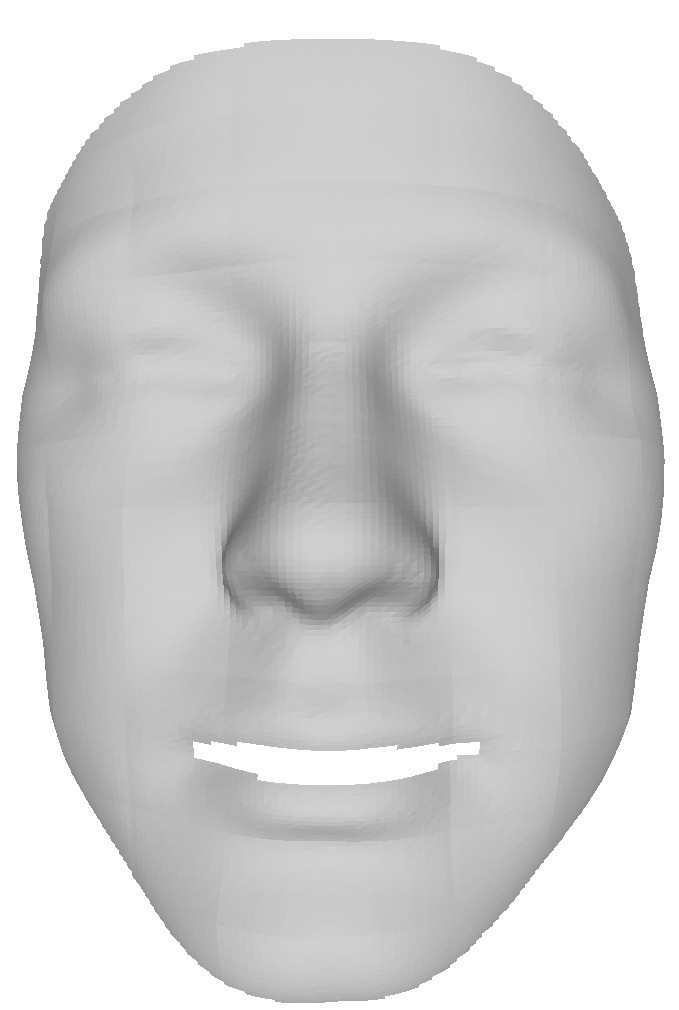} &
\includegraphics[height = 2.5cm]{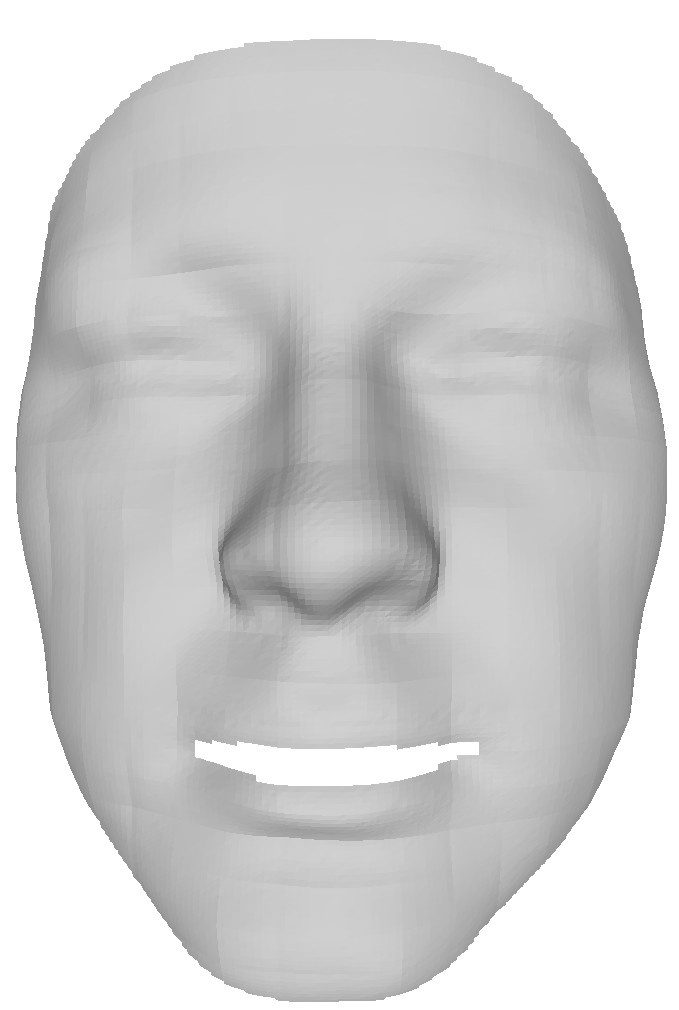} &
\includegraphics[height = 2.5cm]{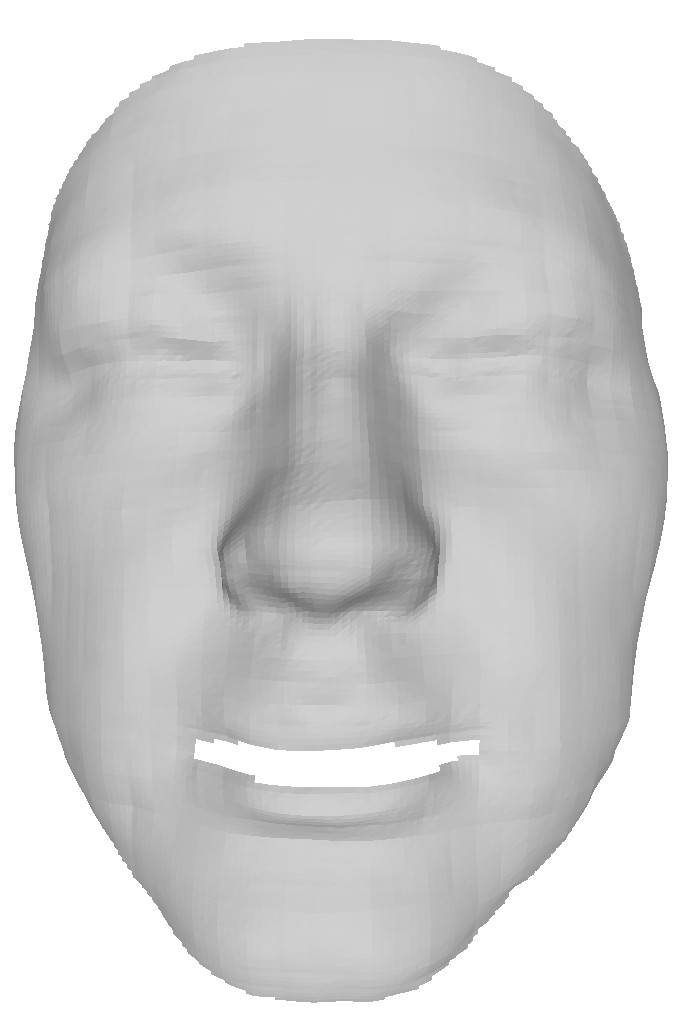} &
\includegraphics[height = 2.5cm]{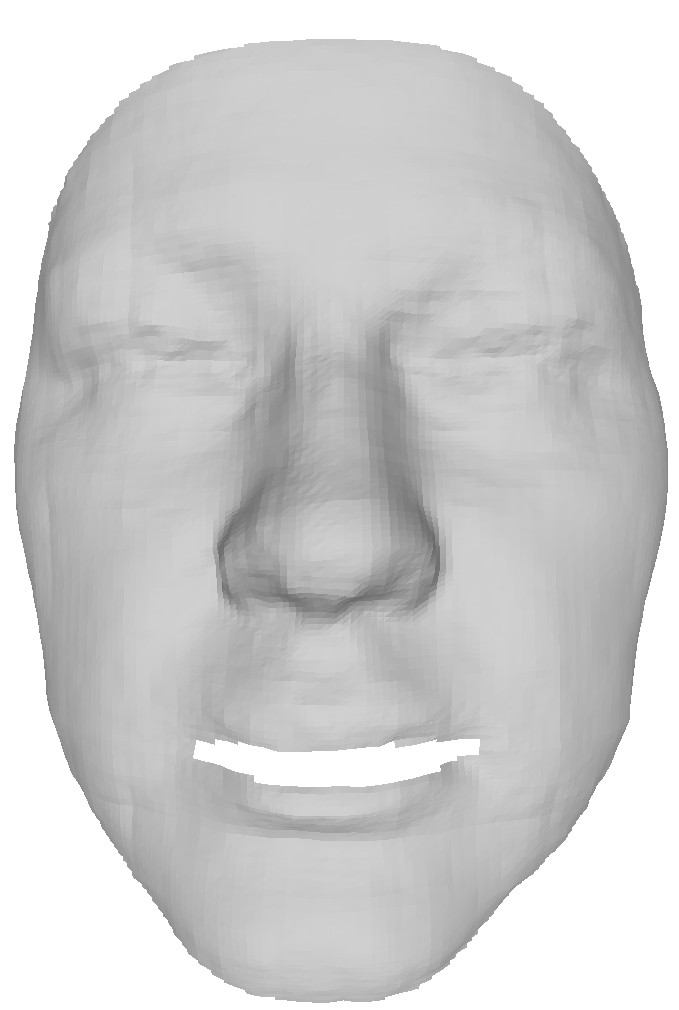} &
\includegraphics[height = 2.5cm]{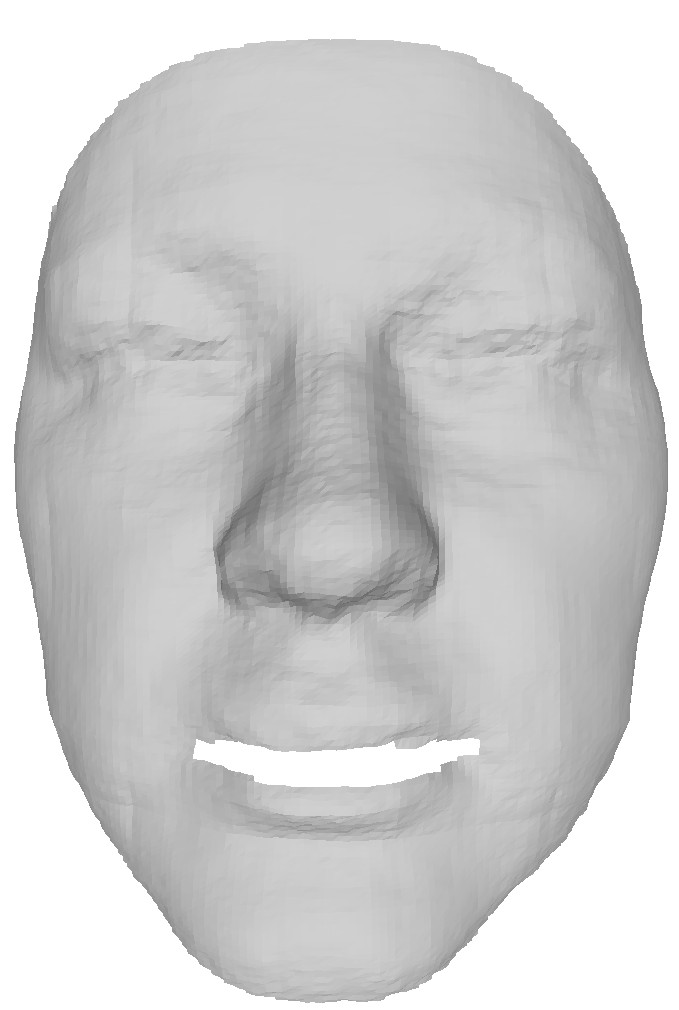} &
\includegraphics[height = 2.5cm]{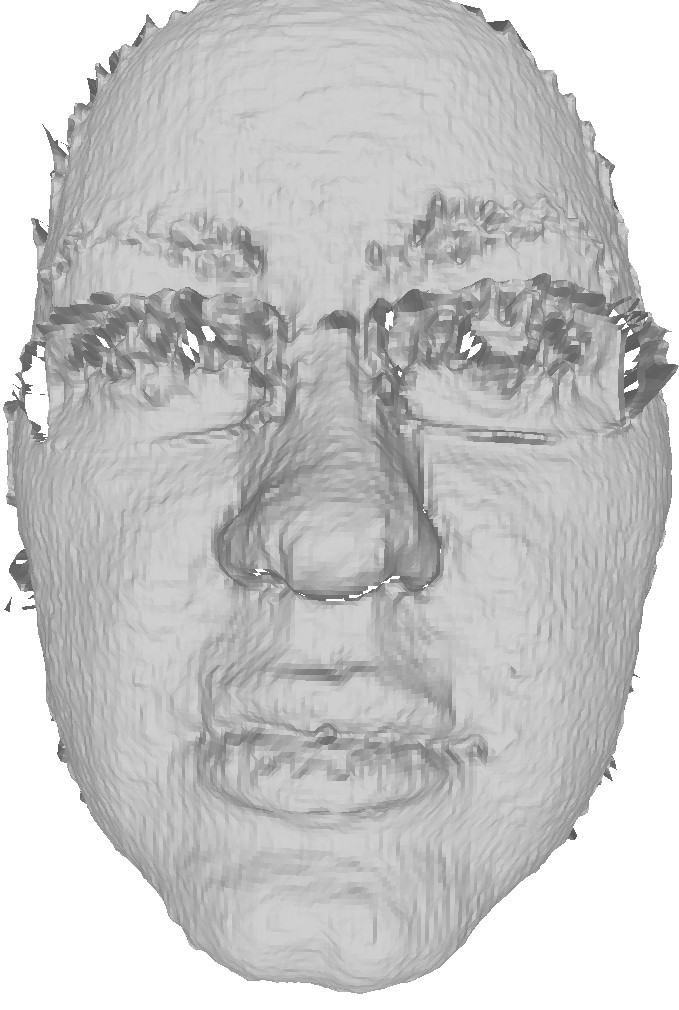} \\
\small{level 0} & \small{level 1} & \small{level 2} & \small{level 3} & \small{level 4} & \small{level 5} & \small{input data}\\
\end{tabular}
\caption{\emph{Reconstruction of a model from the Bosphorus database using different levels of the local shape space.}}
\label{fig_multiple_local_levels}
\end{figure*}

\abheading{Timing} The global model can be fitted in under a minute per face for the data we are using. For the wavelet model the more levels that are used, the more accurate the reconstruction becomes, but also the more time-consuming. While the reconstruction up to level zero runs in a few seconds, the reconstruction up to level five runs in slightly over one hour. Figure~\ref{fig_multiple_local_levels} shows the reconstruction of a Bosphorus scan when optimizing the shape coefficients of the local shape space up to different levels. This gives a way to trade off computation time and reconstruction accuracy. For all the experiments in the following, we evaluate the accurate reconstructions up to level 5.

\subsection{Error Measures}
\label{sec_comp_eval_err_measure}

In addition to the statistical measures discussed in Section~\ref{sec_model_training_eval} to evaluate the shape space itself, we use three ways to evaluate the \emph{fitting} results for different initialization strategies and under different types of occlusion. The first two evaluation measures are quantitative measures, while the third measure is qualitative.

\begin{figure*}
\centering
\includegraphics[width=4.5cm]{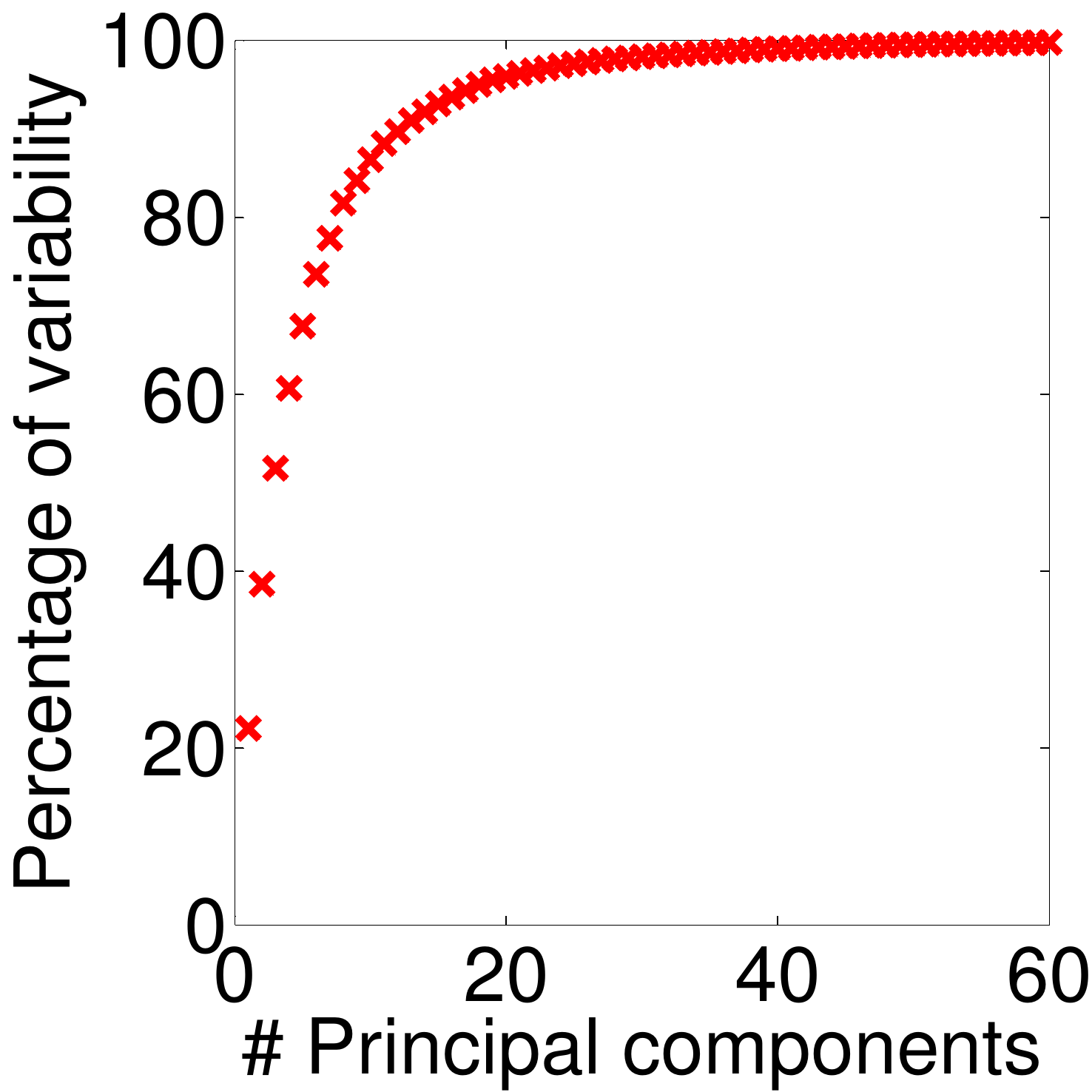}	
\includegraphics[width=5.0cm]{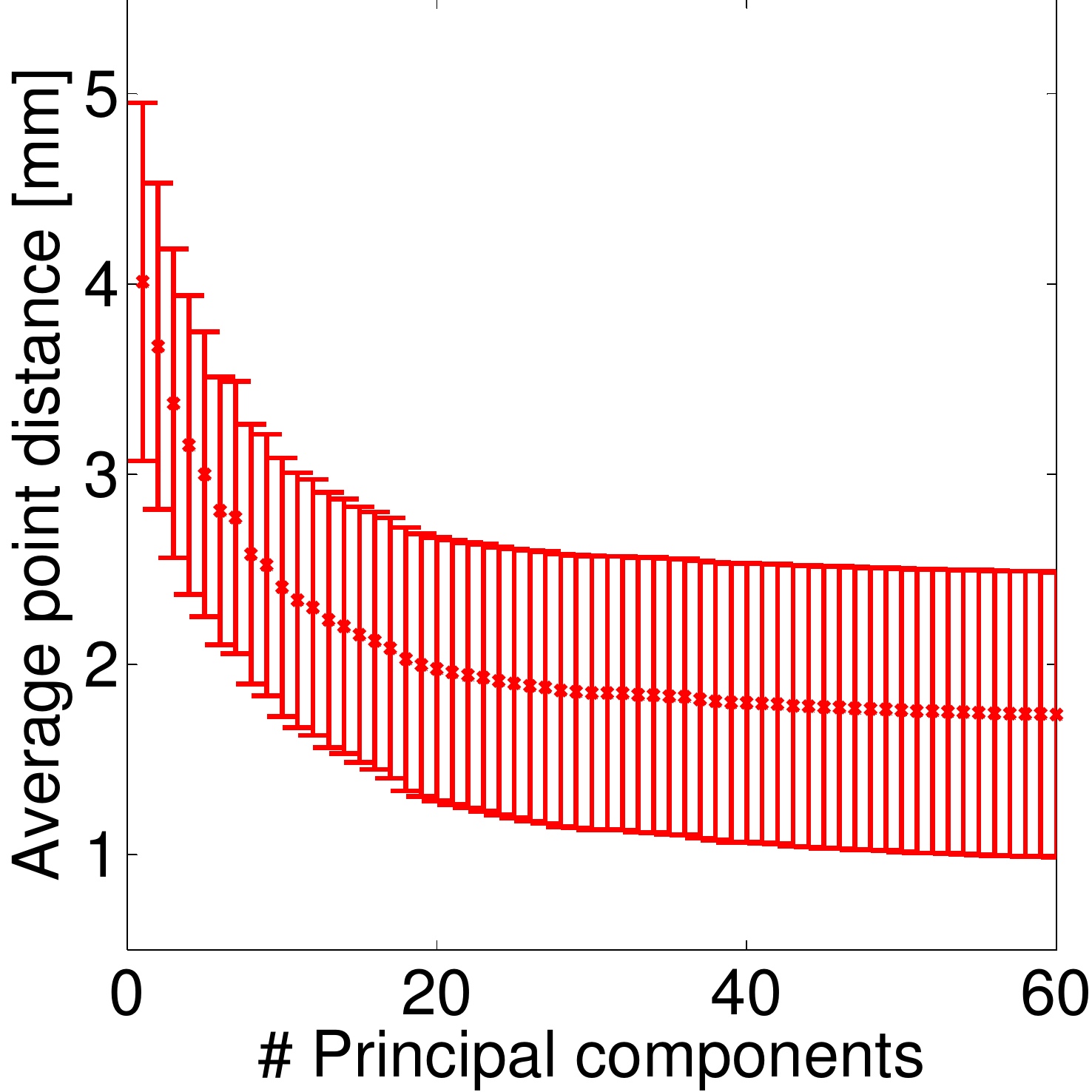}	
\includegraphics[width=5.0cm]{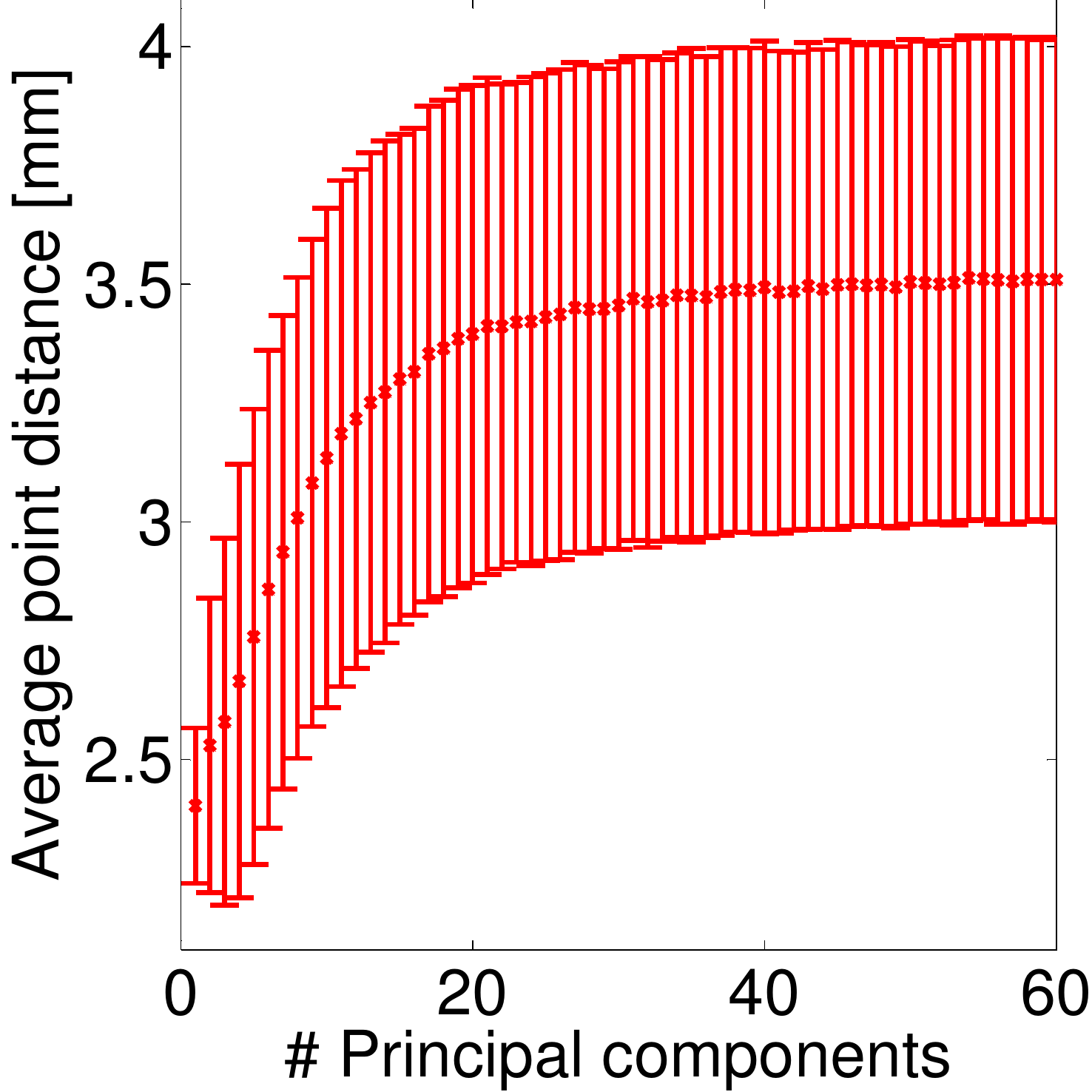}	
\caption{\emph{Compactness, generalization and specificity for the global PCA model.}}
\label{fig_pca_error_measures}
\end{figure*}

\abheading{Statistical Measures} A shape space should ideally be compact, general, and specific. Figure~\ref{fig_pca_error_measures} shows that for the global PCA model, $30$ principal components explain more than $98\%$ of the data variability. Furthermore, for more than $30$ components, the generalization error only decreases slightly, which implies that the benefit of choosing more components is small. Finally, the specificity error still increases for more than $30$ components, which means the model represents plausible faces. Hence, we choose dimensionality $d=30$ for the global model. In the last column of Figure~\ref{fig_pca_error_measures}, $10000$ random samples are chosen and the mean and standard deviation over all samples are shown.

For the Wavelet model, compactness and generalization are predetermined by the fact that we retain all variability in the model. For compactness, the percentage of variability explained by the model is $100\%$. For generalization, the error will be $0\mbox{mm}$. For specificity, we measured the mean point distance to be $3.87\mbox{mm}$ with a standard deviation of $1.12\mbox{mm}$. As expected, this is slightly higher than the value for the global model, since the local wavelet model is less specific to the training data.

\begin{figure}
\centering
\includegraphics[width=3.0cm]{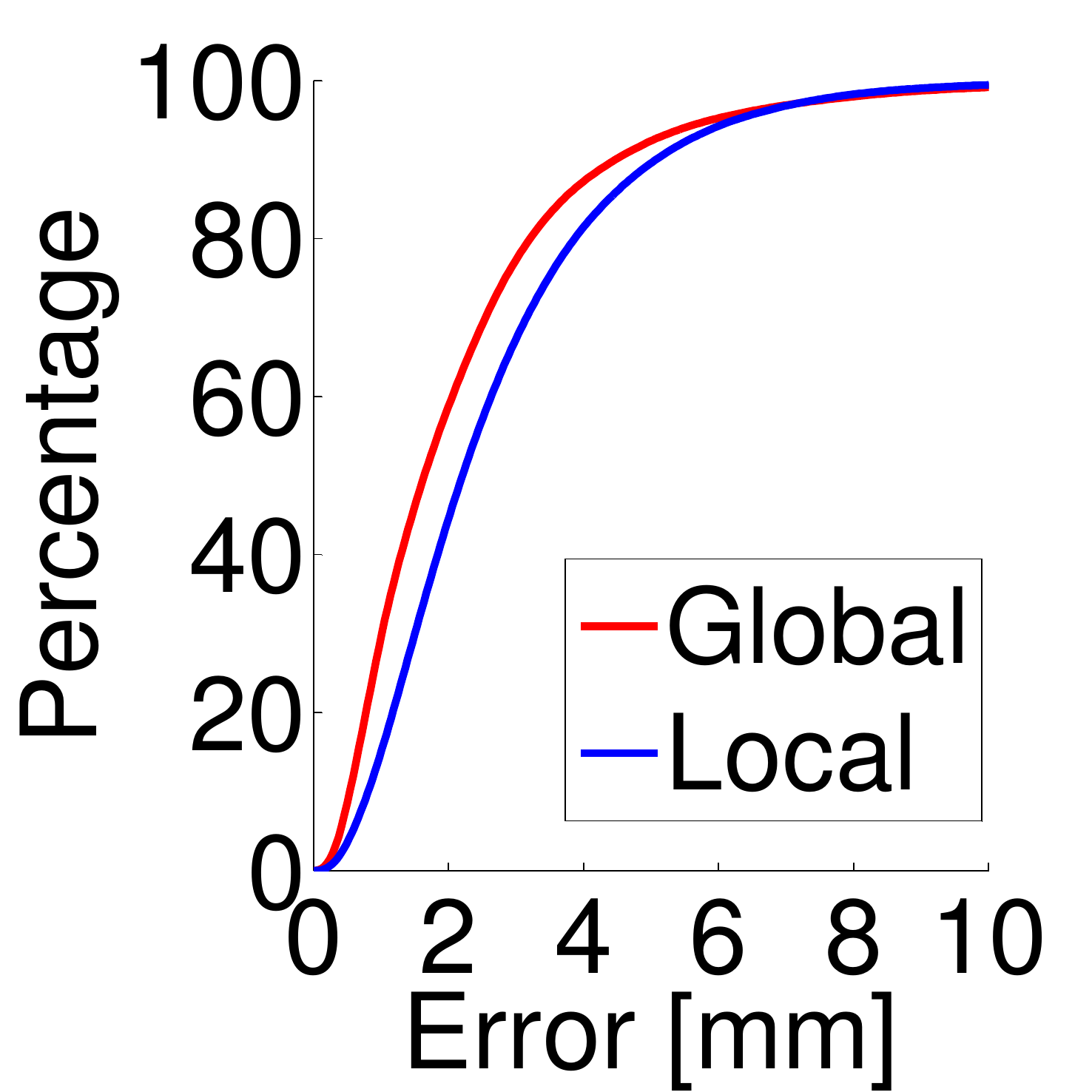}
\caption{\emph{Cumulative fitting error of 10-fold cross valiation computed on the training data.}}
\label{fig_cumulative_10foldcv_error}
\end{figure}

\abheading{Fitting 10-fold Cross Validation} We perform a $10$-fold cross valiation on the training data, to evaluate the generalization of the model fitting for the global and local models. We first randomly split the training data into $10$ groups, where each group consists of half male and half female subjects. We then learn a statistical shape model on $9$ groups and fit it to each face of the remaining group. Figure~\ref{fig_cumulative_10foldcv_error} shows the error for both models, measured by the distance between each vertex of the fitting result and its corresponding vertex on the input face. Both models fit the data well, where $80\%$ of the vertices are with error $\leq 3.23\mbox{mm}$ for the global model and $\leq 3.88\mbox{mm}$ for the local model. The error for the global method is slightly smaller than for the local method. 

\abheading{Landmark Distance} We evaluate the fitting quality using a subset of the landmarks (the ones shown in blue in Figure~\ref{fig_bosphorus_lnd}a) located on the input data. Note that these landmarks are not used in the initial alignment. Also recall that not all of these landmarks are present in all target scans. We only evaluate the error for those landmarks present in a given scan. The landmarks that are present in the test data are considered the ground truth landmark locations. We manually placed all of the landmarks used for testing on the mean shape of the aligned training data. The position of these landmarks after fitting are the estimated landmark positions. The distance between these estimates and the available ground truth landmark positions is the error we measure for the test data. This evaluation is commonly considered an accurate form of error measurement. However, unfortunately, it is possible only for a small subset of surface points since we require ground truth landmarks for this test. In the following, we give a less accurate, yet more dense evaluation, using a surface distance.

\abheading{Surface Distance} We evaluate the distance between the input scan and the fitted model over the entire surface by computing the distance to the nearest neighbor on the input data for each point on the fitted model. This gives a lower bound on the fitting error in terms of semantically meaningful correspondences, but it can be computed for the entire surface. 

\abheading{Visual Qualitative Evaluation} Finally, we evaluate the results visually by showing the distance from the fitted model to the input scan.

\subsection{Influence of Initialization} 
\label{sec_comp_eval_initialize}

We first evaluate the influence of the two different initialization strategies discussed in Section~\ref{sec_model_fitting_init_align} on the results. Tables~\ref{tab_landmark_distance_spin_image_alignment} and~\ref{tab_surface_distance_spin_image_alignment} give the error statistics of the landmark distance and the surface distance for models without occlusions. Note that the two initialization strategies yield results of similar quality for our test database for both statistical models. This implies that both models are robust with respect to changes in the initialization.

We observed a similar behavior of the two initialization strategies for all occlusion levels with the exception that for two models with severe hair occlusion the initialization using Spin images fails, which leads to poor results.

\begin{table}[ht]
\centering
{\small%footnotesize
\begin{tabular}{l | c | c | c | c }
Initialization & Mean & Median & Std. Dev. & Max \\
\hline
\multicolumn{5}{c}{Global Model} \\
\hline
manual landmarks & 5.74	&	4.75	&	4.15	&	26.39	\\
Spin image alignment &  5.51 & 4.50	& 3.90 	& 25.39 	\\
\hline
\multicolumn{5}{c}{Local Model} \\
\hline
manual landmarks & 5.96	& 5.09 & 4.29 & 30.07 \\
Spin image alignment & 5.64 & 4.50	& 4.00 	& 25.27 	\\
\end{tabular}
}
\caption{\emph{Landmark distance statistics (in mm) for different initialization strategies.}}
\label{tab_landmark_distance_spin_image_alignment}
\end{table}

\begin{table}[ht]
\centering
{\small%footnotesize
\begin{tabular}{l | c | c | c | c }
Initialization & Mean & Median & Std. Dev. & Max \\
\hline
\multicolumn{5}{c}{Global Model} \\
\hline
manual landmarks & 1.06	&  0.81	&  0.91	&  14.00	\\
Spin image alignment &  0.91 & 0.74	& 0.64 & 9.18 	\\
\hline
\multicolumn{5}{c}{Local Model} \\
\hline
manual landmarks & 0.80 & 0.47 & 1.06 & 15.63 \\
Spin image alignment & 0.63 & 0.43	& 0.72 & 16.51 	\\
\end{tabular}
}
\caption{\emph{Surface distance statistics (in mm) for different initialization strategies.}}
\label{tab_surface_distance_spin_image_alignment}
\end{table}

\subsection{Influence of Occlusion} 

We now evaluate the methods in the presence of severe occlusion. For these results, we initialize using the given landmarks to remove a potential source of error. Table \ref{tab_landmark_distance} shows the statistics of the landmark distances. Recall that both methods are evaluated against ground truth landmarks that may be shifted by up to $1cm$, as shown in Figure \ref{fig_bosphorus_lnd}. In light of this, the quality of the reconstruction from the two models, as measured by landmark distance, is not distinguishable. 

\begin{table}[ht]
\centering
{\small%footnotesize
\begin{tabular}{c | c | c | c | c | c }
Model & Occlusion & Mean & Median & Std. Dev. & Max \\
\hline
\multirow{2}{*}[-0.5cm]{global}	& none		&  5.74	&	4.75	&	4.15	&	26.39	\\
																& glasses	&  6.68	&	5.64	& 4.71	&	31.05 \\
																& eye			&  6.87	&	5.29	& 5.29	&	35.74	\\
																& mouth		&  8.83	&	6.65	&	6.73	&	38.48	\\
																& hair		&  6.88	&	5.43	&	5.07	&	35.08	\\
\hline
\multirow{2}{*}[-0.5cm]{local}	& none		&  5.96	& 5.09 & 4.29 & 30.07 \\
																& glasses	&  6.75 & 5.48 & 4.83 & 30.48 \\
																& eye			&  6.99 & 5.71 & 5.47 & 35.87 \\
																& mouth		&  7.30 & 6.01 & 4.83 & 27.62 \\
																& hair		&  6.79 & 5.34 & 5.28 & 36.56 \\
\end{tabular}
}
\caption{\emph{Landmark distance statistics (in mm) for global and local models.}}
\label{tab_landmark_distance}
\end{table}

Table \ref{tab_surface_distance} shows the surface distances. We see that the local model produces slightly higher mean and standard deviation, but lower median errors. This reflects the fact that for the local model there are a relatively few points where there is not enough pull from the energy function to the data, and these points are not fitted well. In areas of the surface where the data is close enough to the initial alignment, however, the local model better fits to the surface than the global model due to the retained variability in the model. Conversely, the global model does not get as close to any localized area, but the global information allows the overall shape to guide it in areas where the initial alignment is not close enough to the data.

\begin{table}[ht]
\centering
{\small%footnotesize
\begin{tabular}{c | c | c | c | c | c }
Model & Occlusion & Mean & Median & Std. Dev. & Max \\
\hline
\multirow{2}{*}[-0.5cm]{global}	& none		&  1.06	&  0.81	&  0.91	&  14.00	\\
																& glasses	&  1.31	&  0.97	&  1.17	& 13.14	\\
																& eye			&  2.69	&  1.16	&  5.01	& 65.63	\\
																& mouth		&  3.57	&  1.62	&  5.33	& 46.93	\\
																& hair		&  3.17	&  1.34	&  5.63	&  46.53	\\
\hline
\multirow{2}{*}[-0.5cm]{local}	& none		&  0.80 & 0.47 & 1.06 & 15.63 \\
																& glasses	&  1.10 & 0.59 & 1.46 & 16.73 \\
																& eye			&  2.03 & 0.56 & 4.51 & 64.79 \\
																& mouth		&  2.42 & 0.75 & 4.13 & 47.70 \\
																& hair		&  2.28 & 0.67 & 4.80 & 51.26 \\
\end{tabular}
}
\caption{\emph{Surface distance statistics (in mm) for global and local models.}}
\label{tab_surface_distance}
\end{table}

\begin{figure}[t]
\centering
\begin{tabular}{cccc}
	no occlusion & &  &\\
	\includegraphics[height = 2.3cm]{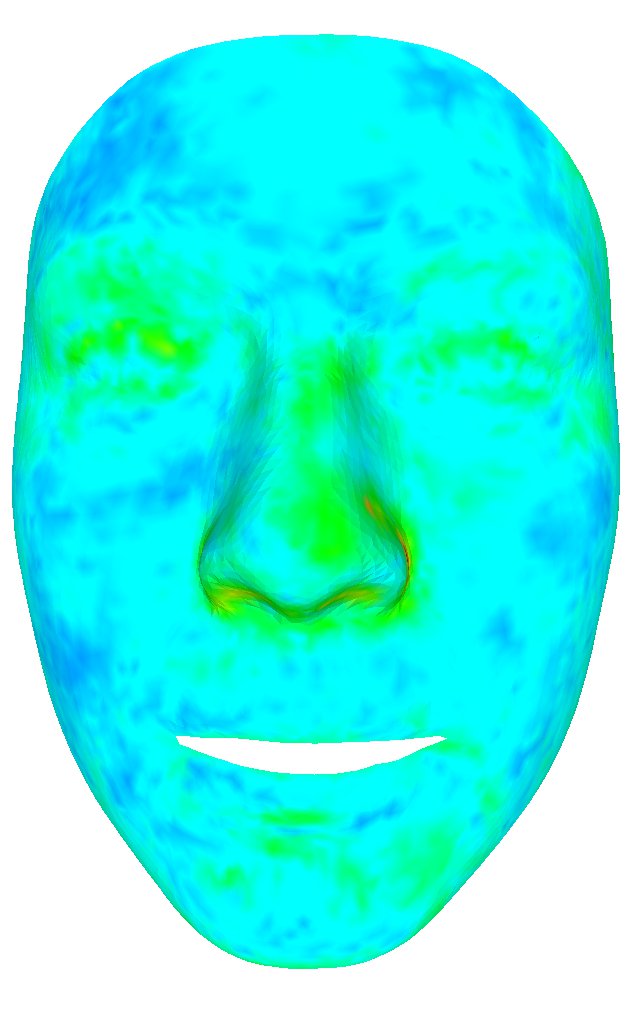}	& \includegraphics[height = 2.3cm]{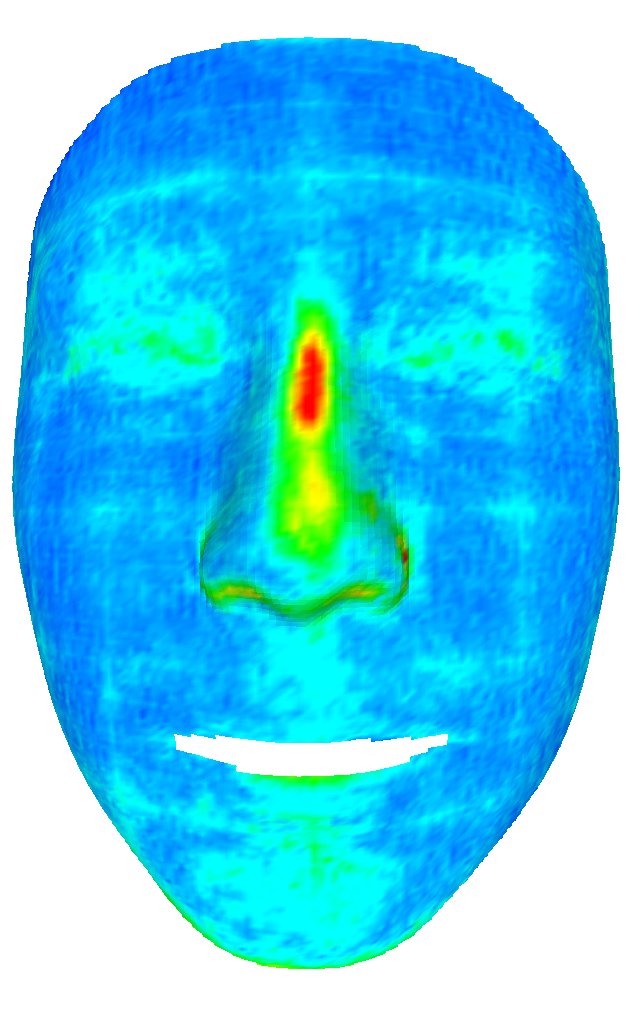}  
	& \includegraphics[height = 2.2cm]{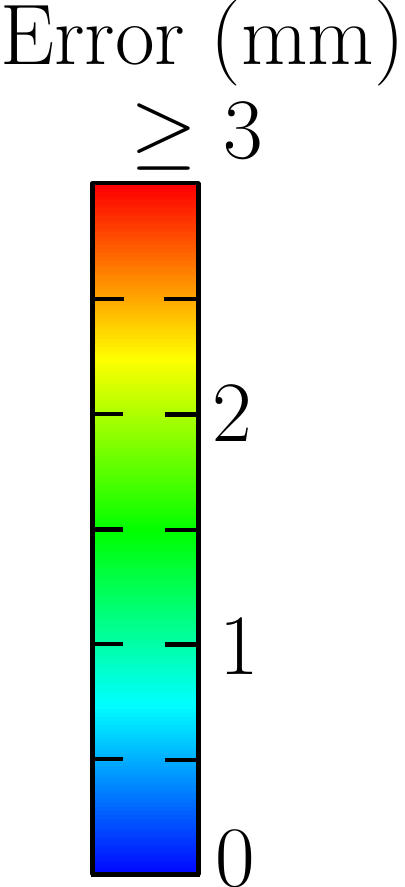} & \includegraphics[height = 2.2cm]{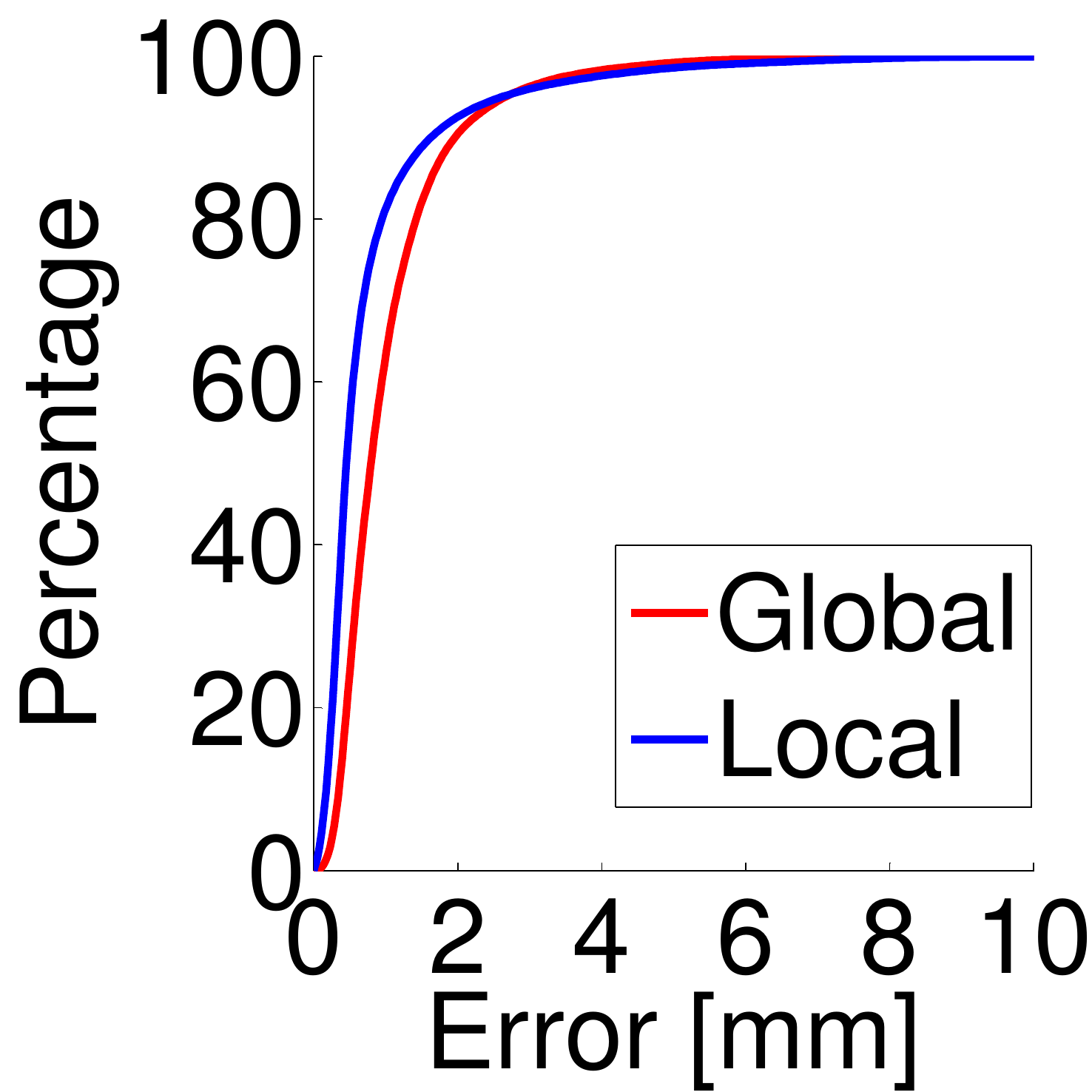} \\
	glasses  & & & \\
	\includegraphics[height = 2.3cm]{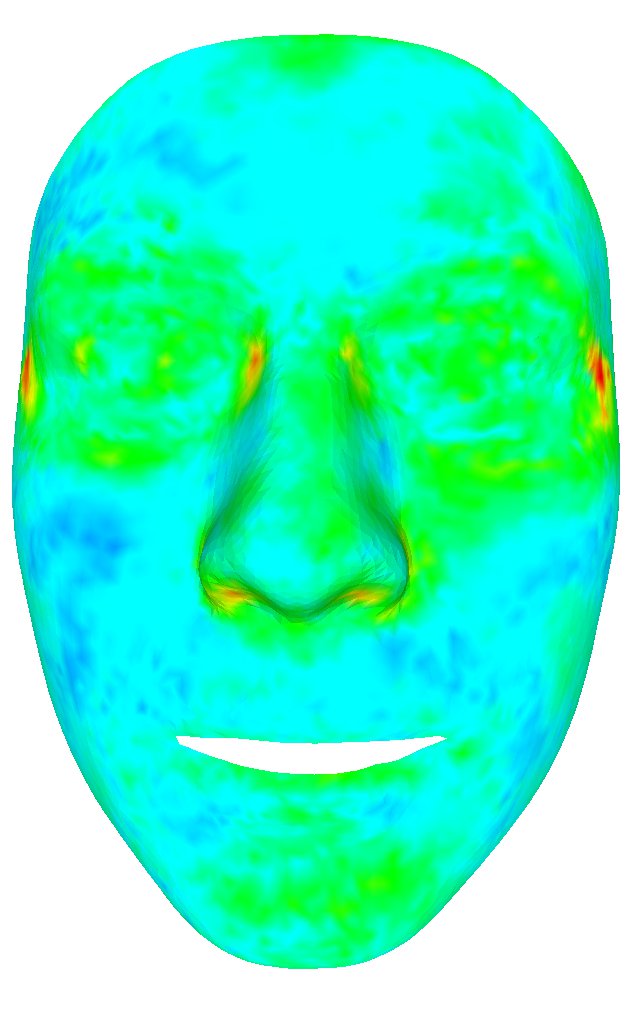}	& \includegraphics[height = 2.3cm]{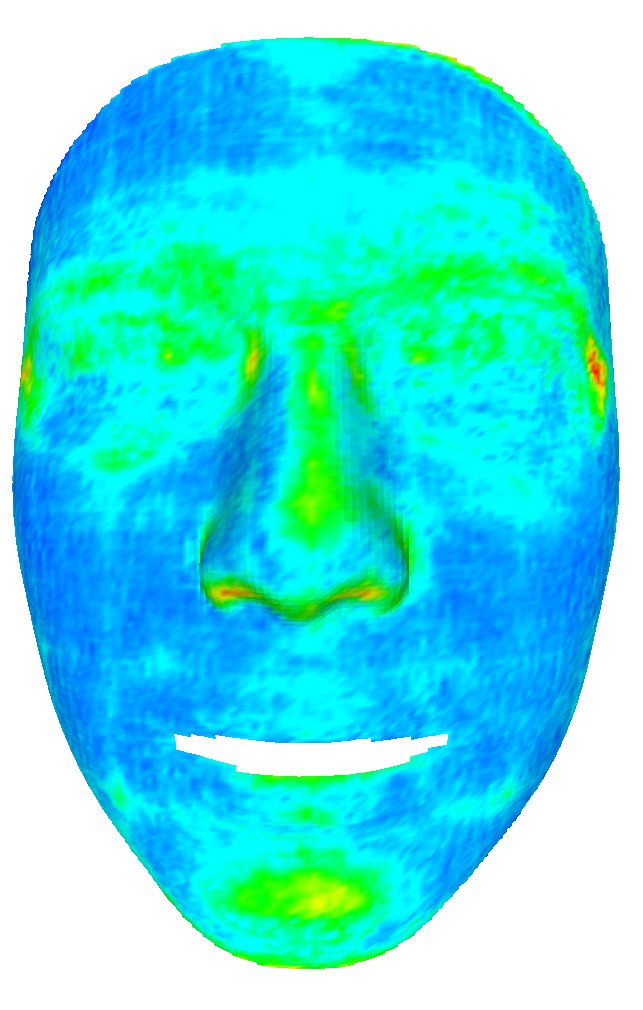}
	& \includegraphics[height = 2.2cm]{ErrorBar.pdf} & \includegraphics[height = 2.2cm]{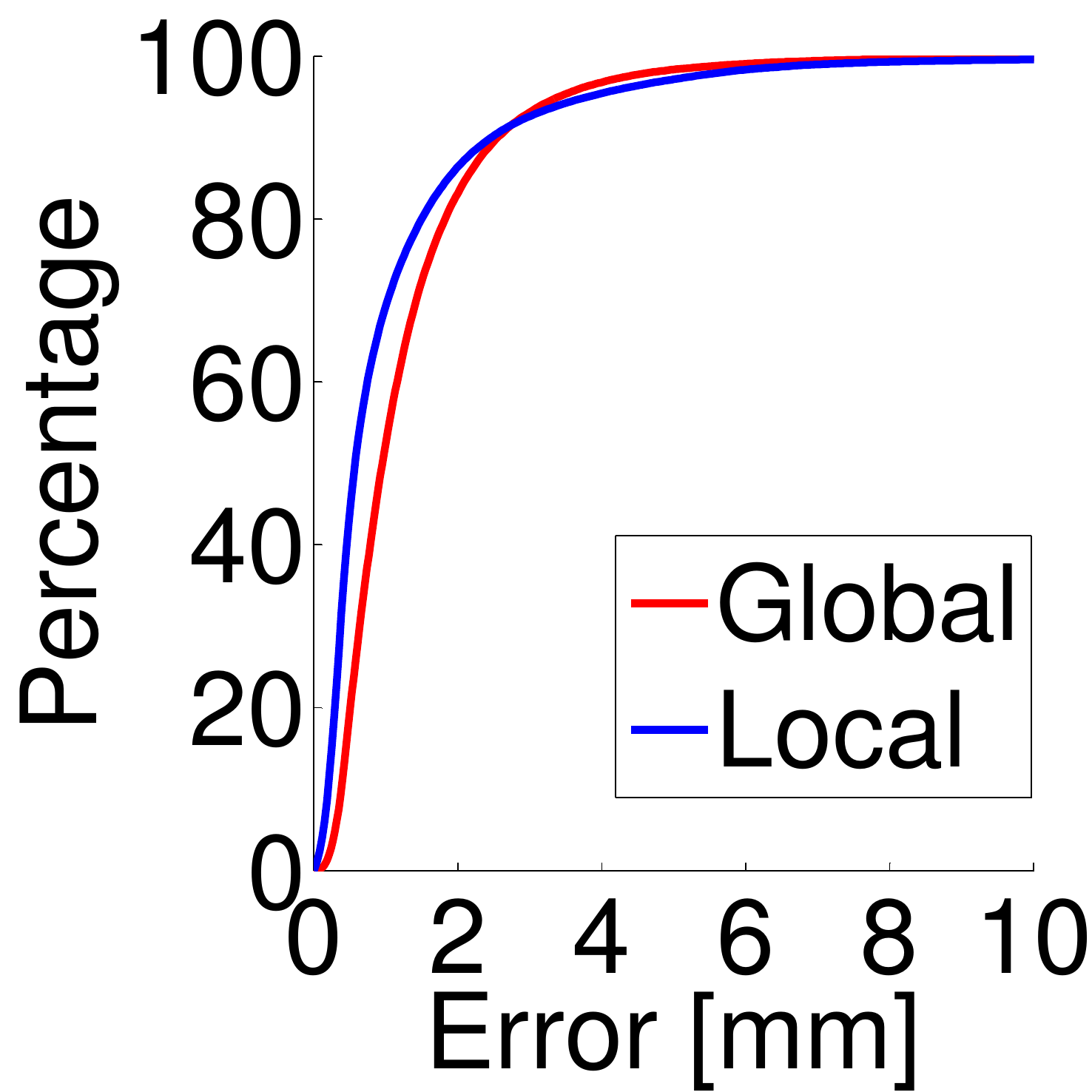} \\
	eye & & & \\
	\includegraphics[height = 2.3cm]{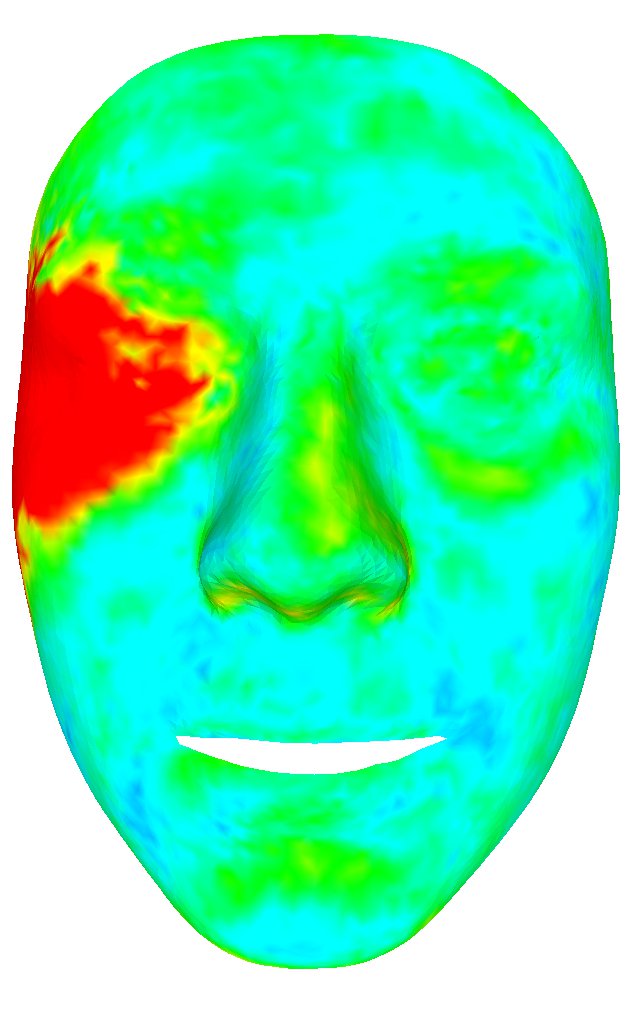}	& \includegraphics[height = 2.3cm]{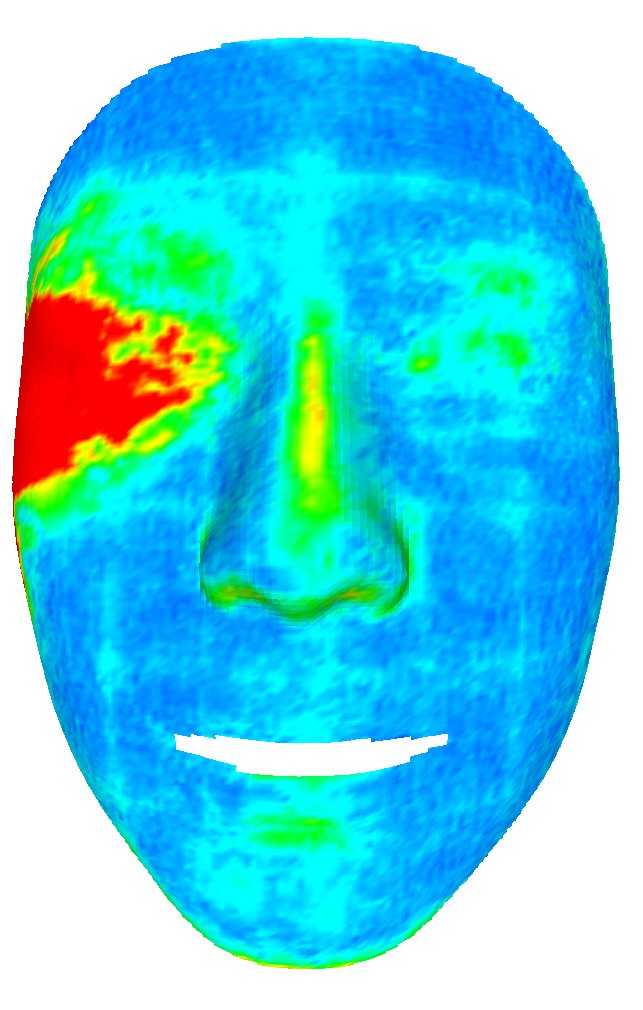} 
	& \includegraphics[height = 2.2cm]{ErrorBar.pdf} & \includegraphics[height = 2.2cm]{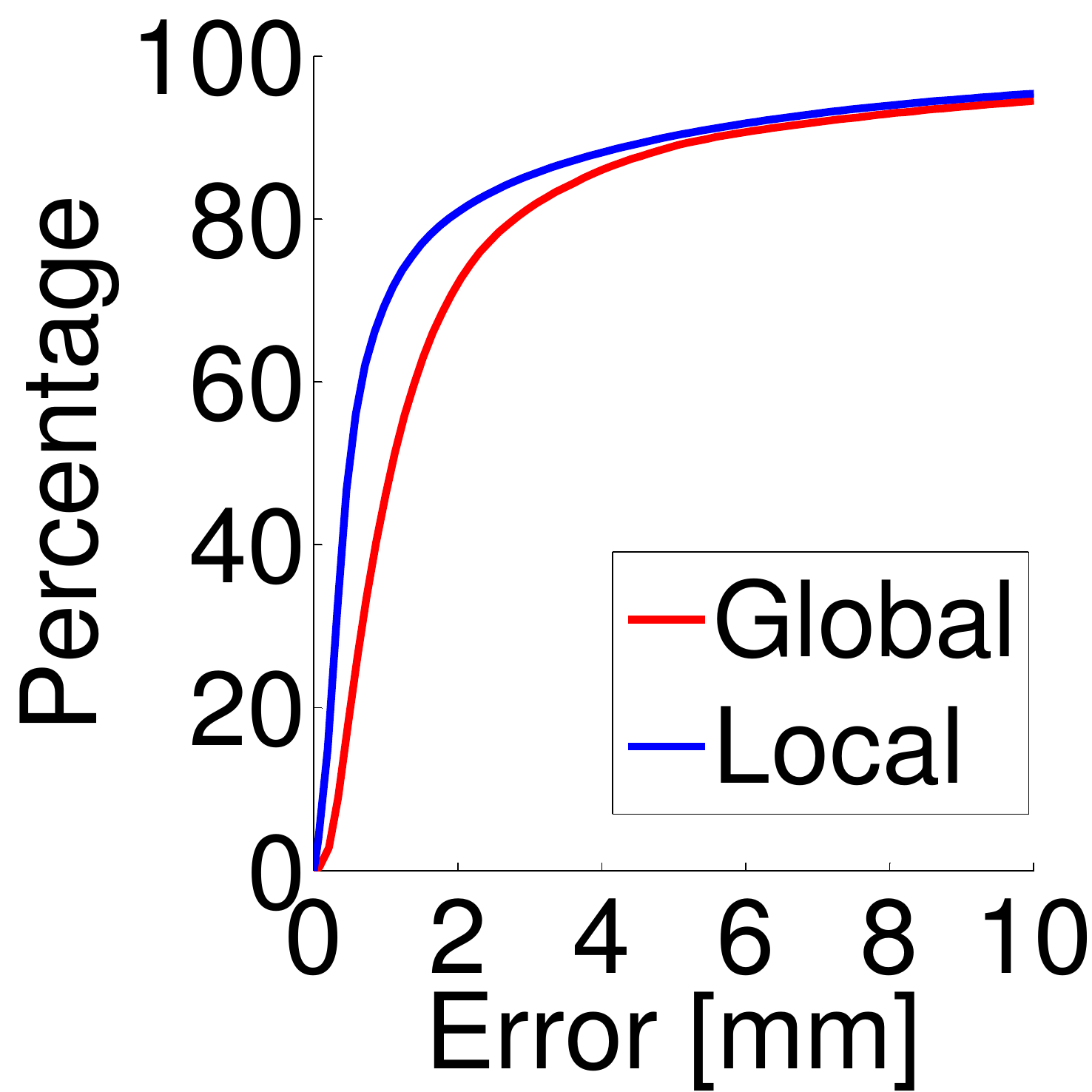} \\
	mouth & & & \\
	\includegraphics[height = 2.3cm]{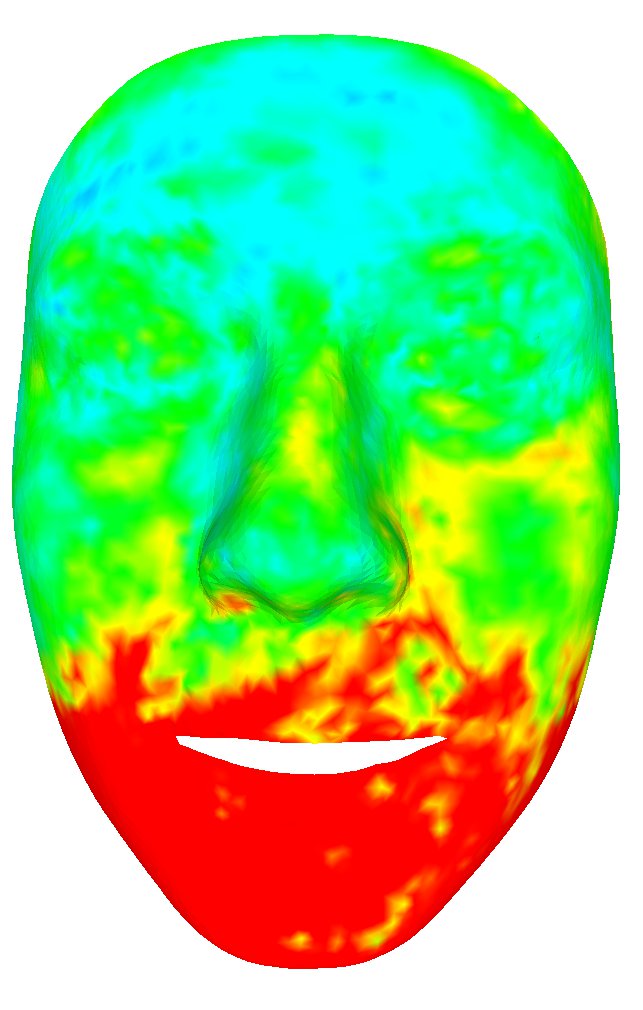} &	\includegraphics[height = 2.3cm]{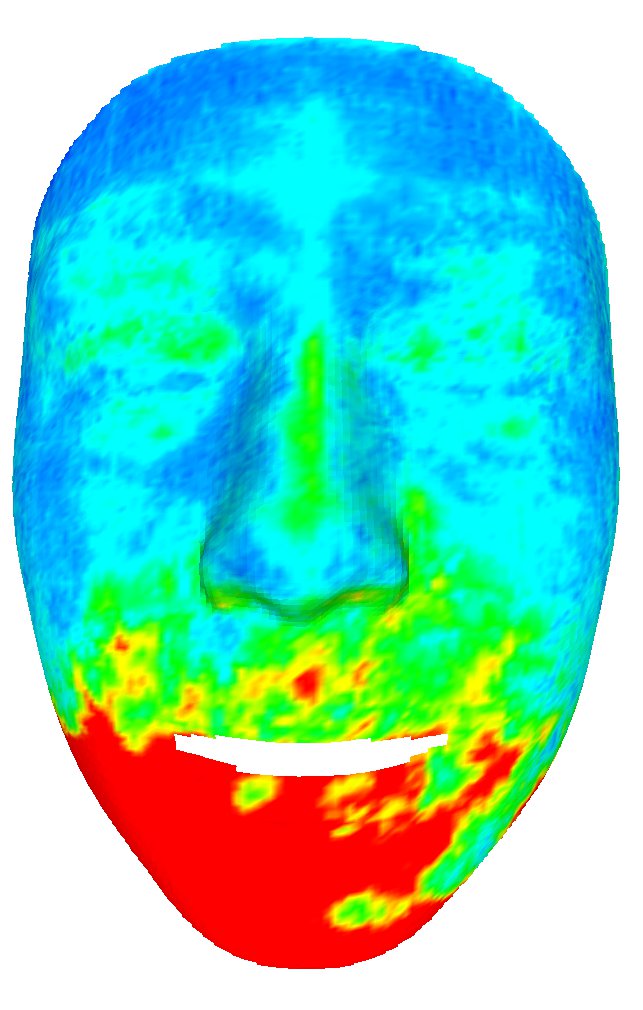}
	& \includegraphics[height = 2.2cm]{ErrorBar.pdf} & \includegraphics[height = 2.2cm]{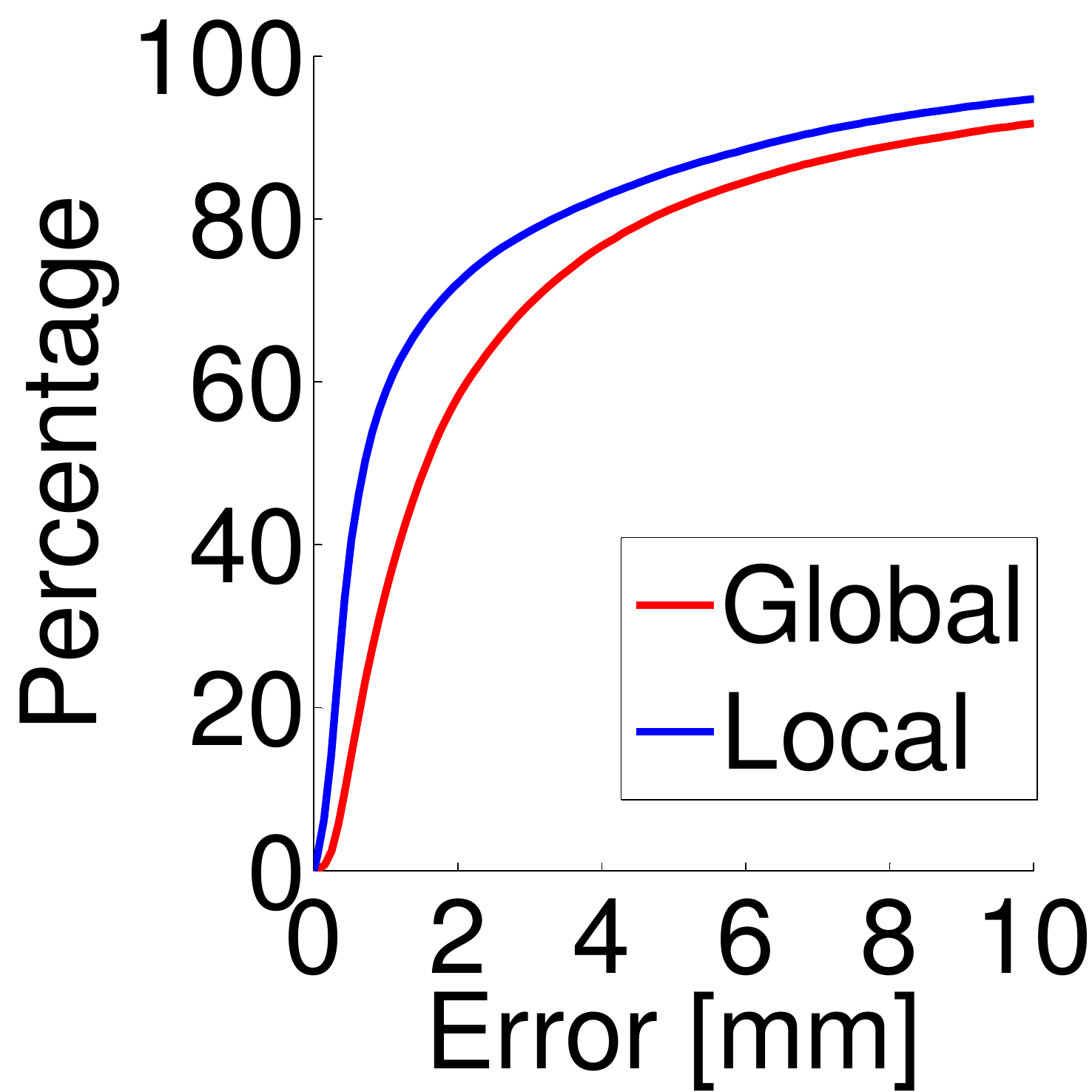} \\
	hair & & & \\
	\includegraphics[height = 2.3cm]{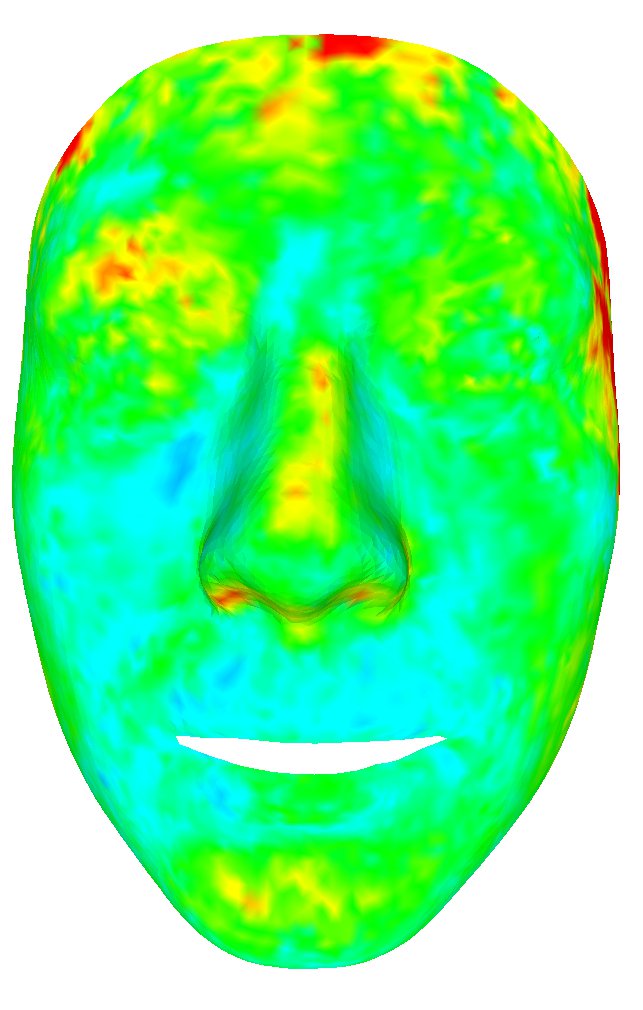}	& \includegraphics[height = 2.3cm]{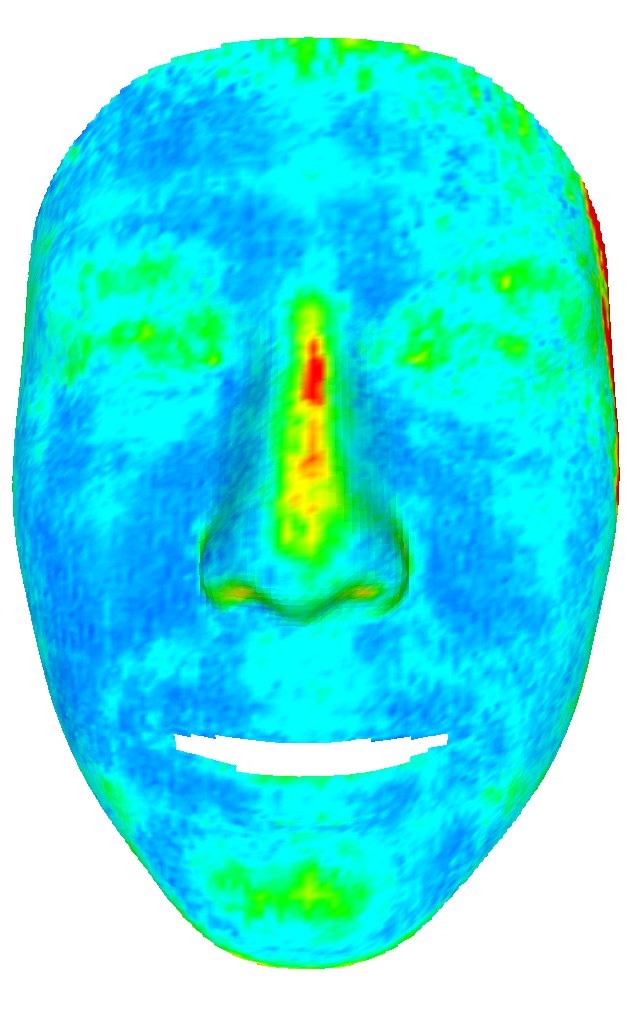}
	& \includegraphics[height = 2.2cm]{ErrorBar.pdf} & \includegraphics[height = 2.2cm]{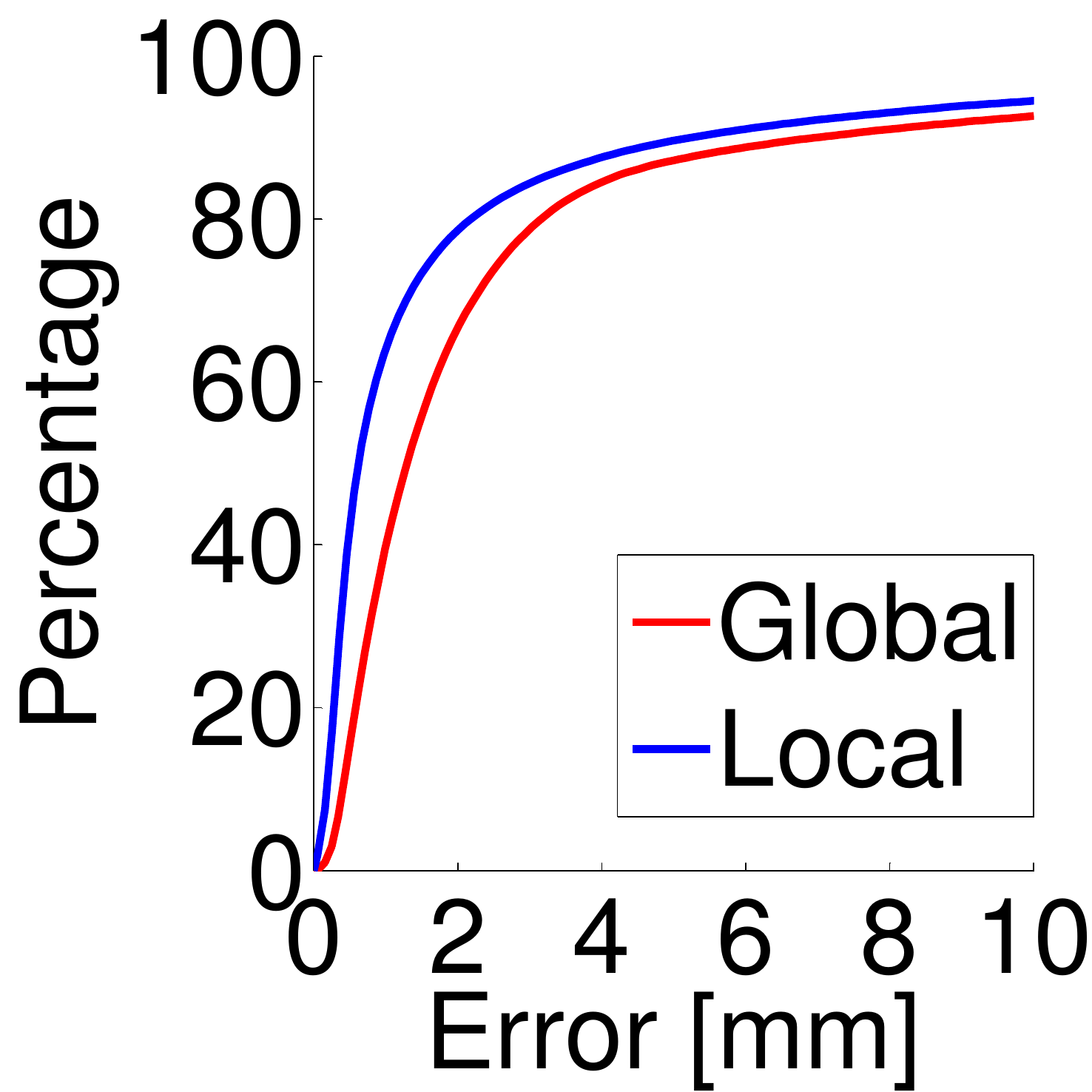} \\
  Global & Local & & \\
\end{tabular}
\caption{\emph{False color visualizations of the median nearest neighbor distances for different types of occlusion and cumulative error per vertex.}}
\label{fig:distancesNN+Occlusions}
\end{figure}

Figure~\ref{fig:distancesNN+Occlusions} shows the color coded median surface distances for all points. For the results without occlusion, in most regions of the face, the local shape space yields results that are closer to the input surface. However, at the nose tip and the chin, the global model is closer to the input surface than the local model. The reason is that the initialization is often poor in these regions and that as a result the local model does not fit these areas to the surface. For the models with occlusion, the additional error caused by the occlusion is generally more localized when using the local shape space than when using the global shape space. This is especially visible in the region around the left eye for the examples where the right eye is occluded by a hand (third row of Figure~\ref{fig:distancesNN+Occlusions}). For the local shape space, the region around the left eye has low average fitting error, while for the global shape space, this region has larger average fitting error because it is influenced by errors in the (symmetric) region around the right eye. 

\begin{figure*}
\centering
\begin{tabular}{l c c c c c c}
\multirow{1}{*}[2.0cm]{{\small no occlusion}} & 
\includegraphics[width = 0.14\textwidth]{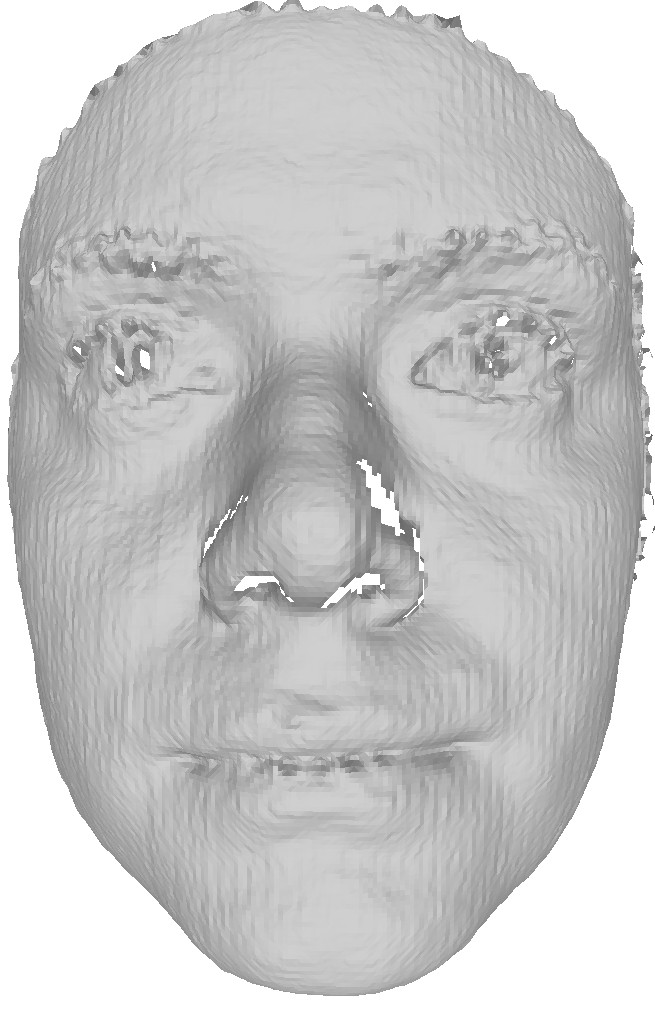} & 
\includegraphics[width = 0.14\textwidth]{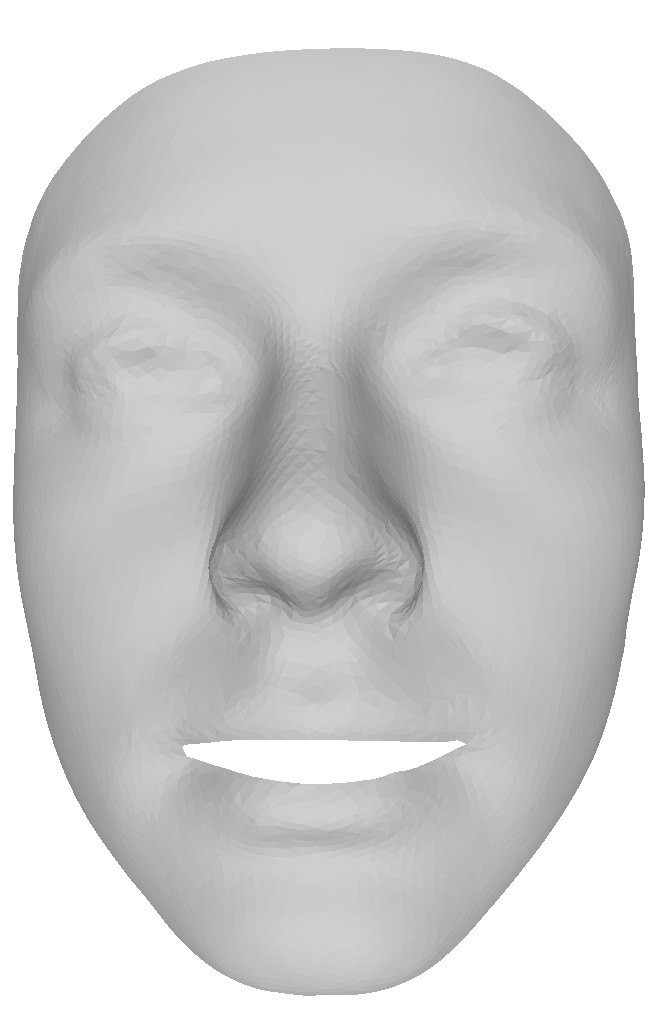} & 
\includegraphics[width = 0.14\textwidth]{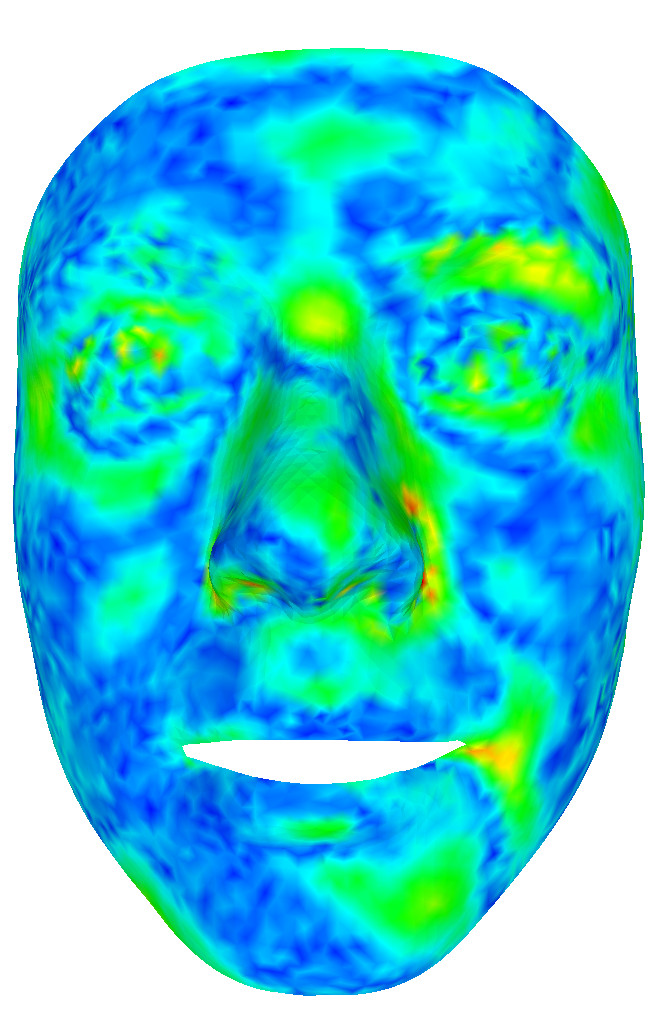} &  
\includegraphics[width = 0.14\textwidth]{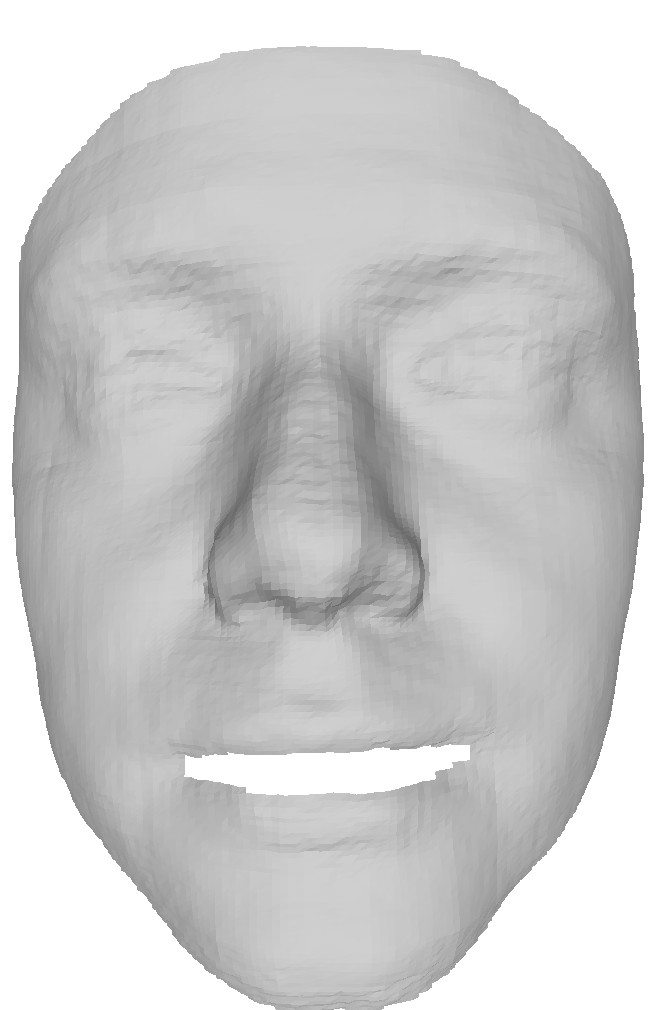} &  
\includegraphics[width = 0.14\textwidth]{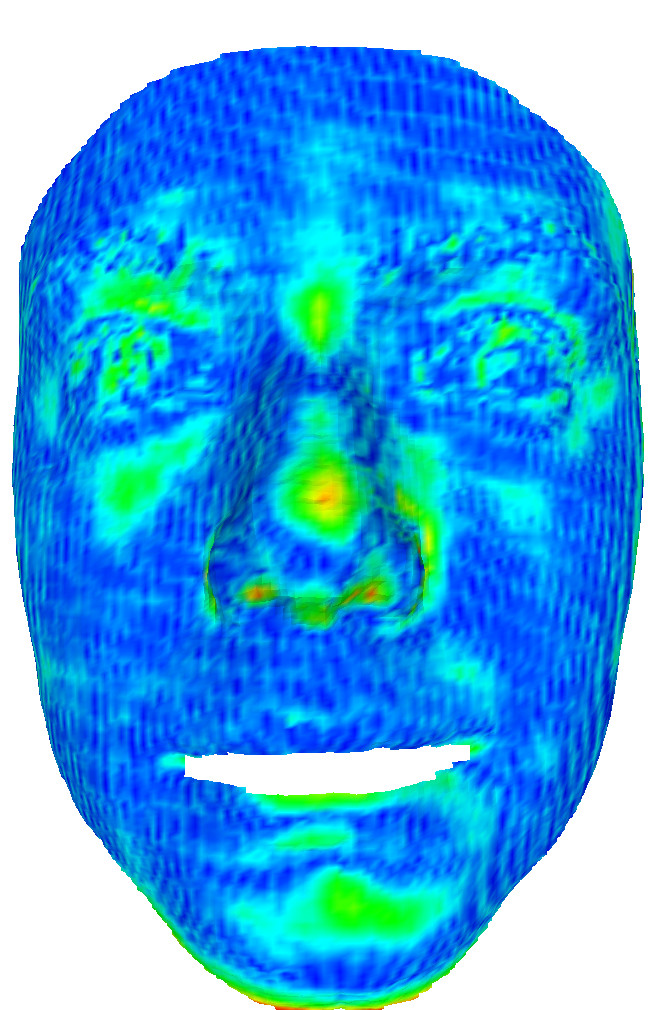} &
\includegraphics[width = 1.2cm]{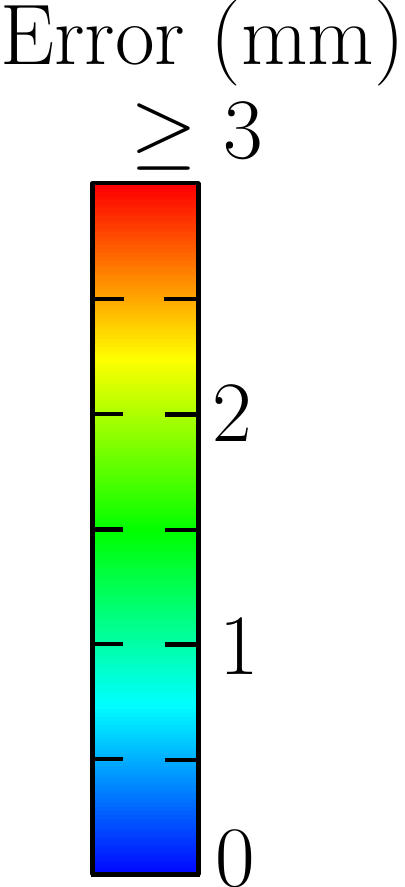}\\

\multirow{1}{*}[2.0cm]{{\small glasses}} &
\includegraphics[width = 0.14\textwidth]{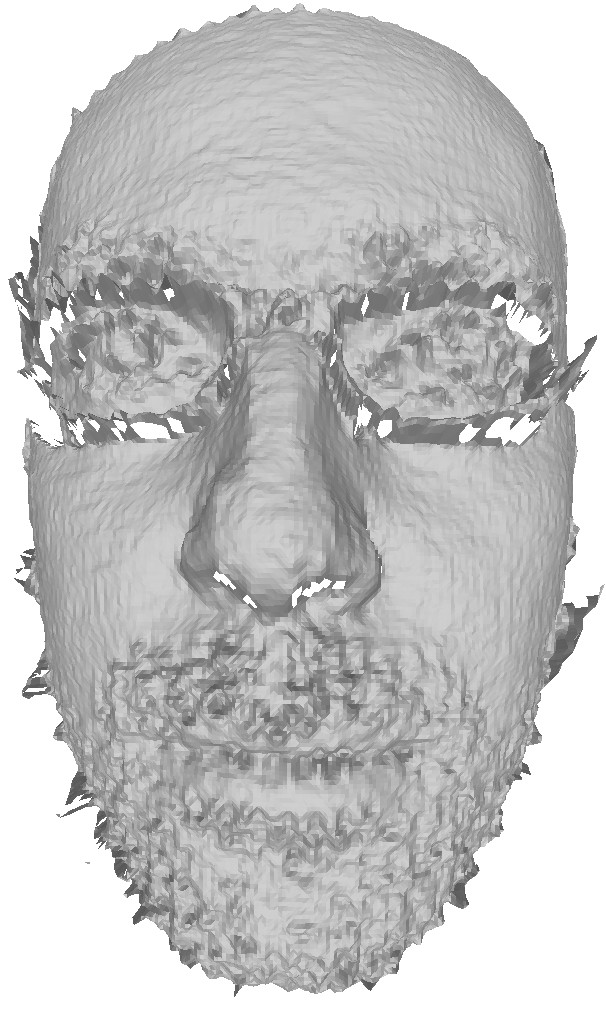} & 
\includegraphics[width = 0.14\textwidth]{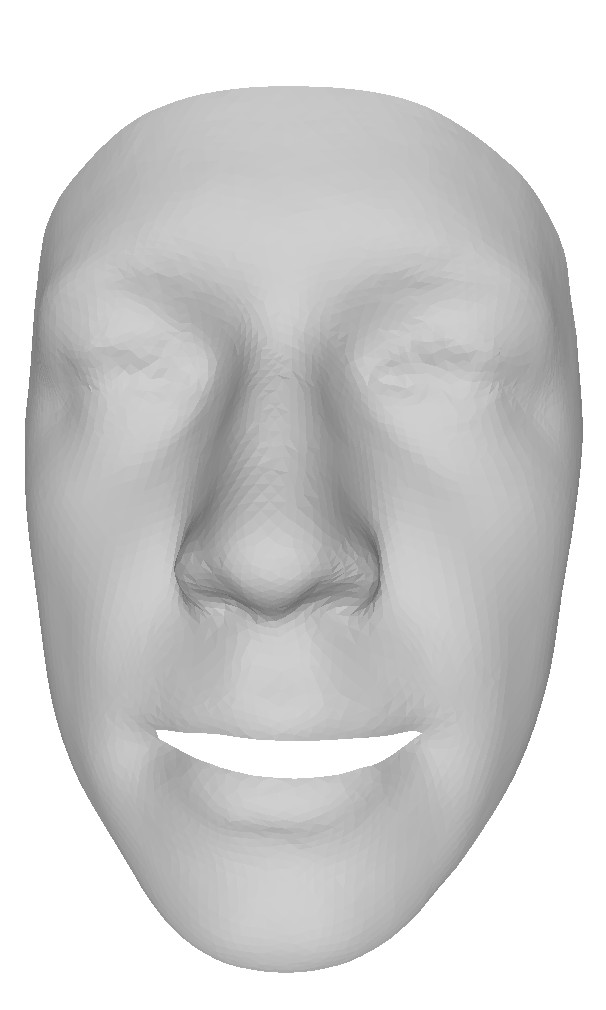} & 
\includegraphics[width = 0.14\textwidth]{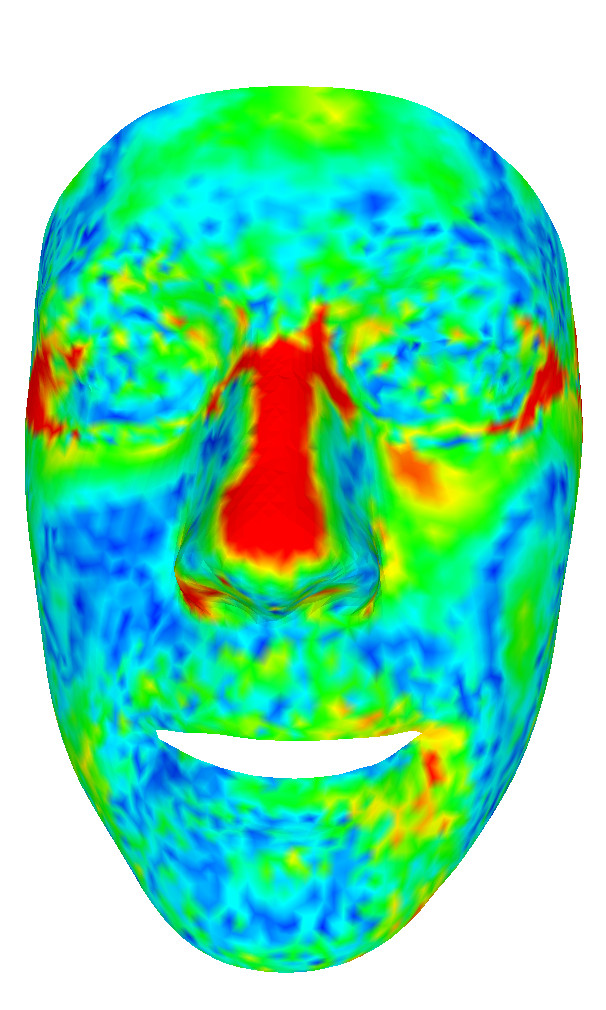} &  
\includegraphics[width = 0.14\textwidth]{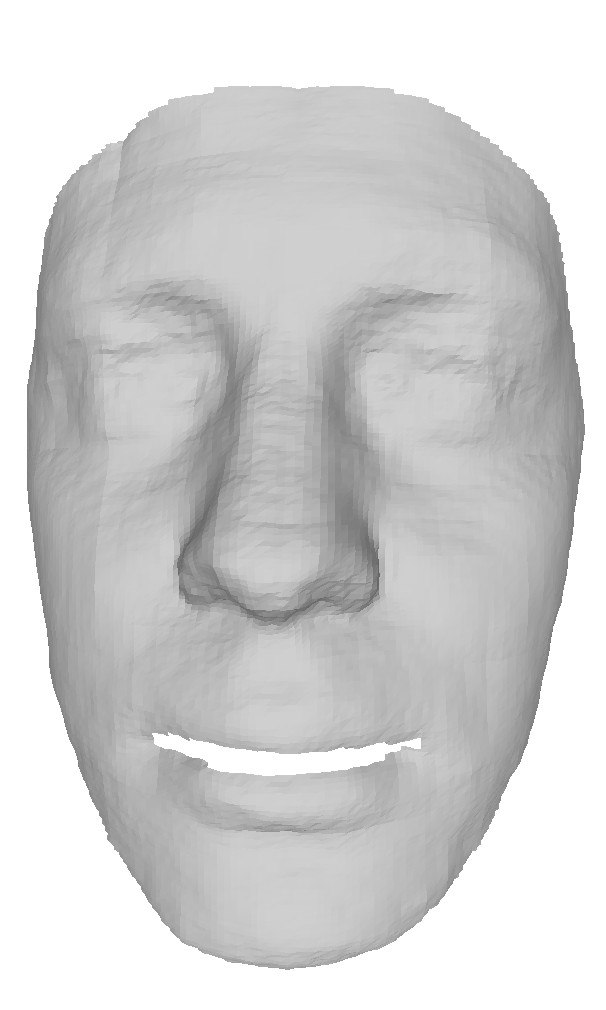} &  
\includegraphics[width = 0.14\textwidth]{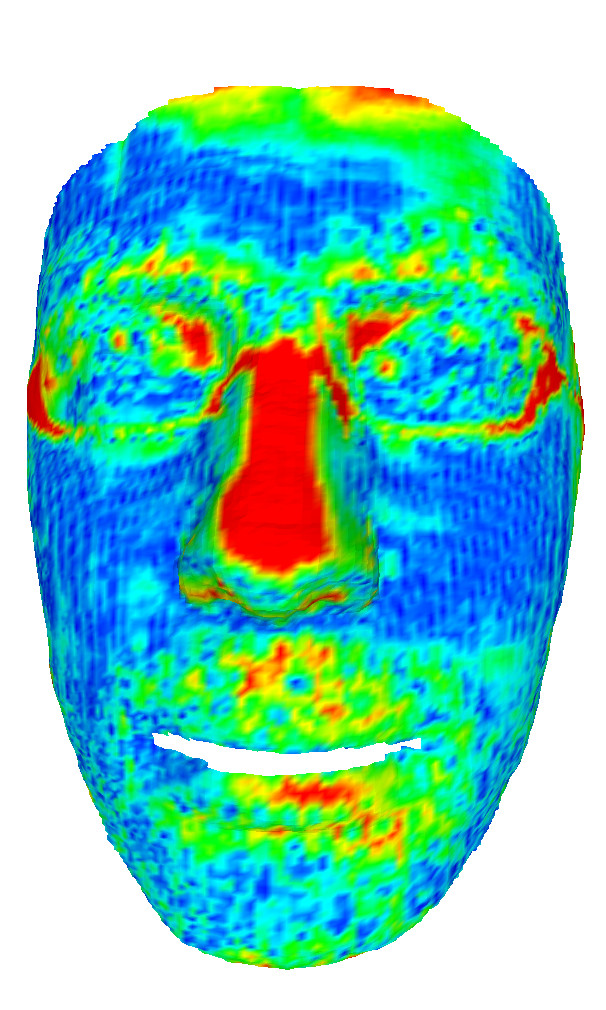} &
\includegraphics[width = 1.2cm]{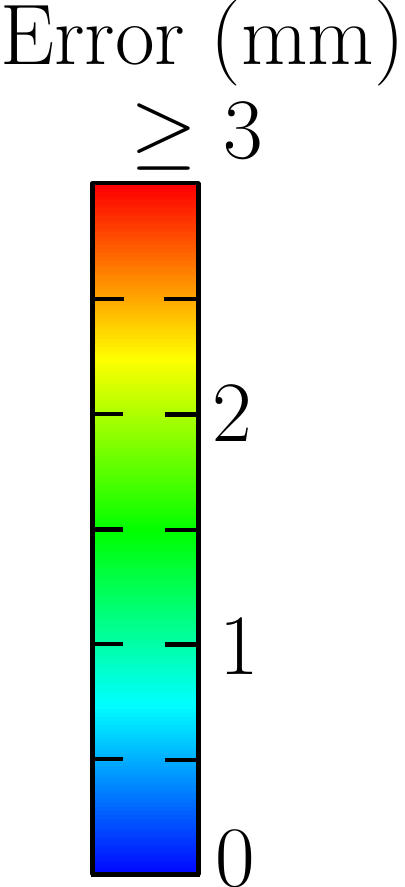}\\

\multirow{1}{*}[2.0cm]{{\small eye}} &
\includegraphics[width = 0.14\textwidth]{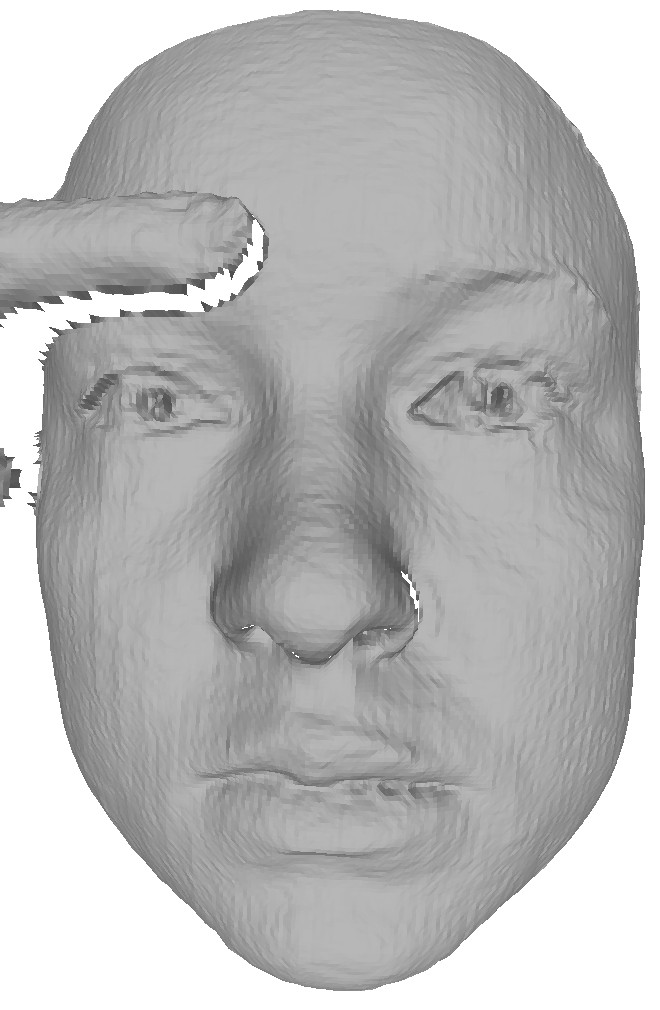} & 
\includegraphics[width = 0.14\textwidth]{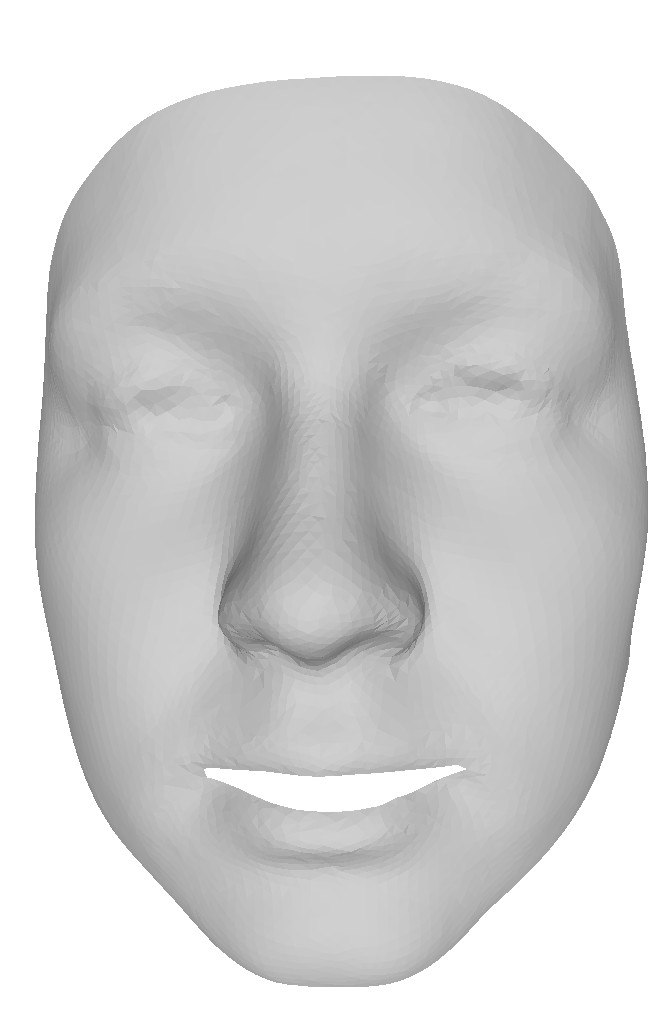} & 
\includegraphics[width = 0.14\textwidth]{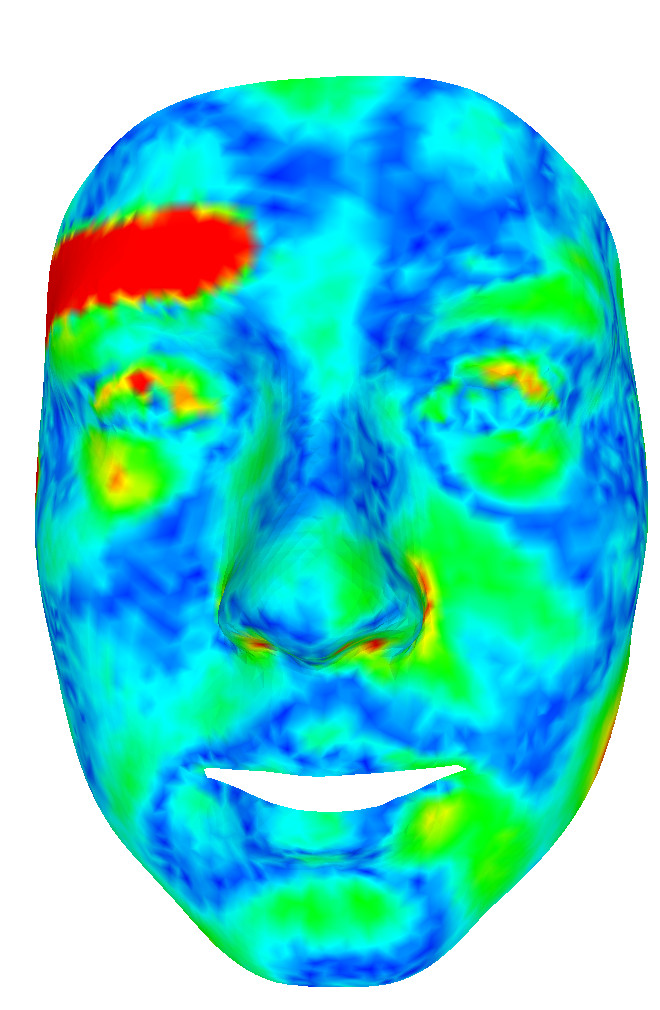} &  
\includegraphics[width = 0.14\textwidth]{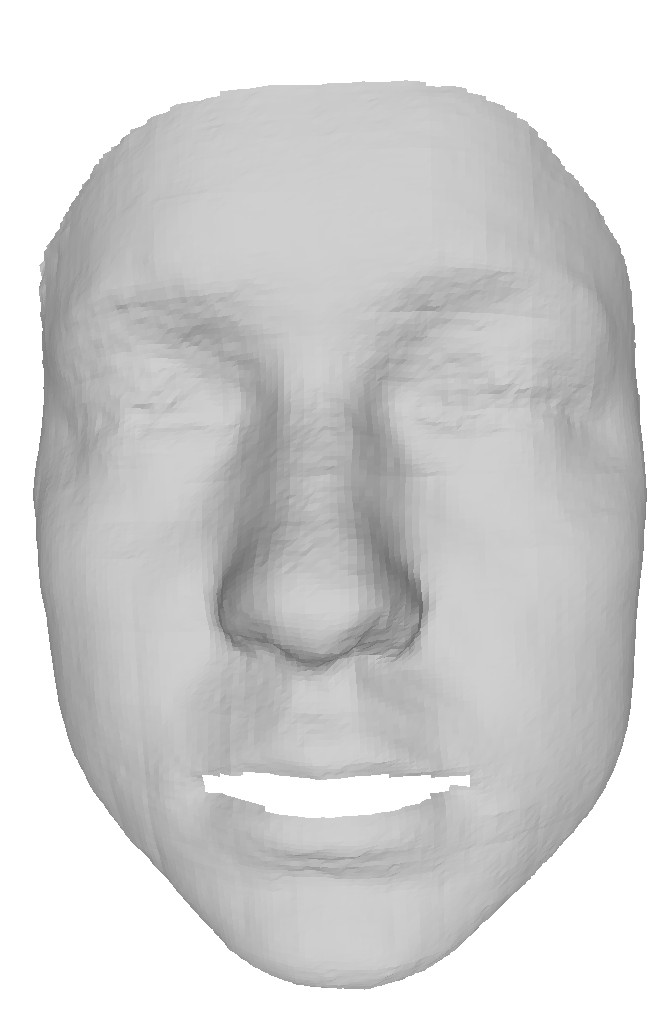} &  
\includegraphics[width = 0.14\textwidth]{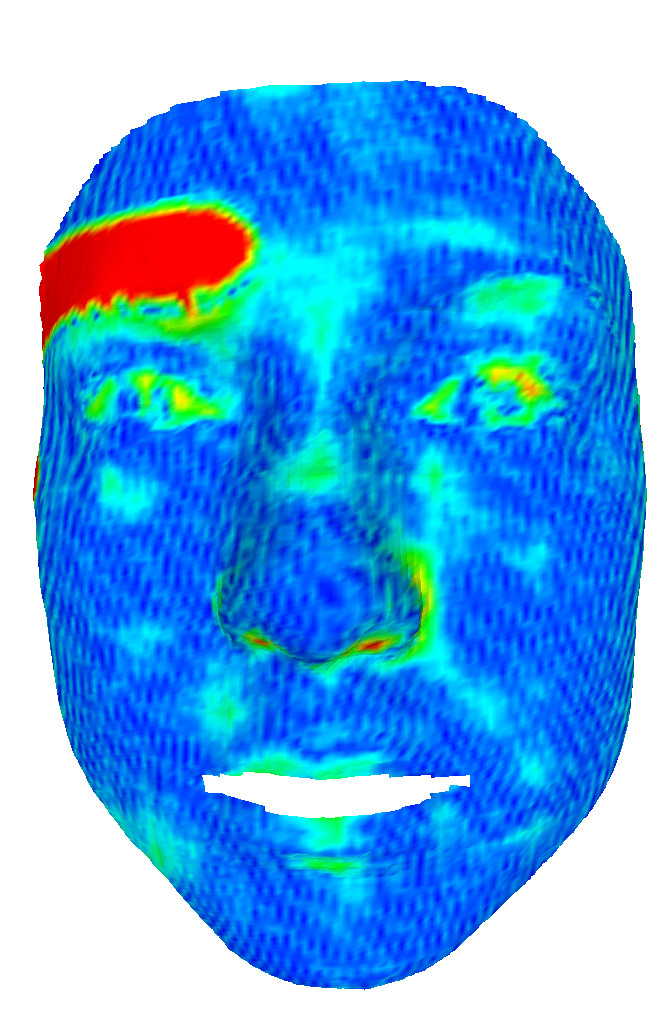} &
\includegraphics[width = 1.2cm]{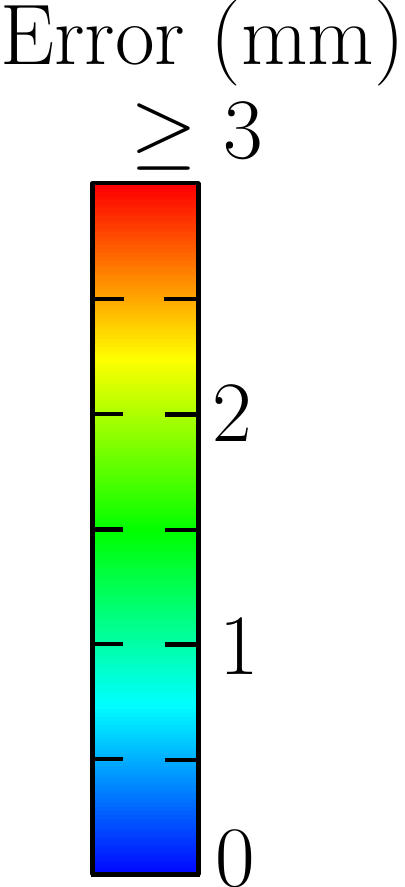}\\

\multirow{1}{*}[2.0cm]{{\small mouth}} &
\includegraphics[width = 0.14\textwidth]{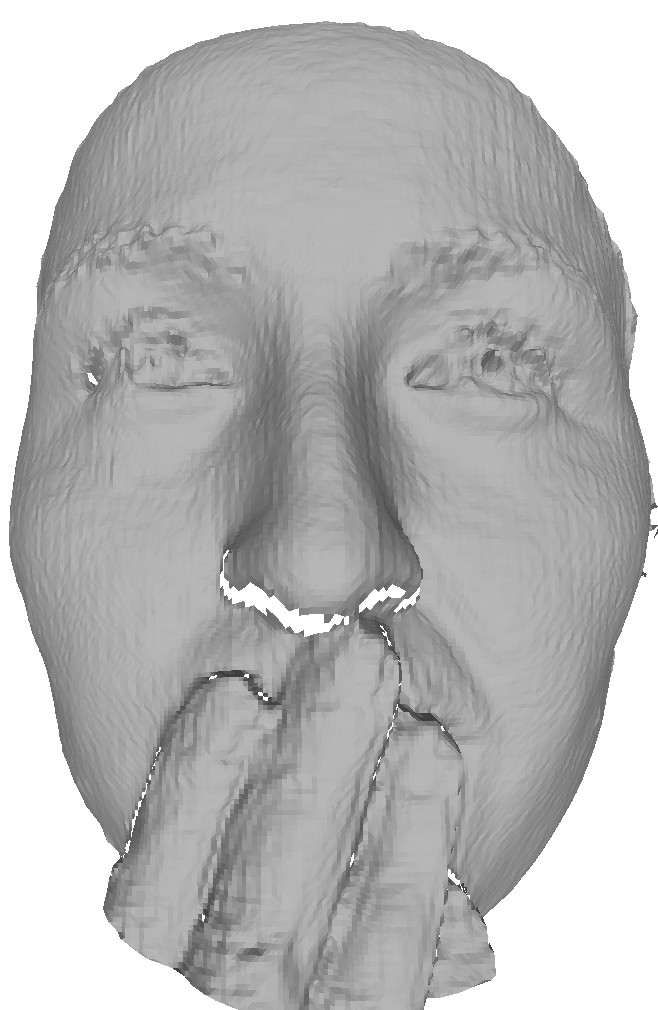} & 
\includegraphics[width = 0.14\textwidth]{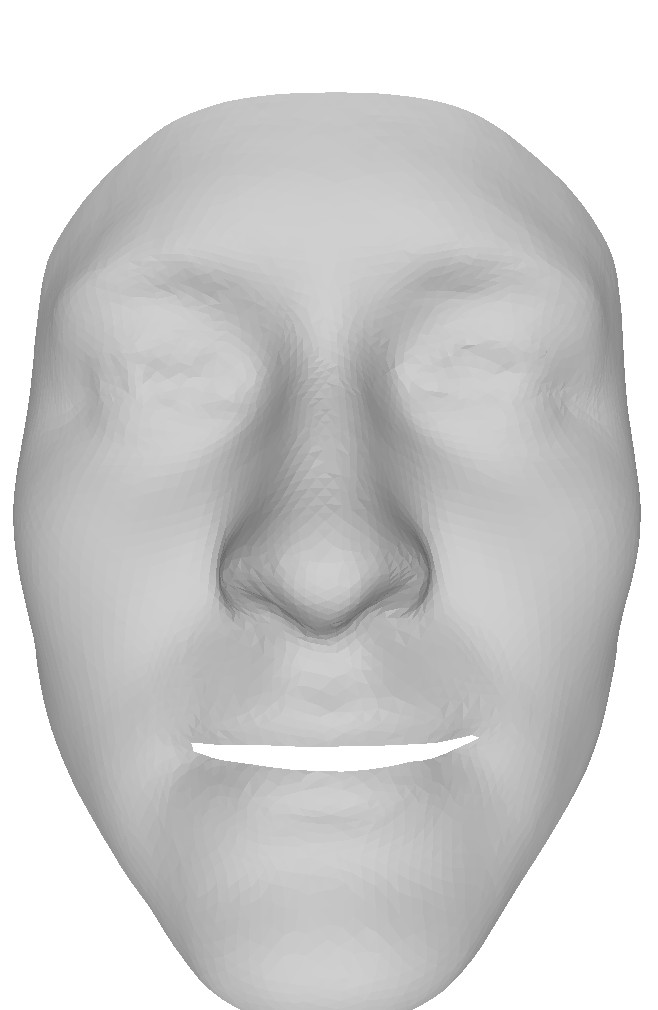} & 
\includegraphics[width = 0.14\textwidth]{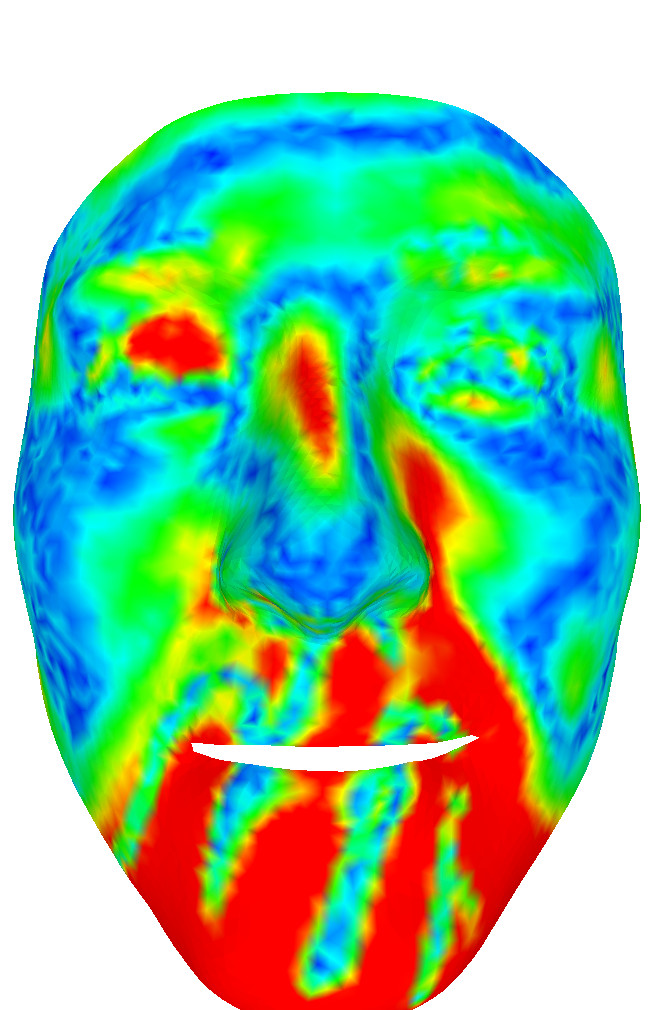} &  
\includegraphics[width = 0.14\textwidth]{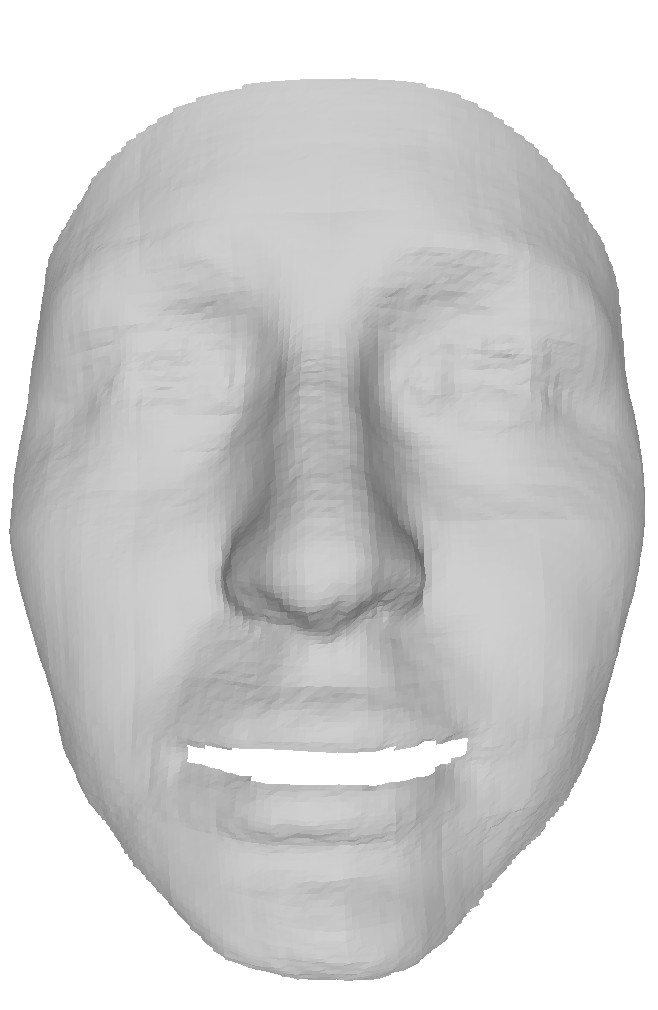} &  
\includegraphics[width = 0.14\textwidth]{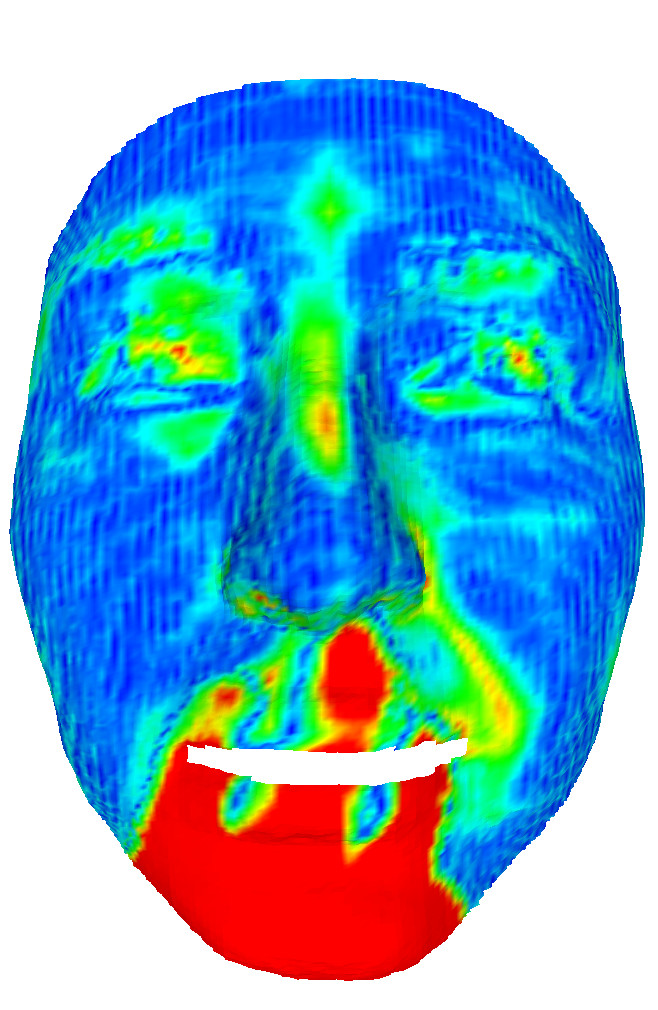} &
\includegraphics[width = 1.2cm]{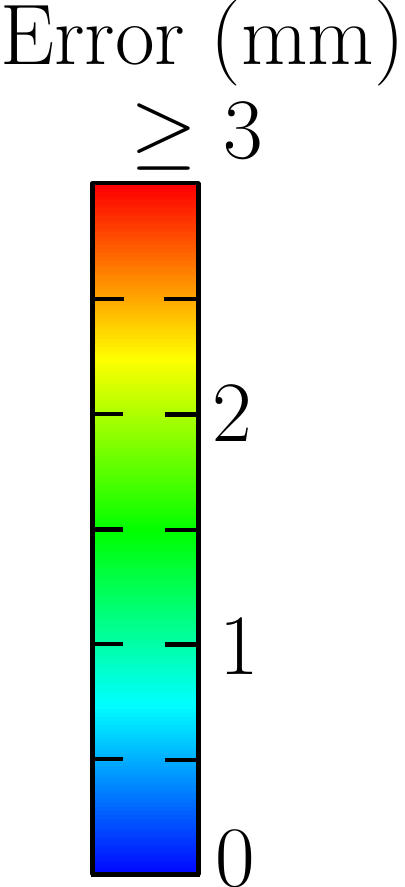}\\

\multirow{1}{*}[2.0cm]{{\small hair}} &
\includegraphics[width = 0.14\textwidth]{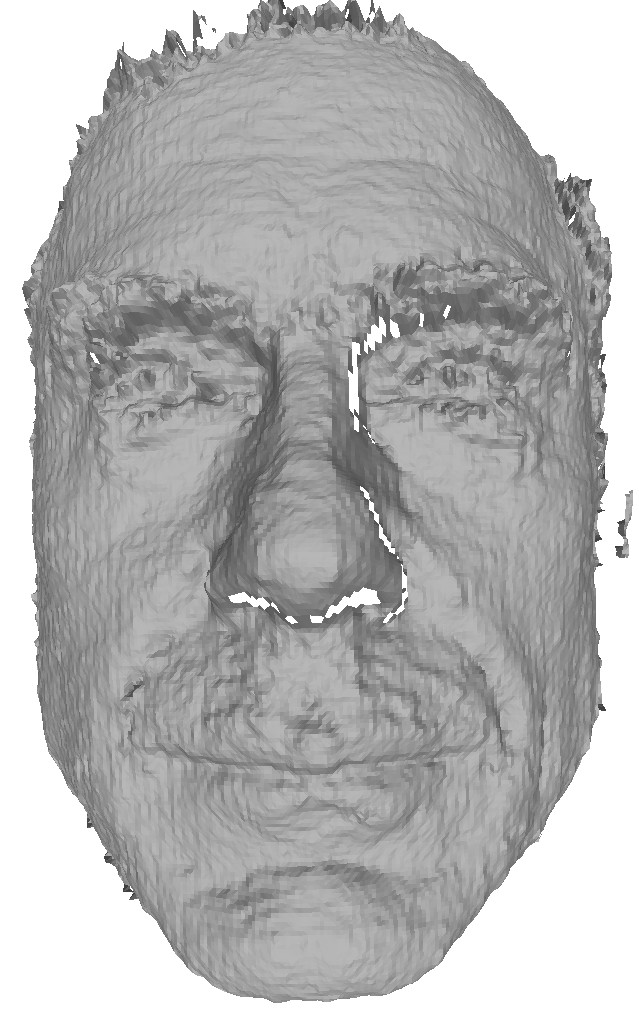} & 
\includegraphics[width = 0.14\textwidth]{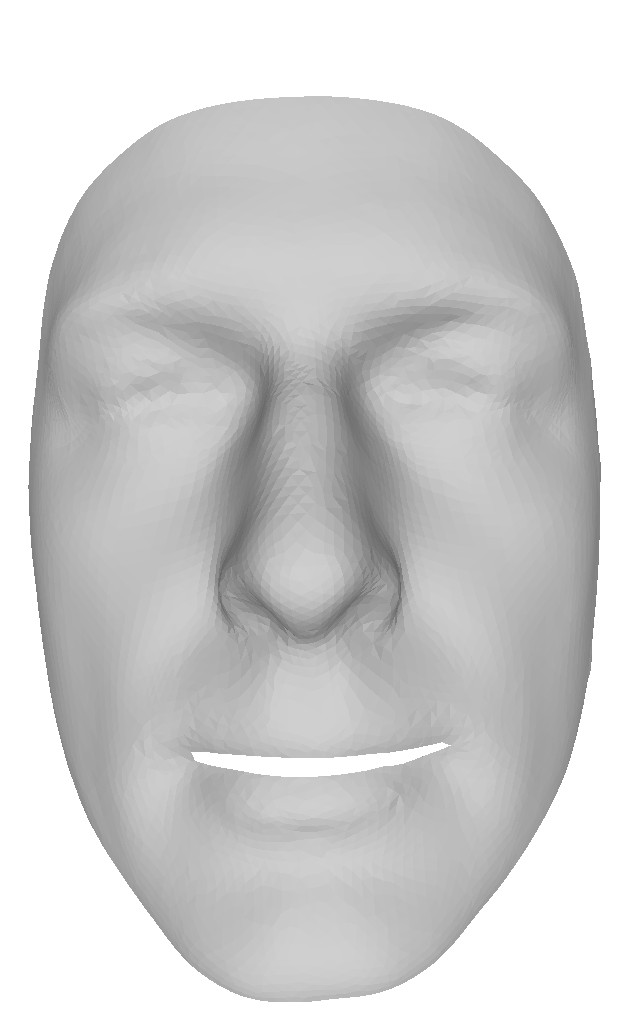} & 
\includegraphics[width = 0.14\textwidth]{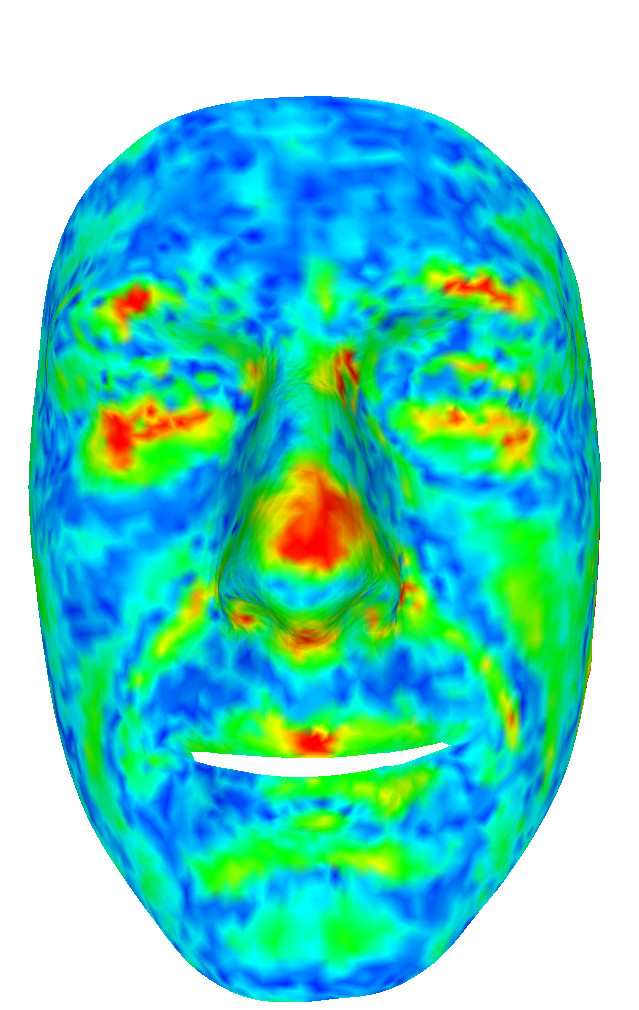} &  
\includegraphics[width = 0.14\textwidth]{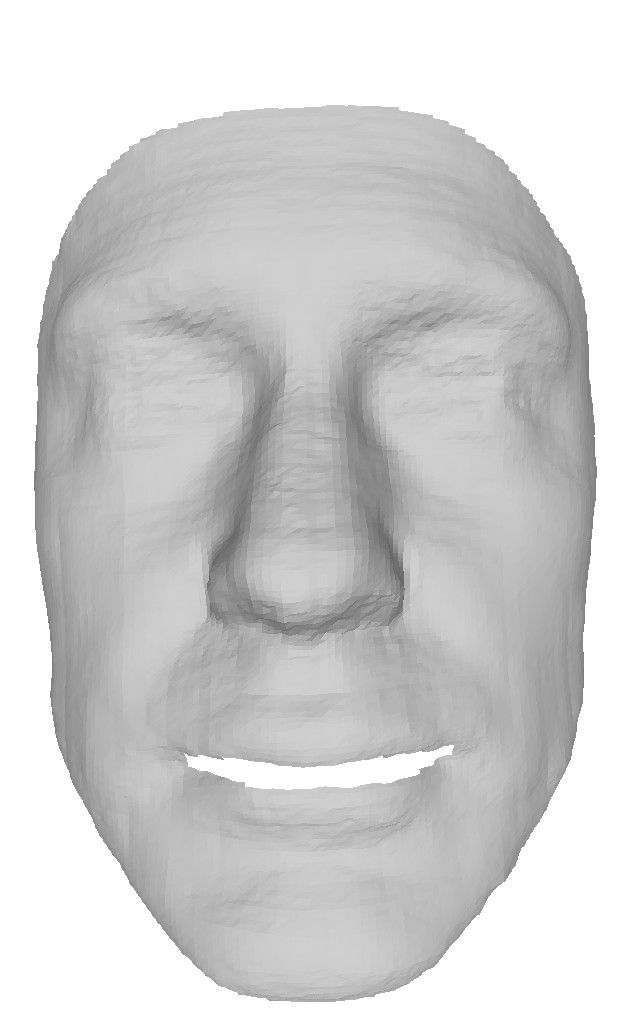} &  
\includegraphics[width = 0.14\textwidth]{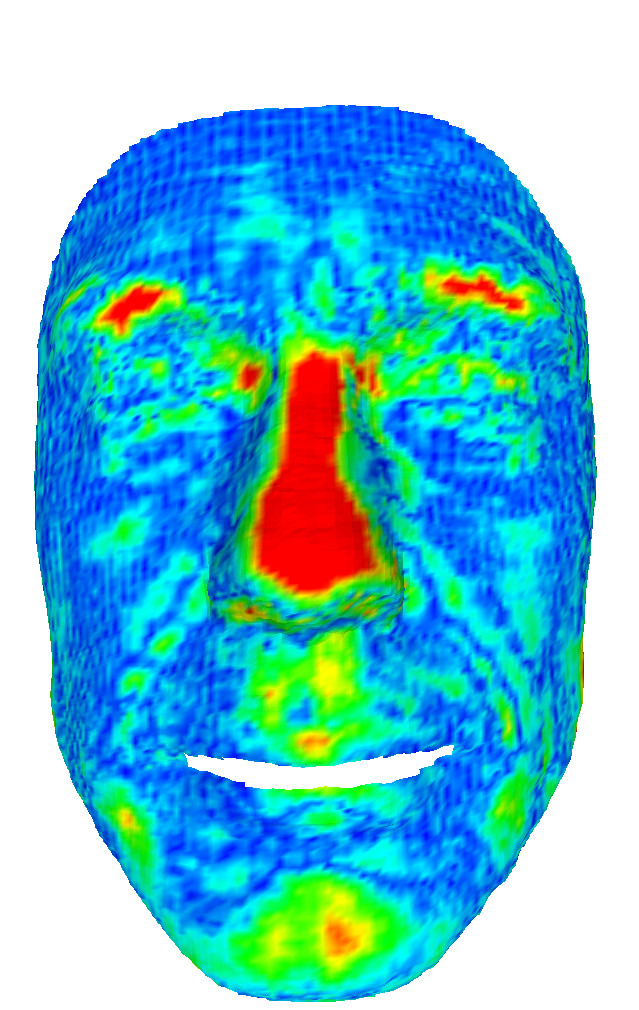} &
\includegraphics[width = 1.2cm]{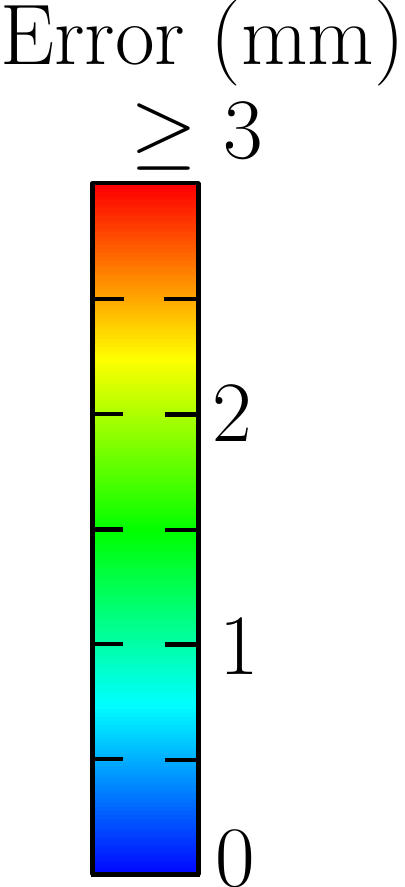}\\

& {\small input} & \multicolumn{2}{c}{{\small global model}} & \multicolumn{2}{c}{{\small local model}} & \\
\end{tabular}
\caption{\emph{Some fitting results. Each row shows from left to right: input data, result of global fitting, color coding of distances between global fitting result and input data, result of local fitting, color coding of distances between local fitting result and input data, and the legend for the color coding.}}
\label{fig:someResults}
\end{figure*}

Finally, we show a visual evaluation. Figure~\ref{fig:someResults} shows some examples of the fitted models for visual evaluation. Both models fit the shape model close to the input data for all of the examples. Note that overall, the results of the local method capture more shape detail than the results of the global method and that in most areas of the face, the results of the local method are fitted closer to the data than the results of the global method. A notable exception is the nose area of the subject shown in the last row of the figure. The reason is that the initialization is poor in this region, which is discussed in detail above. 

The third row of Figure~\ref{fig:someResults} shows a facial expression that is asymmetric in the cheek area. The output of the global method is a fairly symmetric face since the global shape prior learned the symmetric structure of the face. The output of the local method correctly captures the asymmetry in the reconstruction since the local shape prior allows for more flexibility in localized shape differences.

\begin{figure}[t]
\centering
\begin{tabular}{c c c}
\includegraphics[width = 0.12\textwidth]{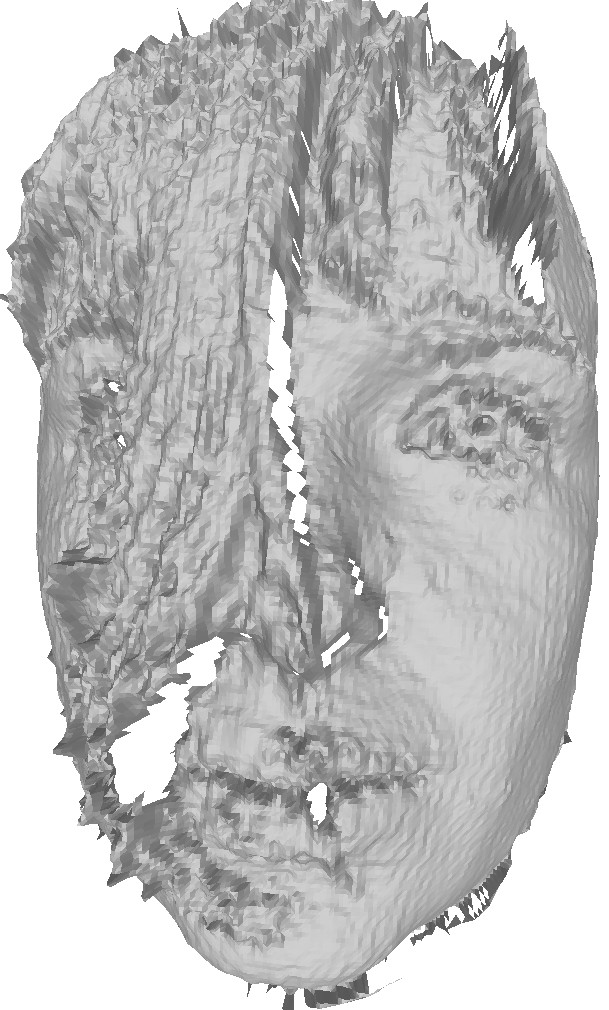} &  
\includegraphics[width = 0.12\textwidth]{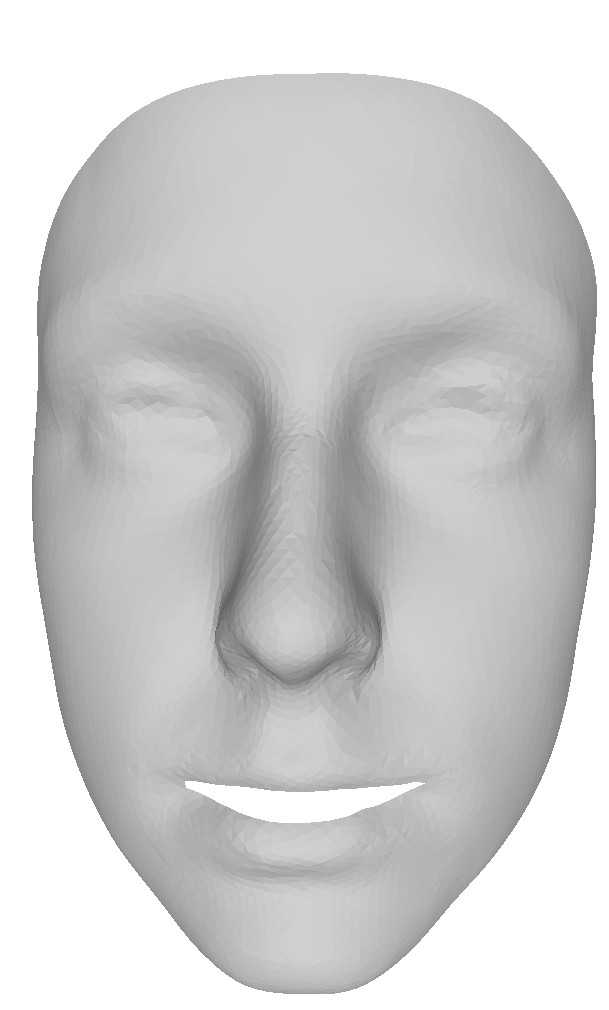} &
\includegraphics[width = 0.12\textwidth]{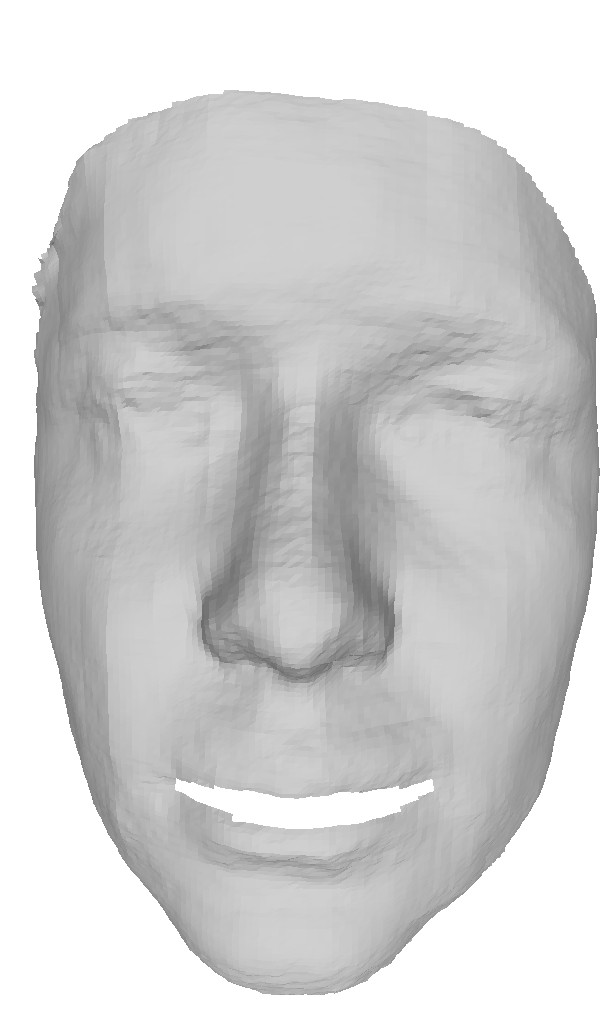}\\
{\small input} & {\small global model} & {\small local model} \\
\end{tabular}
\caption{\emph{Fitting results for model with extreme occlusion.}}
\label{fig:extremeHair}
\end{figure}

Figure~\ref{fig:extremeHair} shows the results for a challenging case where a large part of the input face is occluded by hair. Note that in spite of the large occlusions, visually satisfactory results are found by both methods.

\subsection{Evaluation with Noisy Scans}

\begin{figure}[ht]
\centering
\begin{tabular}{c c c}
\includegraphics[height = 3.0cm]{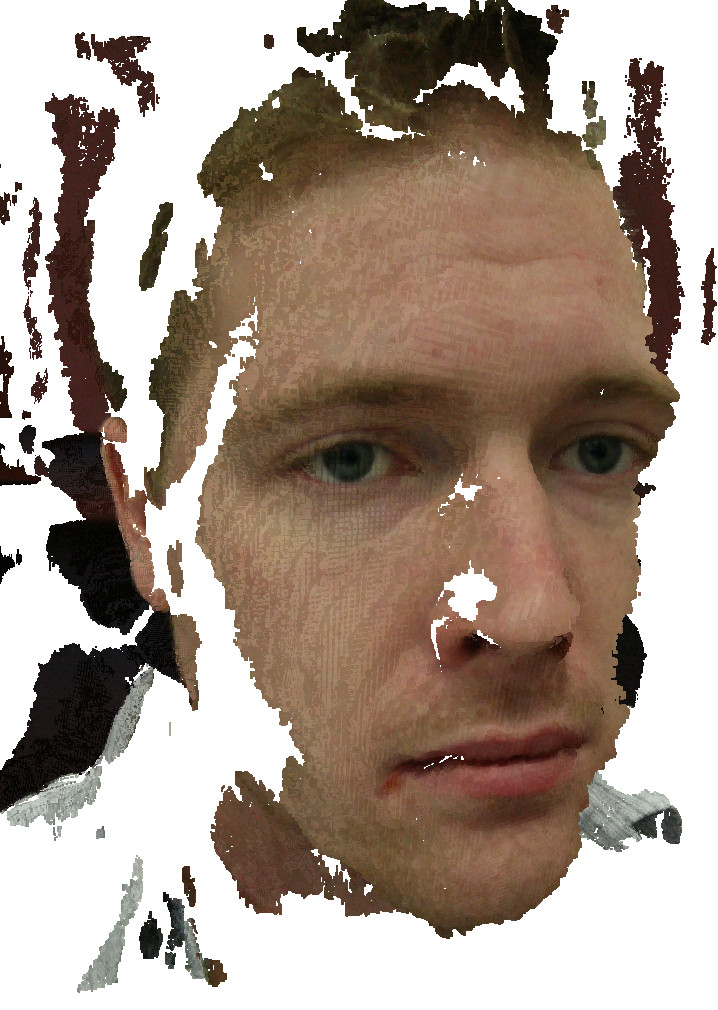} &  
\includegraphics[height = 3.0cm]{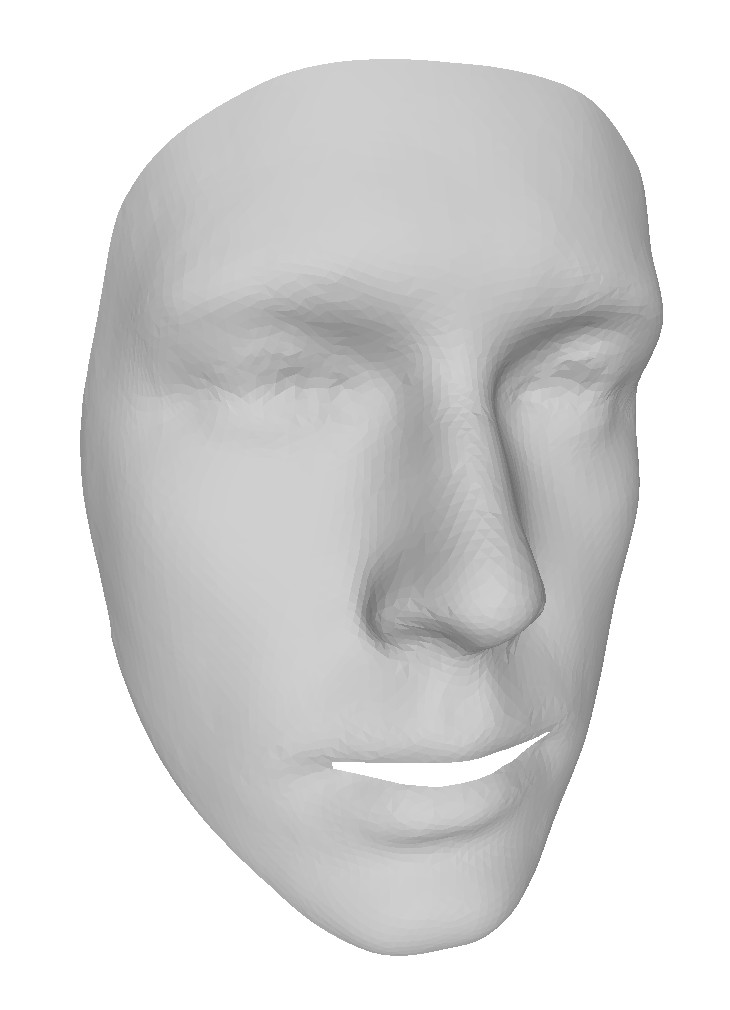} &
\includegraphics[height = 3.0cm]{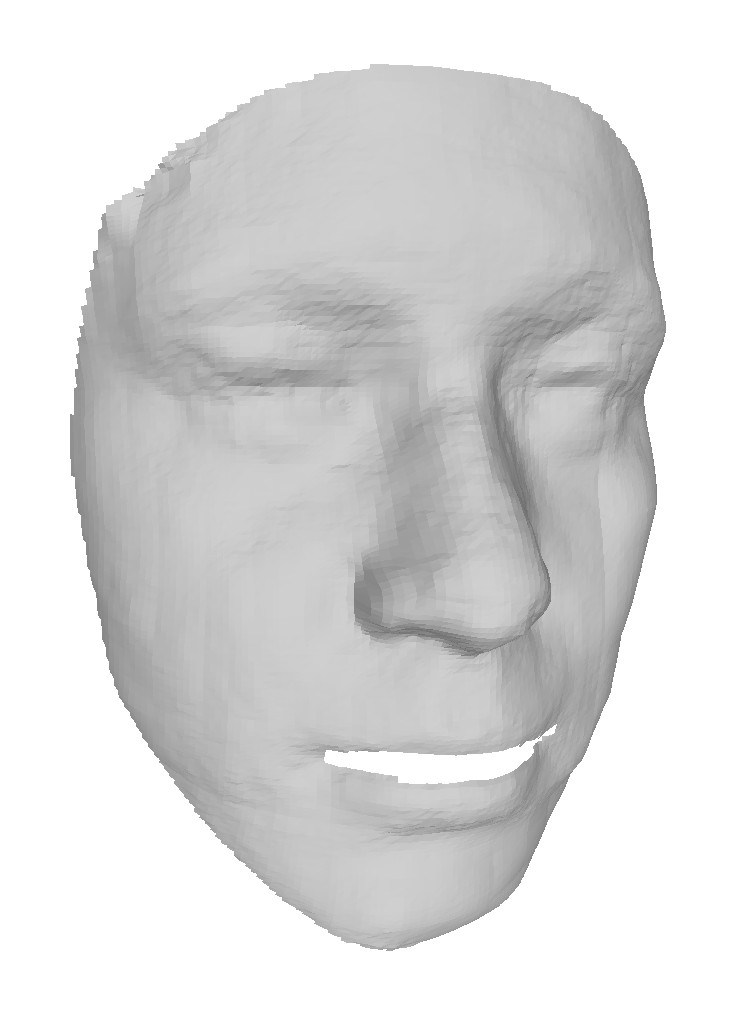}\\

\includegraphics[height = 3.0cm]{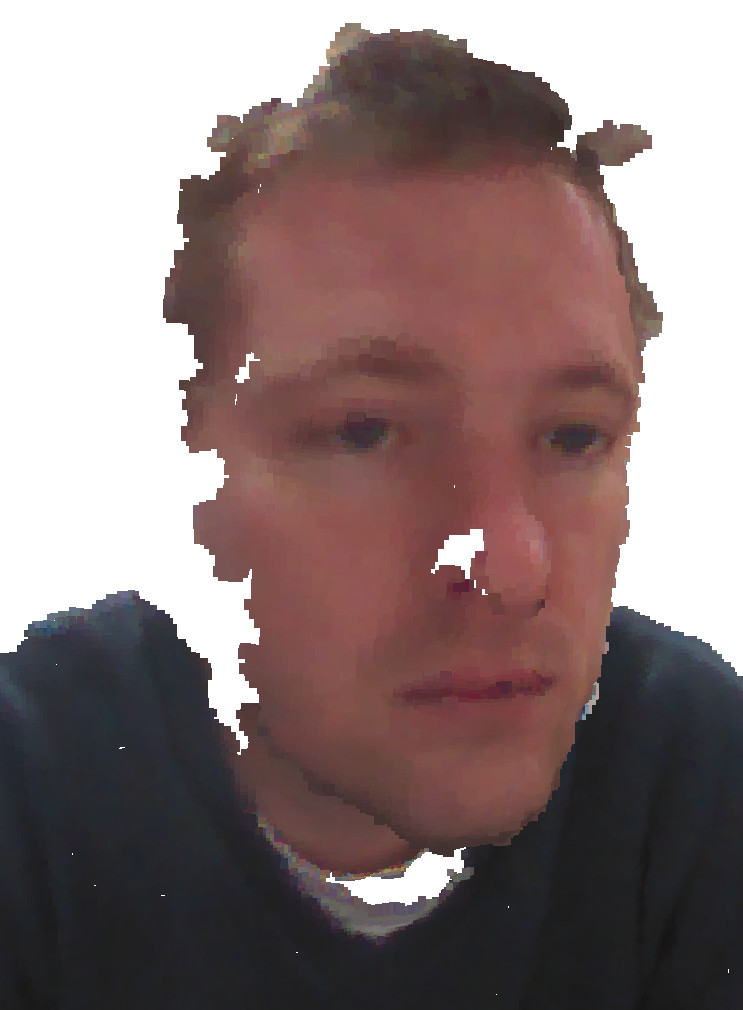} &  
\includegraphics[height = 3.0cm]{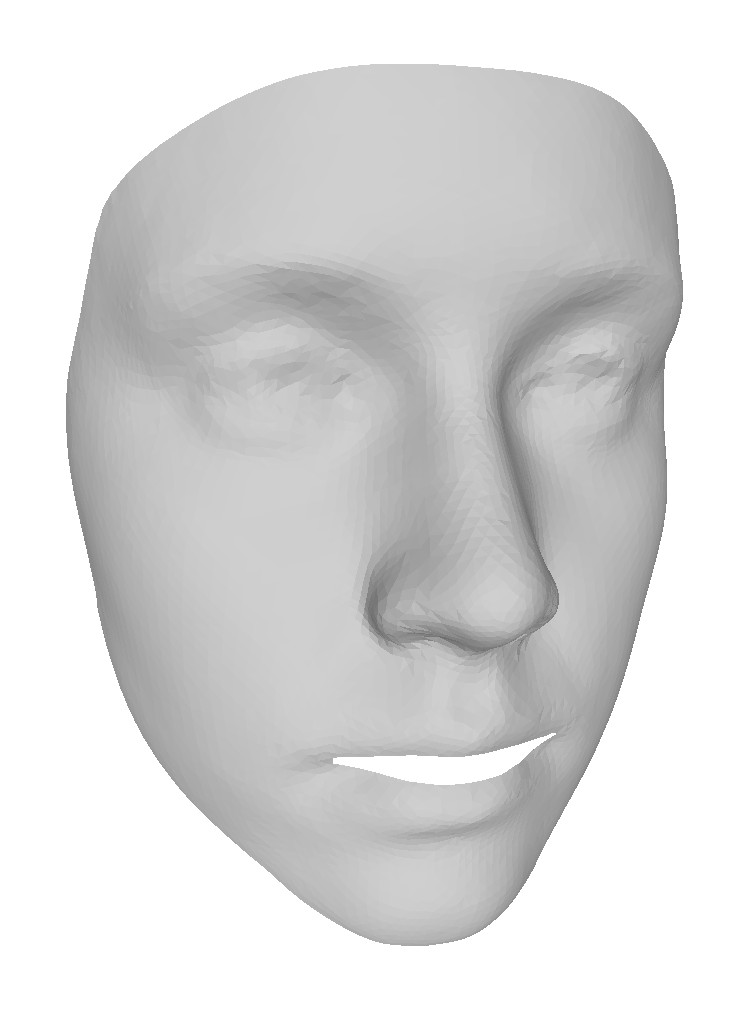} &
\includegraphics[height = 3.0cm]{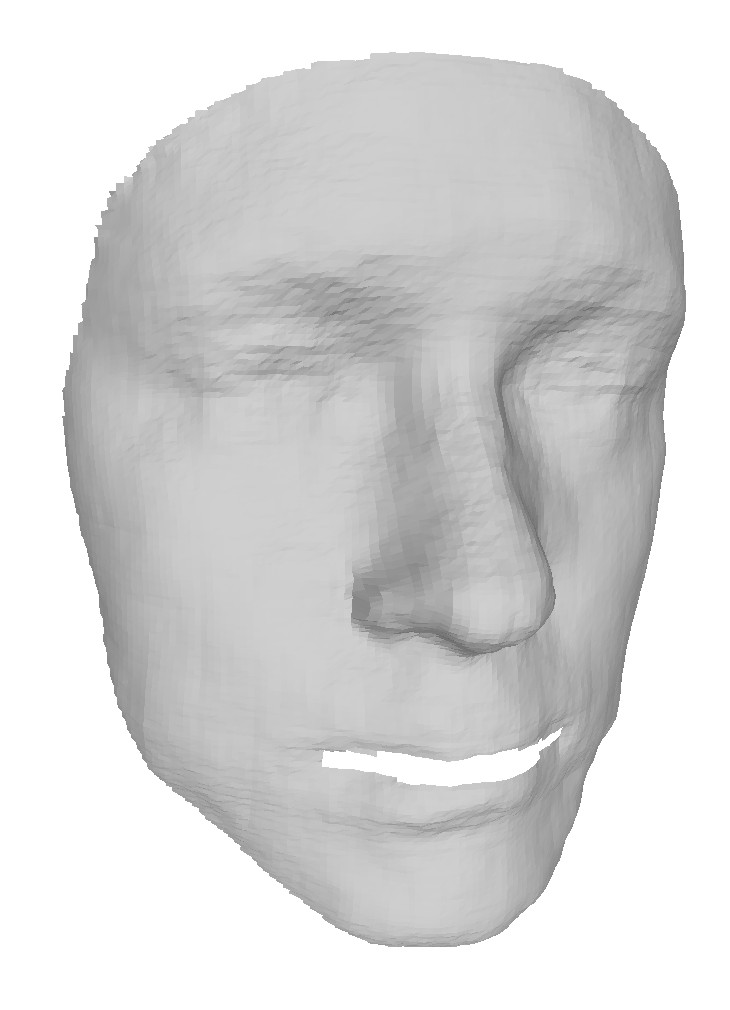}\\

{\small input} & {\small global model} & {\small local model} \\
\end{tabular}
\caption{\emph{Fitting to noisy data. Top row: stereo data. Bottom row: Kinect data. From left to right: input point cloud, result of global fitting, result of local fitting.}}
\label{fig:stereo_and_kinect}
\end{figure}

Figure~\ref{fig:stereo_and_kinect} shows two results obtained using noisy and incomplete stereo and range data. The 3D stereo data used as input to our comparison is obtained using the approach by Brunton et al.~\cite{brunton_etal_stereo_3dimpvt_2012} from two input images. The resulting point cloud has missing data, which is typical for data obtained using passive stereo approaches. The range data used as input to our comparison is obtained using a Kinect sensor. This dataset has low resolution, missing data, and significant data noise. In spite of these problems, both models fit the shape to the input data well in both cases. As in previous experiments, the result using the global model contains less localized shape detail than the result using the local model. 

\subsection{Influence of Fitting Parameters}
Values of fitting parameters used for both models were held constant in this study. This includes the threshold for nearest neighbor distance $\tau$ used in Eq.~(\ref{eqn_nearest_neighbor}) to limit the influence of noise, occlusions, or other outliers in pulling the model to the wrong solution. Another example is the number of standard deviations model parameters are allowed to be from the mean, as controlled by $c$ in Eq.~(\ref{eqn_hyperbox}).

As mentioned in Section \ref{sec_exper_setup}, in this study the settings $\tau=10\mbox{mm}$ and $c=1.0$ were used. These parameters influence the two models in different ways. For example, allowing greater deviation from the mean ($c>1.0$) would allow the global model to capture more detail present in the input. Since both models treat the model parameters as independent, and the global model has many fewer parameters than the local one, setting all parameters to the same distance from the mean results in a much higher probability in the global case than in the local one. On the other hand, allowing a higher threshold for nearest neighbor distance ($\tau>10\mbox{mm}$) would allow the local model to fit better to the nose in some scans. Due to the (least-squares) rigid initial alignment, the tip and ridge of the nose of the model template are often beyond this threshold from the input point cloud, and the local influence of the model parameters means the shape of the rest of the face cannot pull this part of the model towards the data.

The parameters specific to each model were chosen based on the application to faces, or from the training data. The number of principle components kept in the global model was chosen based on the measures of the statistical model given in Section \ref{sec_model_training_eval}. The dimensions of the base grid of the local model was chosen so that the a single scaling coefficient was assigned to the approximate area of important facial features. Choosing a coarser base grid might allow global shape properties to be better reconstructed. The number of samples in the optimization of the local model was chosen to balance precision versus running time.

\section{Implications and Observations}
\label{sec_impl_obs}
Let us now consider possible applications, and how this study can provide insight into which statistical model to use for a particular application. We have seen that increased localization and decorrelation in the wavelet model allow better generalization and therefore better detail recovery in the fitting process. Such detail can be important in applications such as ergonomic design (eg. designing sizes for eye-glasses that sufficiently cover a population from a relatively small training set), tele-presence, and detailed facial capture in the entertainment industry, where state of the art methods require subject specific models~\cite{Garrido_2013} or sophisticated setups~\cite{Beeler_2011}. On the other hand, less detail and a lower-dimensional model are generally preferable for applications such as face and expression recognition~\cite{Scherbaum_2013_ICCV,MpiperisMS08}, model-based manipulation of images and videos~\cite{Blanz1999,Vlasic2005,Dale2011,Yang2012}, and control of virtual avatars~\cite{li2013realtime}.

By making our statistical models public~\cite{bsbw:statmods:2013}, along with code to use them, we make it possible for researchers with knowledge of specific applications to test them thoroughly as they see fit, for their intended application. The provided code uses the bare-bones fitting energy we have used in this paper to illuminate the differences between the two types of models. However, it should be straightforward for others to modify the code to experiment with adding more terms to this energy, such as smoothing terms, landmark terms, or descriptor matching terms. 

An interesting observation in Section \ref{sec_comp_eval_initialize} is that performing initial alignment with the spin-image + RANSAC approach~\cite{JohnsonHebert1997} results in equal or better surface fitting and landmark errors than using manual landmarks. This means that for neutral expression, rather well-known methods suffice to rigidly align face scans to a model. It also likely reflects the fact that landmarks in the test database~\cite{bosphorus}, were placed digitally by clicking the scan rather than placed physically by an anthropometrist.

%--------------------------------------------------------------------------------------------------------------------------------------------------------------------------------------------------------
\section{Conclusion}
\label{sec_conclusions}

In this paper we have reviewed different statistical shape models in the focused context of 3D data fitting, and performed a comparative analysis, both theoretically and experimentally, of global and local statistical shape models for fitting to 3D face data. We have found the following differences between the two types of models: Local models capture details better at the cost of greater computational requirements. This is in part due to the optimization strategy used in this investigation (a sampling strategy that avoids local minima), but is also due to the much higher dimensionality of the local model. The global model has much lower dimensionality and can thus be fitted to input data much faster. In some cases, the global model better captures the overall shape, height and width, of the face. The local model avoids overfitting, because local surface patches are not likely to be biased for a particular database the way the shape of the entire face can be; local surface patches from human faces have much lower shape variation than entire faces, hence a limited training set has a better chance of capturing the full variability in the local model. The local model also better contains erroneous reconstruction due to occlusion to the affected areas, whereas the global model typically captures approximately symmetric shapes of human faces; an occlusion of the left side will cause poor fitting on the right as well. The local model can capture additional details by subdivision resampling.

While to a large extent, these conclusions reflect the motivations behind the use of these models, our observations about the dimensionality of the shape space with respect to the size of the training set provide useful insight into the appropriate choice of model for practitioners. If a limited number of training samples are available, the local wavelet model is preferable because each training sample will be decomposed into many independent low-dimensional samples. Conversely, if a vast amount of parametrized data is available, the global model may be preferable. If for a particular class of shapes, a part-based decomposition can be reliably and efficiently obtained, a part-based model may be preferable. This is reflected in how different models have been preferred in the literature for different types of data: for human faces and bodies, global and part-based models have traditionally been preferred, whereas for medical data, local models are more commonly used.

Applications of statistical model fitting to 3D data include face or body recognition, expression or pose recognition, biometric passwords, and virtual change rooms.
In this study, we have kept the focus on model fitting to static data, however, tracking of noisy dynamic point clouds with occlusion remains an open problem. 

Looking forward, sparse statistical models are a rapidly developing and expanding research area, and the use of sparsity inducing priors in statistical model fitting of 3D data provides many open avenues for future research. Initial steps have been taken in this direction~\cite{neumann_2013}, although so far this is limited to sparsity through locality. Recent methods for biological data formulate finding the basis of the shape space and attaching the statistical prior as an optimization problem that allows to trade-off between locality and compactness~\cite{alcantara_pami2009}. Further, simultaneous parametrization and model learning has only been addressed by a few existing methods~\cite{davies_twining_taylor_mdl,Hirshberg_2012}. There is no doubt room for more innovation in this direction.

%--------------------------------------------------------------------------------------------------------------------------------------------------------------------------------------------------------
\section*{Acknowledgments}
We thank Eric Dubois, Jochen Lang, Chang Shu, Michael Wand, and Tino Weinkauf for helpful discussions. We thank the anonymous reviewers for their helpful and insigntful feedback and suggestions. This work has been partially funded by the Cluster of Excellence on \textit{Multimodal Computing and Interaction} within the Excellence Initiative of the German Federal Government. 

\small{

}


\begin{thebibliography}{10}
\expandafter\ifx\csname url\endcsname\relax
  \def\url#1{\texttt{#1}}\fi
\expandafter\ifx\csname urlprefix\endcsname\relax\def\urlprefix{URL }\fi
\expandafter\ifx\csname href\endcsname\relax
  \def\href#1#2{#2} \def\path#1{#1}\fi

\bibitem{bsbw:statmods:2013}
T.~Bolkart, A.~Brunton, A.~Salazar, S.~Wuhrer,
  \href{http://statistical-face-models.mmci.uni-saarland.de/}{Statistical 3d
  shape models of human faces} (2013).
\newline\urlprefix\url{http://statistical-face-models.mmci.uni-saarland.de/}

\bibitem{Li2010}
H.~Li, T.~Weise, M.~Pauly, Example-based facial rigging, ACM Transactions on
  Graphics 29~(4) (2010) 32:1--6.

\bibitem{VanKaick2011}
O.~van Kaick, H.~Zhang, G.~Hamarneh, D.~Cohen-Or, A survey on shape
  correspondence, Computer Graphics Forum 30~(6) (2011) 1681--1707.

\bibitem{tam_survey_2013}
G.~Tam, Z.-Q. Cheng, Y.-K. Lai, F.~Langbein, Y.~Liu, D.~Marshall, R.~Martin,
  X.-F. Sun, P.~Rosin, Registration of 3d point clouds and meshes: A survey
  from rigid to non-rigid, IEEE Transactions on Visualization and Computer
  Graphics 19~(7) (2013) 1199--1217.

\bibitem{cashman_fitzgibbon_shape_dolphins}
T.~Cashman, A.~Fitzgibbon, What shape are dolphins? {B}uilding 3d morphable
  models from 2d images, {IEEE} Transactions on Pattern Analysis and Machine
  Intelligence 35 (2013) 232--244.

\bibitem{alcantara_pami2009}
D.~A. Alcantara, O.~Carmichael, W.~Harcourt-Smith, K.~Sterner, S.~R. Frost,
  R.~Dutton, P.~Thompson, E.~Delson, N.~Amenta, Exploration of shape variation
  using localized components analysis, IEEE Transactions on Pattern Analysis
  and Machine Intelligence 31~(8) (2009) 1510--1516.

\bibitem{Blanz1999}
V.~Blanz, T.~Vetter, A morphable model for the synthesis of 3d faces, in: ACM
  Conference on Computer Graphics and Interactive Techniques, 1999, pp.
  187--194.

\bibitem{amberg_etal_iccv07}
B.~Amberg, A.~Blake, A.~Fitzgibbon, S.~Romdhani, T.~Vetter, Reconstructing high
  quality face-surfaces using model based stereo, in: {IEEE} International
  Conference on Computer Vision, 2007, pp. 1--8.

\bibitem{patel_smith_morphableModelRevisited_09}
A.~Patel, W.~Smith, 3d morphable face models revisited, in: IEEE Conference on
  Computer Vision and Pattern Recognition, 2009, pp. 1327--1334.

\bibitem{Yang2011}
F.~Yang, J.~Wang, E.~Shechtman, L.~Bourdev, D.~Metaxas, Expression flow for
  3d-aware face component transfer, ACM Transactions on Graphics 30~(4) (2011)
  60:1--10.

\bibitem{amberg_etal_fg08}
B.~Amberg, R.~Knothe, T.~Vetter, Expression invariant {3D} face recognition
  with a morphable model, in: {IEEE} International Conference on Automatic Face
  and Gesture Recognition, 2008, pp. 1--6.

\bibitem{Vlasic2005}
D.~Vlasic, M.~Brand, H.~Pfister, J.~Popovi\'{c}, Face transfer with multilinear
  models, ACM Transactions on Graphics 24~(3) (2005) 426--433.

\bibitem{Dale2011}
K.~Dale, K.~Sunkavalli, M.~Johnson, D.~Vlasic, W.~Matusik, H.~Pfister, Video
  face replacement, ACM Transactions on Graphics 30~(6) (2011) 130:1--10.

\bibitem{Yang2012}
F.~Yang, L.~Bourdev, J.~Wang, E.~Shechtman, D.~Metaxas, Facial expression
  editing in video using a temporally-smooth factorization, in: {IEEE}
  International Conference on Computer Vision and Pattern Recognition, 2012,
  pp. 861--868.

\bibitem{Bolkart2013}
T.~Bolkart, S.~Wuhrer, Statistical analysis of 3d faces in motion, in: {IEEE}
  International Conference on {3D} Vision, 2013, pp. 103--110.

\bibitem{basso_verri_2007}
C.~Basso, A.~Verri, Fitting 3d morphable models using implicit representations,
  in: Conference on Computer Vision, Imaging and Computer Graphics Theory and
  Applications, 2007, pp. 45--52.

\bibitem{haar_veltkamp_2008}
F.~ter Haar, R.~Veltkamp, 3d face model fitting for recognition, in: European
  Conference on Computer Vision, 2008, pp. 652--664.

\bibitem{smet_vanGool_2010}
M.~Smet, L.~V. Gool, Optimal regions for linear model-based 3d face
  reconstruction, in: Asian Conference on Computer Vision, 2010, pp. 276--289.

\bibitem{kakadiaris_etal_2007_deformable_model}
I.~Kakadiaris, G.~Passalis, G.~Toderici, M.~Murtuza, Y.~Lu, N.~Karamelpatzis,
  T.~Theoharis, Three-dimensional face recognition in the presence of facial
  expressions: An annotated deformable model approach, IEEE Transactions on
  Pattern Analysis and Machine Intelligence 29~(4) (2007) 640--649.

\bibitem{Golovinskiy_2006}
A.~Golovinskiy, W.~Matusik, H.~Pfister, S.~Rusinkiewicz, T.~Funkhouser, A
  statistical model for synthesis of detailed facial geometry, ACM Transactions
  on Graphics 25~(3) (2006) 1025--1034.

\bibitem{Brunton2011}
A.~Brunton, C.~Shu, J.~Lang, E.~Dubois, Wavelet model-based stereo for fast,
  robust face reconstruction, in: Canadian Conference on Computer and Robot
  Vision, 2011, pp. 347--354.

\bibitem{Allen2003}
B.~Allen, B.~Curless, Z.~Popovi\'{c}, The space of human body shapes:
  reconstruction and parameterization from range scans, ACM Transactions on
  Graphics 22~(3) (2003) 587--594.

\bibitem{seo_etal_shape_from_silhouette}
H.~Seo, Y.~I. Yeo, K.~Wohn, 3d body reconstruction from photos based on range
  scan, Technologies for E-Learning and Digital Entertainment (2006) 849--860.

\bibitem{chen:learning}
Y.~Chen, R.~Cipolla, Learning shape priors for single view reconstruction, in:
  {IEEE} International Workshop on 3-D Digital Imaging and Modeling, 2009, pp.
  1425--1432.

\bibitem{boisvert_etal_shape_from_silhouette}
J.~Boisvert, C.~Shu, S.~Wuhrer, P.~Xi, Three-dimensional human shape inference
  from silhouettes: Reconstruction and validation, Machine Vision and
  Applications 24~(1) (2013) 145--157.

\bibitem{wuhrer_shu_acc_shape_measurement}
S.~Wuhrer, C.~Shu, Estimating 3d human shapes from measurements, Machine Vision
  and Applications 24~(6) (2013) 1133--1147.

\bibitem{anguelov_srinivasan_koller_thrun_rodgers_05_shapecomp}
D.~Anguelov, P.~Srinivasan, D.~Koller, S.~Thrun, J.~Rodgers, J.~Davis, Scape:
  shape completion and animation of people, ACM Transactions on Graphics 24~(3)
  (2005) 408--416.

\bibitem{guan_etal}
P.~Guan, A.~Weiss, A.~O. Balan, M.~J. Black, Estimating human shape and pose
  from a single image, in: International Conference on Computer Vision, 2009,
  pp. 1381--1388.

\bibitem{balan_black_08_naked_truth}
A.~Balan, M.~Black, The naked truth: Estimating body shape under clothing, in:
  European Conference on Computer Vision, 2008, pp. 15--29.

\bibitem{ParametricReshaping2010}
S.~Zhou, H.~Fu, L.~Liu, D.~Cohen-Or, X.~Han, Parametric reshaping of human
  bodies in images, ACM Transactions on Graphics 29~(4) (2010) 126:1--10.

\bibitem{Jain:2010:MovieReshape}
A.~Jain, T.~Thorm\"{a}hlen, H.-P. Seidel, C.~Theobalt, Movie{R}eshape: Tracking
  and reshaping of humans in videos, ACM Transactions on Graphics 29~(5) (2010)
  148:1--10.

\bibitem{weiss_etal_bodyShapeFromKinect_2011}
A.~Weiss, D.~Hirshberg, M.~Black, Home 3d body scans from noisy image and range
  data, in: International Conference on Computer Vision, 2011, pp. 1951--1958.

\bibitem{Hirshberg_2012}
D.~Hirshberg, M.~Loper, E.~Rachlin, M.~Black, Coregistration: Simultaneous
  alignment and modeling of articulated 3d shape, in: European Conference on
  Computer Vision, 2012, pp. 242--255.

\bibitem{HasStoSunRosSei09}
N.~Hasler, C.~Stoll, M.~Sunkel, B.~Rosenhahn, H.-P. Seidel, A statistical model
  of human pose and body shape, Computer Graphics Forum 28~(2) (2009) 337--346.

\bibitem{hasler_etal_smi09}
N.~Hasler, C.~Stoll, B.~Rosenhahn, T.~Thorm\"{a}hlen, H.-P. Seidel, Estimating
  body shape of dressed humans, Computers \& Graphics 33~(3) (2009) 211--216.

\bibitem{Hasler2010}
N.~Hasler, H.~Ackermann, B.~Rosenhahn, T.~Thorm{\"a}hlen, H.-P. Seidel,
  Multilinear pose and body shape estimation of dressed subjects from image
  sets, in: IEEE Conference on Computer Vision and Pattern Recognition, 2010,
  pp. 1823--1830.

\bibitem{wuhrer_etal_2012_pose_inv_statistics}
S.~Wuhrer, C.~Shu, P.~Xi, Posture-invariant statistical shape analysis using
  laplace operator, Computers \& Graphics 36 (2012) 410--416.

\bibitem{Xi_etal_segmented_body_2007}
P.~Xi, W.-S. Lee, C.~Shu, A data-driven approach to human body cloning using a
  segmented body database, in: Pacific Graphics, 2007, pp. 139--147.

\bibitem{Chen_2013_CVPR}
Y.~Chen, Z.~Liu, Z.~Zhang, Tensor-based human body modeling, in: {IEEE}
  International Conference on Computer Vision and Pattern Recognition, 2013,
  pp. 105--112.

\bibitem{cootes_etal_95_ASM}
T.~Cootes, C.~Taylor, D.~Cooper, J.~Graham, Active shape models -- their
  training and application, Computer Vision and Image Understanding 61~(1)
  (1995) 38--59.

\bibitem{cootes_taylor_01_asm}
T.~Cootes, C.~Taylor, Statistical models of appearance for medical image
  analysis and computer vision, in: SPIE Medical Imaging, 2001, pp. 236--248.

\bibitem{davies_twining_taylor_mdl}
R.~Davies, C.~Twining, C.~Taylor, Statistical Models of Shape: Optimisation and
  Evaluation, Springer, 2008.

\bibitem{fletcher_etal_pga_2004}
P.~Fletcher, C.~Lu, S.~Pizer, S.~Joshi, Principal geodesic analysis for the
  study of nonlinear statistics of shape, IEEE Transactions on Medical Imaging
  23~(8) (2004) 995--1005.

\bibitem{toews_etal_06}
M.~Toews, D.~Collins, T.~Arbel, A statistical parts-based appearance model of
  inter-subject variability, in: International Conference on Medical Image
  Computing and Computer Assisted Intervention, 2006, pp. 232--–240.

\bibitem{lecron_etal_2012}
F.~Lecron, J.~Boisvert, S.~Mahmoudi, H.~Labelle, M.~Benjelloun, Fast 3d spine
  reconstruction of postoperative patients using a multilevel statistical
  model, in: International Conference on Medical Image Computing and Computer
  Assisted Intervention, 2012, pp. 446--453.

\bibitem{davatzikos_etal}
C.~Davatzikos, X.~Tao, D.~Shen, Hierarchical active shape models, using the
  wavelet transform, IEEE Transactions on Medical Imaging 22~(3) (2003)
  414--423.

\bibitem{nain_etal_MICCAI05}
D.~Nain, S.~Haker, A.~Bobick, A.~Tannenbaum, Multiscale 3d shape analysis using
  spherical wavelets, in: International Conference on Medical Image Computing
  and Computer Assisted Intervention, 2005, pp. 459--467.

\bibitem{nain_etal_MICCAI06}
D.~Nain, S.~Haker, A.~Bobick, A.~Tannenbaum, Shape-driven 3d segmentation using
  spherical wavelets, in: International Conference on Medical Image Computing
  and Computer Assisted Intervention, 2006, pp. 66--74.

\bibitem{li_etal_CVPR07}
Y.~Li, T.-S. Tan, I.~Volkau, W.~Nowinski, Model-guided segmentation of {3D}
  neuroradiological image using statistical surface wavelet model, in: {IEEE}
  International Conference on Computer Vision and Pattern Recognition, 2007,
  pp. 1--7.

\bibitem{hierarchical-diffusion-wavelet-shape-priors}
S.~Essafi, G.~Langs, Hierarchical {3D} diffusion wavelet shape priors, in:
  {IEEE} International Conference on Computer Vision, 2009, pp. 1717--1724.

\bibitem{cortical_folding}
P.~Yu, B.~T.~T. Yeo, P.~E. Grant, B.~Fischl, P.~Golland, Cortical folding
  development study based on over-complete spherical wavelets, in: {IEEE}
  International Conference on Computer Vision, 2007, pp. 1--8.

\bibitem{Weise2011}
T.~Weise, S.~Bouaziz, H.~Li, M.~Pauly, Realtime performance-based facial
  animation, ACM Transactions on Graphics 30~(4) (2011) 77:1--10.

\bibitem{dryden_mardia_shape_analysis}
I.~Dryden, K.~Mardia, Statistical Shape Analysis, Wiley, 2002.

\bibitem{Styner2003}
M.~Styner, K.~Rajamani, L.-P. Nolte, G.~Zsemlye, G.~Szekely, C.~Taylor,
  R.~Davies, Evaluation of 3d correspondence methods for model building, in:
  Information Processing in Medical Imaging, 2003, pp. 63--75.

\bibitem{sweldens_lifting_1996}
W.~Sweldens, The lifting scheme: A custom-design construction of biorthogonal
  wavelets, Applied and Computational Harmonic Analysis 3~(2) (1996) 186--200.

\bibitem{mallat_wavelet_tour_1999}
S.~Mallat, A Wavelet Tour of Signal Processing, Elsevier, 1999.

\bibitem{spherical_wavelets}
P.~Schr\"{o}der, W.~Sweldens, Spherical wavelets: Efficiently representing
  functions on the sphere, in: {ACM} Conference on Computer Graphics and
  Interactive Techniques, 1995, pp. 161--172.

\bibitem{bspline_subdiv_wavelets}
M.~Bertram, M.~Duchaineau, B.~Hamann, K.~I. Joy, Generalized {B-Spline}
  subdivision-surface wavelets for geometry compression, {IEEE} Transactions on
  Visualization and Graphics 10~(3) (2004) 326--338.

\bibitem{creusot_landmark_labelling_3dor2010}
C.~Creusot, N.~Pears, J.~Austin, {3D} face landmark labelling, in: {ACM}
  Workshop on {3D} Object Retrieval, 2010, pp. 27--32.

\bibitem{salazar_auto_expr_face_reg_arxiv2012}
A.~Salazar, S.~Wuhrer, S.~Chu, F.~Prieto, Fully automatic expression-invariant
  face correspondence, Machine Vision and Applications 25~(4) (2014) 859--879.

\bibitem{JohnsonHebert1997}
A.~Johnson, M.~Hebert, Recognizing objects by matching oriented points, in:
  Conference on Computer Vision and Pattern Recognition, 1997, pp. 684--692.

\bibitem{FischlerBolles1981}
M.~Fischler, R.~Bolles, Random sample consensus: a paradigm for model fitting
  with applications to image analysis and automated cartography, Communications
  of the ACM 24~(6) (1981) 381--395.

\bibitem{ANN}
D.~Mount, S.~Arya, \href{http://www.cs.umd.edu/~mount/ANN/}{{ANN}: A library
  for approximate nearest neighbor searching} (2010).
\newline\urlprefix\url{http://www.cs.umd.edu/~mount/ANN/}

\bibitem{patel_explore_id_manifold_eccv2010}
A.~Patel, W.~Smith, Exploring the identity manifold: Constrained operations in
  face space, in: European Conference on Computer Vision, 2010, pp. 112--125.

\bibitem{liu_nocedal_lbfgsb}
D.~Liu, J.~Nocedal, On the limited memory method for large scale optimization,
  Mathematical Programming 45~(3) (1989) 503--528.

\bibitem{lee_kdquadtree_1977}
D.~Lee, C.~Wong, Worst-case analysis for region and partial region searches in
  multidimensional binary search trees and balanced quad trees, Acta
  Informatica 9~(1) (1977) 23--29.

\bibitem{bu-3dfe}
L.~Yin, X.~We, Y.~Sun, J.~Wang, M.~Rosato, A {3D} facial expression database
  for facial behavior research, in: {IEEE} International Conference on
  Automatic Face and Gesture Recognition, 2006, pp. 211--216.

\bibitem{bosphorus}
A.~Savran, N.~Alyuz, H.~Dibeklioglu, O.~Celiktutan, B.~G\"{o}kberk, B.~Sankur,
  L.~Akarun, Bosphorus database for {3D} face analysis, in: Workshop on
  Biometrics and Identity Management, 2008, pp. 47--56.

\bibitem{brunton_etal_stereo_3dimpvt_2012}
A.~Brunton, J.~Lang, E.~Dubois, Efficient multi-scale stereo of high-resolution
  planar and spherical images, in: IEEE Conference on 3D Imaging, Modeling,
  Processing, Visualization, and Transmission, 2012, pp. 120--127.

\bibitem{Garrido_2013}
P.~Garrido, L.~Valgaerts, C.~Wu, C.~Theobalt, Reconstructing detailed dynamic
  face geometry from monocular video, ACM Transactions on Graphics 32~(6)
  (2013) 158:1--10.

\bibitem{Beeler_2011}
T.~Beeler, F.~Hahn, D.~Bradley, B.~Bickel, P.~Beardsley, C.~Gotsman, R.~W.
  Sumner, M.~Gross, High-quality passive facial performance capture using
  anchor frames, ACM Transactions on Graphics 30 (2011) 75:1--10.

\bibitem{Scherbaum_2013_ICCV}
K.~Scherbaum, J.~Petterson, R.~S. Feris, V.~Blanz, H.-P. Seidel, Fast face
  detector training using tailored views, in: {IEEE} International Conference
  on Computer Vision, 2013.

\bibitem{MpiperisMS08}
I.~Mpiperis, S.~Malassiotis, M.~G. Strintzis, Bilinear models for 3-d face and
  facial expression recognition, IEEE Transactions on Information Forensics and
  Security 3~(3) (2008) 498--511.

\bibitem{li2013realtime}
H.~Li, J.~Yu, Y.~Ye, C.~Bregler, Realtime facial animation with on-the-fly
  correctives, ACM Transactions on Graphics 32~(4).

\bibitem{neumann_2013}
T.~Neumann, K.~Varanasi, S.~Wenger, M.~Wacker, M.~Magnor, C.~Theobalt, Sparse
  localized deformation components, {ACM} Transactions on Graphics 32~(6)
  (2013) 179:1--10.

\end{thebibliography}
\end{document}